%% file: main.tex
\documentclass[runningheads]{llncs}

\usepackage{eccv}

\usepackage{eccvabbrv}
\input{preamble}
\usepackage{graphicx}
\usepackage{booktabs}

\usepackage[accsupp]{axessibility}  

\usepackage{hyperref}

\usepackage{orcidlink}

\newcommand{\TIP}{SPIRE}
\newcommand{\Processing}{Restoration}

\newcommand{\textdrivenimagerestoration}{Semantic Prompt-Driven Image Restoration}
\newcommand{\underlinetextdrivenimagerestoration}{\textbf{\underline{S}}emantic \textbf{\underline{P}}rompt-Driven \textbf{\underline{I}}mage \textbf{\underline{RE}}storation}

\begin{document}

\title{{SPIRE: Semantic Prompt-Driven Image Restoration}}


\author{Chenyang Qi\inst{1,2}*\orcidlink{0000-0002-6462-6534} \and
Zhengzhong Tu\inst{1}\orcidlink{0000-0002-7594-2292} \and
Keren Ye\inst{1}\orcidlink{0000-0002-7349-7762}
Mauricio Delbracio\inst{1}\orcidlink{0000-0001-7539-2991} \and
Peyman Milanfar\inst{1} \and
Qifeng Chen\inst{2}\orcidlink{0000-0003-2199-3948} \and
Hossein Talebi\inst{1}
}
\authorrunning{Qi et al.}

\institute{ Google Research \and HKUST}








\input{sec/figs/teaser}
\let\thefootnote\relax\footnote{*This work was done during an internship at Google Research.}

\input{sec/0_abstract}

\input{sec/1_intro}
\input{sec/2_related}

\input{sec/3_method}

\input{sec/4_exp}

\input{sec/5_conclusion}








\bibliographystyle{splncs04}
\bibliography{main}

\newpage
\appendix
\input{sec/X_suppl}


















\end{document}

%% file: preamble.tex
\usepackage[dvipsnames]{xcolor}
\usepackage{algorithm}
\usepackage{algpseudocode}
\usepackage{enumitem}

\usepackage{wrapfig}

\newcommand{\todo}[1]{{\color{red}[todo: #1]}}

\usepackage{lipsum}
\usepackage{arydshln}
\usepackage{multirow}
\usepackage{comment}

\usepackage{amsmath} 
\usepackage{bm}

\usepackage{bbding}



%% file: sec/figs/teaser.tex


{
\maketitle

\begin{center}
    \captionsetup{type=figure}
    \vspace{-1em}
\newcommand{\imwidth}{0.99\textwidth}
\newcommand{\imheight}{0.6\textwidth}

\parbox{\imwidth}{\includegraphics[width=\imwidth,height=\imheight]{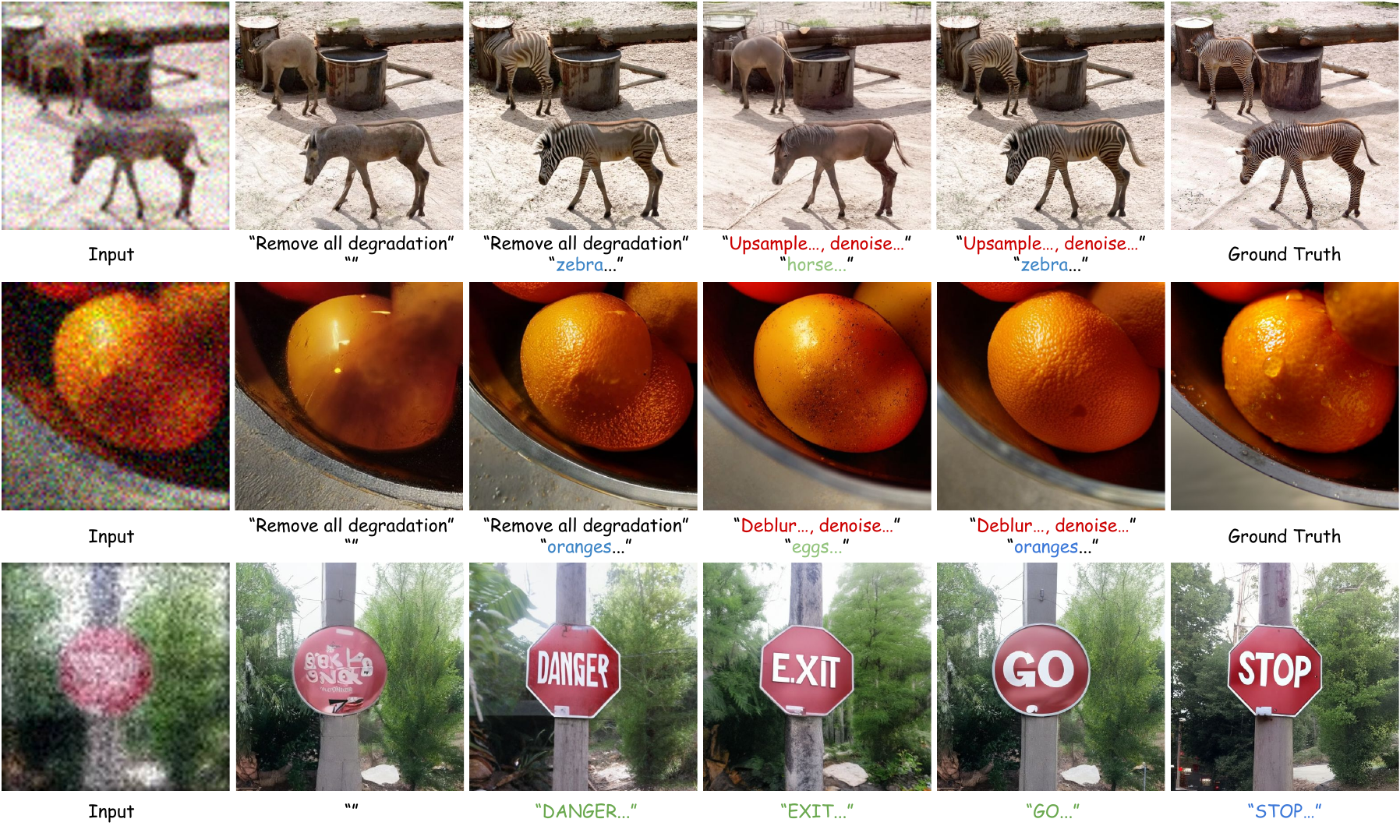}}
\vspace{-1em}
\captionof{figure}{We present 
\textbf{\TIP{}: 
\textdrivenimagerestoration{}
}, a text-based foundation model for all-in-one, instructed image restoration. \TIP{} allows users to flexibly leverage either semantic-level content prompt, or quantitative degradation-aware restoration prompt, or both, to obtain their desired enhancement results based on personal preferences.
In other words, \TIP{} can be easily prompted to conduct blind restoration, semantic restoration, or task-specific granular treatment.
Our framework also enables a new paradigm of \textbf{instruction-based image restoration}, providing a reliable evaluation benchmark to facilitate vision-language models for low-level computational photography applications.
}
\vspace{-3em}
\label{fig:teaser}
\end{center}

}

%% file: sec/0_abstract.tex
\begin{abstract}
Text-driven diffusion models have become increasingly popular for various image editing tasks, including inpainting, stylization, and object replacement. 
However, it still remains an open research problem to adopt this language-vision paradigm for more fine-level image processing tasks, such as denoising, super-resolution, deblurring, and compression artifact removal. 
In this paper, we develop \TIP{}, a Semantic and restoration Prompt-driven Image \Processing{} framework that leverages natural language as a user-friendly interface to control the image restoration process. We consider the capacity of prompt information in two dimensions. 
First, we use content-related prompts to enhance the semantic alignment, effectively alleviating identity ambiguity in the restoration outcomes.
Second, our approach is the first framework that supports fine-level instruction through language-based quantitative specification of the restoration strength, without the need for explicit task-specific design.
In addition, we introduce a novel fusion mechanism that augments the existing ControlNet architecture by learning to rescale the generative prior, thereby achieving better restoration fidelity. 
Our extensive experiments demonstrate the superior restoration performance of \TIP{} compared to the state of the arts, alongside offering the flexibility of text-based control over the restoration effects.
\end{abstract}


%% file: sec/1_intro.tex
\section{Introduction}
\label{sec:intro}


Image restoration or enhancement aims to recover high-quality, pixel-level details from a degraded image, while preserving as much the original semantic information as possible. Although neural network models~\cite{wang2021real-esrgan,zhengzhong2022Maxim,zhendong2022Uformer,liang2021swinir,delbracio2021projected,whang2022deblurring,delbracio2023inversion,wang2023stablesr} have marked significant progress, it still remains challenging to design an effective task-conditioning mechanism instead of building multiple individual models for each task (such as denoise, deblur, compression artifact removal) in practical scenarios.
The advancement of text-driven diffusion models~\cite{stable-diffusion, imagen, dalle2} has unveiled the potential of natural language as a universal input condition for a broad range of image processing challenges, which improves interactivity with users and reduces the cost of task-specific fine-tuning. However, the existing applications of natural language in stylization~\cite{pix2pix-zero, null},
and inpainting~\cite{blended,blended_latent, pnp,p2p,pix2pix-zero,diffedit} predominantly focus on high-level semantic editing, whereas the uniqueness and challenges for low-level image processing have been less explored.

Natural language text prompts in image restoration can play two crucial roles: alleviating semantic ambiguities and resolving degradation type ambiguities.
Firstly, the same degraded image can correspond to different visual objects, leading to ambiguities in content interpretation (\eg, discerning whether the blurred animal in~\cref{fig:teaser} is a horse or a zebra). Typically, unconditional image-to-image restoration leads to a random or average estimation of the identities, resulting in neither fish nor fowl.
Secondly, certain photographic effects that are deliberately introduced for aesthetic purposes, such as \emph{bokeh} with a soft out-of-focus background, can be misconstrued as \emph{blur distortions} by many existing deblurring models, leading to unwanted artifacts or severe hallucinations in the outputs.
Although blind restoration methods can produce clean images by either leveraging frozen generative priors~\cite{wang2021gfpgan, Yang2021GPEN, wang2023stablesr}, or using end-to-end regression~\cite{wang2021real-esrgan}, they do not consider the aforementioned ambiguities 
.


In this paper, we introduce \textbf{\TIP{}}---a \underlinetextdrivenimagerestoration{}
framework that provides a user-friendly interface to fully control \underline{both the restored image semantics and the enhancement granularity} using natural language instructions.
Traditional blind restoration without prompts (Figure~\ref{fig:teaser}) has limited flexibility in solving semantic and degradation ambiguity, thus tends to generate average blurry images. 
Leveraging the semantic and quantitative restoration prompts, we show that the visual quality can be significantly improved. Moreover, our framework can also be used as blind restoration with null prompts.
In addition to higher perceptual quality, our framework provides the flexibility for users to generate more than one plausible result. Specifically, in imaging scenarios with large levels of degradation there are multiple plausible solutions. The proposed interactive framework has the capability to personalize the output by leveraging input prompts according to user preference.
In concurrent studies, the focus has typically been isolated to one of three areas: either blind restoration~\cite{wang2023stablesr,Lin2023diffbir}, only semantic prompting~\cite{SUPIR}, or only discrete restoration types~\cite{luo2023controlling, chen2023Promptsr}.
To the best of our knowledge, this is the first unified model to support the following three distinct features simultaneously:
\begin{enumerate}
\item \textbf{Blind Restoration}: When instructed with a general restoration prompt ``remove all degradations'' and empty semantic prompt ``'', \TIP{} operates as a blind restoration model (``Ours w/o text'' in~\cref{tab:merged main-table}).
\item \textbf{Semantic Restoration}: When provided with a text description of the (desired) visual content, \TIP{} concentrates on restoring the specified identity of the uncertain or ambiguous objects in the degraded image.

\item \textbf{Quantitative Task-Specific Restoration}: Receiving specific restoration type hints (\eg, ``deblur...'', ``deblur... denoise...''), \TIP{} transforms into a task-specific model (\eg, deblur, denoise, or both). Moreover, it understands the numeric nuances of different degradation parameters in language prompts so that the users can control the restoration strength (\eg, ``deblur with sigma 3.0'' in~\cref{fig:restoration_decoupling}).
\end{enumerate}
To train \TIP{}, we first build a synthetic data generation pipeline on a large-scale text-image dataset~\cite{PaLI} upon the second-order degradation process proposed in Real-ESRGAN~\cite{wang2021real-esrgan}. Additionally, we embed the degradation parameters into the restoration prompts to encode finer-grained degradation information (e.g., \emph{``deblur with sigma 3.5''}). Our approach provides richer degradation-specific information compared to contemporary works~\cite{jiang2023autodir, luo2023controlling} which only employ the degradation types (\eg, ``gaussian blur'').
We then finetune a ControlNet adaptor~\cite{controlnet}---which learns a parameter-efficient, parallel branch on top of the latent diffusion models (LDMs)~\cite{stable-diffusion}---on a mixture of diverse restoration tasks (see Fig.~\ref{fig:teaser}). 
The semantic text prompts are processed through the LDM backbone, aligning with LDM's text-to-image pretraining. The degradation text prompts and image-conditioning are implemented in the ControlNet branch, as these aspects are not inherently addressed by the vanilla LDM.
We further improve the ControlNet by introducing a new modulation connection that adaptively fuses the degradation condition with the generative prior with only a few extra trainable parameters yet showing impressive performance gain.

Our extensive experiments demonstrate remarkable restoration quality achieved by our proposed \TIP{} method. Moreover, \TIP{} offers additional freedom for both content and restoration prompt engineering. For example, the semantic prompt \emph{``a very large giraffe eating leaves''} helps to resolve the semantic ambiguity in Fig.~\ref{fig:framework}; the degradation prompt ``deblur with sigma 3.0'' reduces the gaussian blur while maintaining the intentional motion blur in Fig.~\ref{fig:degradation_ambiguity}. We also find out that our model learns a continuous latent space of the restoration strength in~\cref{fig:restoration_decoupling},
even though we do not explicitly fine-tune the CLIP embedding. 
Our contribution can be summarized as follows:
\begin{itemize}[leftmargin=*]

    \item We introduce the first unified text-driven image restoration model that supports both semantic prompts and restoration instructions. Our experiments demonstrate that incorporating semantic prompts and restoration instructions significantly enhances the restoration quality.

    \item Our proposed paradigm empowers users to fully control the semantic outcome of the restored image using different semantic prompts during test time.
    
    \item Our proposed approach provides a mechanism for users to adjust the category and strength of the restoration effect based on their subjective preferences.
    

    \item We demonstrate that text can serve as universal guidance control for low-level image restoration, eliminating the need for task-specific model design.

    




    

\end{itemize}

%% file: sec/2_related.tex
\section{Related Work}
\label{sec:related}

First, we review the literature on image restoration. Our proposed framework aims to address some of the limitations of the existing restoration methods.
Next, we compare multiple text-guided diffusion editing methods. Our method is designed to deal with two different text ambiguities concurrently, which is an unprecedented challenge for existing text-driven diffusion methods.
Finally, we discuss the parameter-efficient fine-tuning of diffusion models, which further motivates our proposed modulation connection.

\noindent\textbf{Image Restoration}
is the task of recovering a high-resolution, clean image from a degraded input. 
Pioneering works in the fields of super-resolution~\cite{dong2016srcnn,lim2017edsr}, motion and defocus deblurring~\cite{su2017deep,abuolaim2021learning,Zamir2021MPRNet}, denoising~\cite{zhang2017dncnn}, and JPEG and artifact removal~\cite{Galteri2017DeepGA,prakash2022interpretable} primarily utilize deep neural network architectures to address specific tasks.
Later, Transformer-based~\cite{liang2021swinir,zhengzhong2022Maxim,zhendong2022Uformer} and adversarial-based~\cite{wang2021real-esrgan} formulations were explored, demonstrating state-of-the-art performance with unified model architecture.
Recently there has been a focus on exploiting iterative restorations, such as the ones from diffusion models, to generate images with much higher perceptual quality~\cite{whang2022deblurring,saharia2022image,saharia2022palette,ren2023multiscale,delbracio2023inversion,SUPIR,wu2024seesr}. 

There have been some recent works attempting to control the enhancement strength of a single model, such as the scaling factor condition in arbitrary super-resolution~\cite{Xiaohang2023Arbitrary_Scale_SR,wang2021scale-arbitrary} and noise-level map for image denoising~\cite{zhang2018ffdnet}. However, these methods require hand-crafting dedicated architectures for each task, which limits their scalability.
In contrast to the task-specific models, other approaches seek to train a single blind model by building a pipeline composed of several classical degradation process, such as BSRGAN~\cite{zhang2021bsrgan} and Real-ESRGAN~\cite{wang2021real-esrgan}.



\input{sec/figs/sec3_method/method_framework}

\noindent\textbf{Text-guided Diffusion Editing.}
Denoising diffusion models are trained to reverse the noising process~\cite{ddpm,ddim,ncsn}. Early methods focused on unconditioned generation~\cite{DiffusionBeatGANs}, but recent trends pivot more to conditioned generation, such as image-to-image translation~\cite{saharia2022palette, saharia2021sr3} and text-to-image generation~\cite{imagen,stable-diffusion}.
%
Latent Diffusion~\cite{stable-diffusion} is a groundbreaking approach that introduces a versatile framework for improving both training and sampling efficiency, while flexible enough for general conditioning inputs.
%
Subsequent works have built upon their pretrained text-to-image checkpoints, and designed customized architectures for different tasks like text-guided image editing.
For instance, SDEdit~\cite{sdedit} generates content for a new prompt by adding noise to the input image. 
Attention-based methods~\cite{pnp,p2p,pix2pix-zero,zhang2024magicbrush} 
show that images can be edited via reweighting and replacing the cross-attention map of different prompts. 
%
Aside from the previously mentioned approaches driven by target prompts, instruction-based editing~\cite{instructpix2pix,Geng23instructdiff} entail modifying a source image based on specific instructions. These textual instructions are typically synthesized using large language models~\cite{radford2019language,openai2023gpt4}.
As CLIP serves as a cornerstone for bridging vision and language, several studies have extensively investigated the representation space of CLIP embeddings~\cite{radford2021clip}.
Instead of directly applying CLIP to fundamental discriminative tasks~\cite{Gu2021OpenvocabularyOD,zhang2021tip-adapter}, these works either tailor CLIP to specific applications, such as image quality assessment~\cite{ke2023vila,liang2023iterative}, serve as semantic guidance in generative adversarial network~\cite{bai2023textir} or aligning image embedding with degradation types in the feature space~\cite{luo2023controlling,jiang2023autodir}. To enhance CLIP's ability to understand numerical information, some works \cite{Paiss2022NoTokenLeftBehind,paiss2023countclip} finetune it using contrastive learning on synthetic numerical data, and then incorporate the fine-tuned text embedding into 
diffusion training.

\vspace{.3em}
\noindent\textbf{Parameter-Efficient Diffusion Model Finetuning.}
To leverage the powerful generative prior in pretrained diffusion models, parameter-efficient components such as text embedding~\cite{gal2022image}, low-rank approximations of model weights~\cite{hu2021lora,Han_2023_ICCV}, and cross attention layers~\cite{kumari2023multi} can be finetuned to personalize the pretrained model.
Adaptor-based finetuning paradigms~\cite{controlnet,t2i-adaptor} propose to keep the original UNet weights frozen, and add new image or text conditions. This adaptor generates residual features that are subsequently added to the frozen UNet backbone.

%% file: sec/figs/sec3_method/method_framework.tex
\begin{figure*}[ht]
  \centering
   \includegraphics[width=0.99\linewidth,]{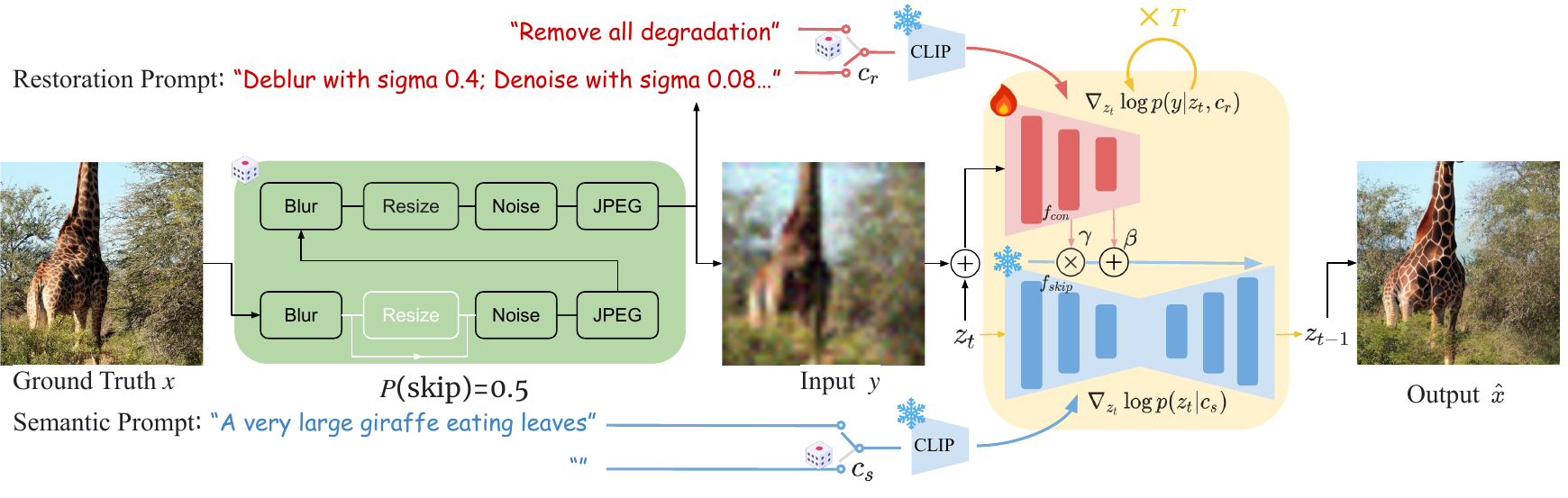}
    \vspace{-1em}
   \caption{\textbf{Framework of \TIP{}}. 
 In the training phase, we begin by synthesizing a degraded version $y$, of a clean image $x$. Our degradation synthesis pipeline also creates a restoration prompt $\bm{c}_r$ , which contains numeric parameters that reflects the intensity of the degradation introduced.
   Then, we inject the synthetic restoration prompt into a ControlNet adaptor, which uses our proposed modulation fusion blocks ($\gamma$, $\beta$) to connect with the frozen backbone driven by the semantic prompt $\bm{c}_s$. During test time, the users can employ the \TIP{} framework as either a blind restoration model with restoration prompt \emph{``Remove all degradation''} and empty semantic prompt $\varnothing$, or manually adjust the restoration $\bm{c}_r$ and semantic prompts $\bm{c}_s$ to obtain what they want.
   }
   \label{fig:framework}
    \vspace{-2em}
\end{figure*}

%% file: sec/3_method.tex
\section{Method}


We introduce a universal approach to combine the above mentioned task-specific, strength-aware, and blind restoration methods within a unified framework (illustrated in Sec.~\ref{ssec:text-driven-ir}). 
We further propose to decouple the learning of content and restoration prompts to better preserve the pre-trained prior while injecting new conditions, as unfolded in Sec.~\ref{ssec:decoupling}.
Sec.~\ref{ssec:learning-to-control-restoration} details our design on how to accurately control the restoration type and strength, as well as our proposed modulation fusion layer that adaptively fuses the restoration features back to the frozen backbone.
We start with some preliminaries in the following section.



\subsection{Preliminaries}
\label{secPreliminary}
\newcommand{\y}{\boldsymbol{y}}
\newcommand{\cs}{\boldsymbol{c}_s}
\newcommand{\cres}{\boldsymbol{c}_r}
\newcommand{\ztlatents}{\boldsymbol{z}_t}

Latent Diffusion Models (LDMs)~\cite{stable-diffusion} are probabilistic generative models that learn the underlying data distribution by 
iteratively removing Gaussian noise in the latent space, which is typically learned using a VAE autoencoder. Formally, the VAE encoder $\mathcal{E}$ compresses an input image $\boldsymbol{x}$ into a compact latent representation $\boldsymbol{z}=\mathcal{E}(\boldsymbol{x})$, which can be later decoded back to the pixel space using the coupled VAE decoder $\mathcal{D}$, often learned under an image reconstruction objective: $\mathcal{D}(\mathcal{E}(\boldsymbol{x}))\approx \boldsymbol{x}$.
During training stage, the output of a UNet~\cite{unet} $\boldsymbol\epsilon_{\theta}\left(\boldsymbol{z}_t, t, \boldsymbol{y} \right)$ conditioned on ${\bm{y}}$ (such as text, images, etc.) is parameterized~\cite{DiffusionBeatGANs,progressivedistillation} to remove Gaussian noise $\boldsymbol{\epsilon}$ in the latent space $\boldsymbol{z}_t$ as: 
%
\begin{equation} \label{eq:diffusion}
\min _\theta \mathbb{E}_{\substack{(\boldsymbol{z}_0,\boldsymbol{y}) \sim p_\text{data}, \boldsymbol{\epsilon} \sim \mathcal{N}(0,I), t }}\left\|\boldsymbol{\epsilon}-\boldsymbol{\epsilon}_\theta\left(\boldsymbol{z}_t, t, \bm{y} \right)\right\|_2^2,
\end{equation}
where $\boldsymbol{z}_t$ is a noisy sample of $\boldsymbol{z}_0$ at sampled timestep $t$. 
The condition $\bm{y}$ is randomly dropped out as $\varnothing$ to make the model unconditional.

At test time, deterministic DDIM sampling~\cite{ddim} is utilized to convert a random noise $z_T$ to a clean latent $z_0$ which is decoded as final result $\mathcal{D}(z_0)$.
In each timestep $t$, classifier-free guidance~\cite{DiffusionBeatGANs,ho2022classifierfree} can be applied to trade-off sample quality and condition alignment:
\begin{eqnarray*}
\overline{\boldsymbol{\epsilon}}_\theta(\boldsymbol{z}_t, t, \boldsymbol{y}) = \boldsymbol{\epsilon}_\theta(\boldsymbol{z}_t, t, \varnothing) + w(\boldsymbol{\epsilon}_\theta(\boldsymbol{z}_t, t, \boldsymbol{y})-\boldsymbol{\epsilon}_\theta(\boldsymbol{z}_t, t, \varnothing)),
\end{eqnarray*}
where $w$ is a scalar to adjust the guidance strength of $y$. Note that the estimated noise $\overline{\boldsymbol{\epsilon}}_\theta(\boldsymbol{z}_t, t, \boldsymbol{y})$ is used to update $z_{t}$ to $z_{t-1}$, which is a approximation of the distribution gradient score~\cite{song2021scorebased} as $\nabla_{z_t}\log p(\boldsymbol{z}_t |\y{}) \propto \overline{\boldsymbol{\epsilon}}_\theta(\boldsymbol{z}_t, t, \boldsymbol{y})$.

\subsection{Text-driven Image Restoration}
\label{ssec:text-driven-ir}


Based on the LDM framework, we propose a new restoration paradigm---text-driven image restoration. 
Our method target to restore images ($\boldsymbol{x} $ or $ \boldsymbol{z}_0$) based on conditions $ \{\boldsymbol{y}, \boldsymbol{c}_s, \boldsymbol{c}_r\}$. Specifically: $\boldsymbol{y}$ denotes the degraded image condition, $\cs{}$ is the \underline{\textbf{s}}emantic prompt describing the clean image $\boldsymbol{x}$ (\eg, ``a panda is sitting by the bamboo'' or ``a panda''), and $\boldsymbol{c}_r$ is the \underline{\textbf{r}}estoration prompt that describes the details of the degradation in terms of both the operation and parameter (e.g., ``deblur with sigma 3.0''). We use $\boldsymbol{y}=\text{Deg}(\boldsymbol{x}, \cres{})$ to denote the degradation process which turns the clean image $\boldsymbol{x}$ into its degraded counterpart $\boldsymbol{y}_i$. 

The above text-driven image restoration model $p(\boldsymbol{z}_t|\{\boldsymbol{y}, \cs{}, \cres{} \})$ can be trained using paired data. We use the text-image dataset~\cite{PaLI} so that each clean image $\boldsymbol{x}$ is paired with a semantic prompt $\cs{}$ (the paired alt-text). Then $\bm{y}=\text{Deg}(\bm{x}, \cres{})$ is simulated using the synthetic degradation~\cite{wang2021real-esrgan} pipeline (\cref{sec:Training_data_Benchmarks}), yielding the final paired training data $(\boldsymbol{x} $ or $\boldsymbol{z}_0, \{\boldsymbol{y}, \cs{} , \cres{} \})\sim p_{data}$.

\vspace{.3em}
\noindent\textbf{Comparison to Blind Restoration.}
Recent existing blind restoration models~\cite{wang2023stablesr,Lin2023diffbir,saharia2021sr3} also leverage diffusion priors to generate high-quality images. However, these methods are prone to hallucinate unwanted, unnatural, or oversharpened textures given that the semantic- and degradation-ambiguities persist everywhere.
\cref{fig:degradation_ambiguity} provides an example, where the input image is degraded simultaneously with Gaussian blur and motion blur. The fully blind restoration method removes all of the degradation, leaving no way to steer the model to remove only the Gaussian blur while preserving the motion effect for aesthetic purposes.
Therefore, having the flexibility to control the restoration behavior and its strength at test time is a crucial requirement.
Our text-driven formulation of the problem introduces additional controlling ability using both the semantic content and the restoration instruction. In our \TIP{} design, $\boldsymbol{c}_r=$\emph{``Remove all degradation''} is also a plausible restoration prompt which makes our \TIP{} to be compatible with the blind restoration .
Further, our text-driven restoration takes advantages of the pretrained content-aware LDM using the additional semantic prompt $\cs{}$, hence able to better handle semantic ambiguities on the noisy, blurry regions whose visual objects are less recognizable.

\subsection{Decoupling Semantic and Restoration Prompts}
\label{ssec:decoupling}

To effectively learn the latent distribution $p(\bm{z}_t|\y{}, \cs{}, \cres{})$, we further decouple the conditions $\{\y{}, \cs{}, \cres{}\}$ into two groups---one for the text-to-image prior ($\cs{} \rightarrow \bm{z}_t$) already imbued in the pretrained LDM model, and the other for both the image-to-image ($\y{} \rightarrow \bm{z}_t$) and restoration-to-image ($\cres{} \rightarrow \bm{z}_t$) that needs to be learnt from the synthetic data. This decoupling strategy prevents catastrophic forgetting in the pretrained diffusion model and enables independent training of the restoration-aware model
, whose
score function~\cite{song2021scorebased} is expressed as:
%
\definecolor{cvprblue}{rgb}{0.21,0.49,0.74}
\vspace{-1em}
\begin{equation}
\begin{split}
    &\nabla_{z_t}{\log p(\boldsymbol{z}_t |\bm{y}, \cs{}, \cres{})}
    \approx
   \underbrace{\nabla_{z_t}\log p(\boldsymbol{z}_t |\cs{})}_{\text{Semantic-aware (\textcolor{cvprblue}{frozen})}} + \underbrace{\nabla_{z_t}\log p(  \y{}| \boldsymbol{z}_t,\cres{}).}_{\text{Restoration-aware (\textcolor{BrickRed}{learnable})}}
\label{eq:text_driven_restoration}
\end{split}
\end{equation}


In the above equation, the first part $\nabla_{\boldsymbol{z}_t} \log p(\boldsymbol{z}_t | \cs{})$ aligns with the text-to-image prior inherent in the pretrained LDM.
The second term $\nabla_{\boldsymbol{z}_t} \log p(\y{}| \boldsymbol{z}_t, \cres{})$ approximates the consistency constraint with degraded image $\y{}$, meaning the latent image $\bm{z}_t$ should be inferred by the degraded image $\y{}$ and the degradation details $\cres{}$. This is somewhat similar to the reverse process of $\y{}=\text{Deg}(\bm{x}_t,\cres{})$ in the pixel space. More derivation is provided in our supplement.
Providing degradation information through a restoration text prompt $\cres{}$ can largely reduce the uncertainly of estimation $p(\y{}| \boldsymbol{z}_t, \cres{})$,
thereby alleviating the ambiguity of the restoration task and leading to better results that are less prone to hallucinations.

\vspace{-2em}
\input{sec/figs/sec3_method/content_degradation_ambiguity}




\vspace{1em}
\subsection{Learning to Control the Restoration.}
\label{ssec:learning-to-control-restoration}


Inspired by ControlNet~\cite{controlnet}, we employ an adaptor to model the restoration-aware branch in Eq.~\eqref{eq:text_driven_restoration}.
First, the pretrained latent diffusion UNet $\boldsymbol\epsilon_{\theta}\left(\boldsymbol{z}_t, t, \cs{} \right)$ is frozen during training to preserve text-to-image prior $\nabla_{\boldsymbol{z}_t} \log p(\boldsymbol{z}_t | \cs{})$.
Secondly, an encoder initialized from pretrained UNet is finetuned to learn the restoration condition $\nabla_{z_t}\log p( \y{}| \boldsymbol{z}_t,\cres{})$, which takes degradation image $\y{}$, current noisy latents $\ztlatents{}$ and the restoration prompt $\cres{}$ as input. The output of encoder is fusion features $\bm{f}_{con}$ for frozen UNet.
In this way, the choice of ControlNet implementation follows our decoupling analysis in~\cref{eq:text_driven_restoration}.

\noindent\textbf{Conditioning on the Degraded Image $\y{}$.}
We apply downsample layers on $\y{}$ to make the feature map matching to the shape of the $\bm{z}_t$, denoted as $\mathcal{E}'(\y{})$. $\mathcal{E}'$ is more compact~\cite{controlnet} than VAE encoder $\mathcal{E}$, and its trainable parameters allow model to perceive and adapt to the degraded images from our synthetic data pipeline. Then, we concatenate the downsampled feature map $\mathcal{E}'(\y{})$ with the latent $\bm{z}_t$, and feed them to the ControlNet encoder.

\noindent\textbf{Conditioning on the Restoration Prompt $\cres{}$.}
Following~\cite{stable-diffusion}, we first use the vision-language CLIP~\cite{radford2021clip} to infer the text sequence embedding $\bm{e}_r \in \mathbb{R}^{M\times d}$ for $\cres{}$, where $M$ is the number of tokens, and $d=768$ is the dimension of the image-text features.
Specifically, given a restoration instruction ``\emph{deblur with sigma 3.0}'', CLIP first tokenizes it into a sequence of tokens
(e.g., ``\emph{3.0}" is tokenized into [``3'', ``.'', ``0'']) and embeds them into $M$ distinct token embeddings.
Then, stacked causal transformer layers~\cite{attention_is_all_you_need} are applied on the embedding sequence to contextualize the token embeddings, resulting in $\bm{e}_r$. Hence, tokens such as ``\emph{deblur}'' may arm with the numeric information such as ``\emph{3}'', ``\emph{.}'',, and ``\emph{0}'', making it possible for the model to distinguish the semantic difference of strength definition in different restoration tasks.
Cross-attention process ensures that the information from $\bm{e}_r$ is propagated to the ControlNet feature $f_{con}$.
While \cite{Paiss2022NoTokenLeftBehind,paiss2023countclip} finetuned CLIP using contrastive learning, while we found frozen CLIP still works. The reason is that the learnable cross-attention layers can somehow ensemble the observations from frozen CLIP and squeeze useful information from it.

\noindent\textbf{Modulation Fusion Layer.} 
The decoupling of learning two branches (Sec.~\ref{ssec:decoupling}), despite offering the benefit of effective separation of prior and conditioning, may cause distribution shift in the learned representations between the frozen backbone feature $\bm{f}_\mathrm{skip}$ and ControlNet feature $\bm{f}_\mathrm{control}$.
To allow for adaptive alignment of the above features, here we propose a new modulation fusion layer to fuse the multi-scale features from ControlNet to the skip connection features of the frozen backbone:
$
\hat{\bm{f}}_{\mathrm{skip}}=(1+\boldsymbol{\gamma}) \bm{f}_{\mathrm{skip}}+\boldsymbol{\beta}; ~ \boldsymbol{\gamma}, \boldsymbol{\beta}=\mathcal{M}({\bm{f}_{\mathrm{con}}}),
$
where $\bm{\gamma}, \bm{\beta}$ are the scaling and bias parameters from $\mathcal{M}$ a lightweight $1\times1$ zero-initialized convolution layer.

\input{sec/tabs/main_table}
\input{sec/tabs/degradation_pipeline}

%% file: sec/figs/sec3_method/content_degradation_ambiguity.tex



\begin{wrapfigure}{r}{0.6\textwidth}
  \begin{center}
      \vspace{-2.5em}
    \includegraphics[width=0.6\textwidth]{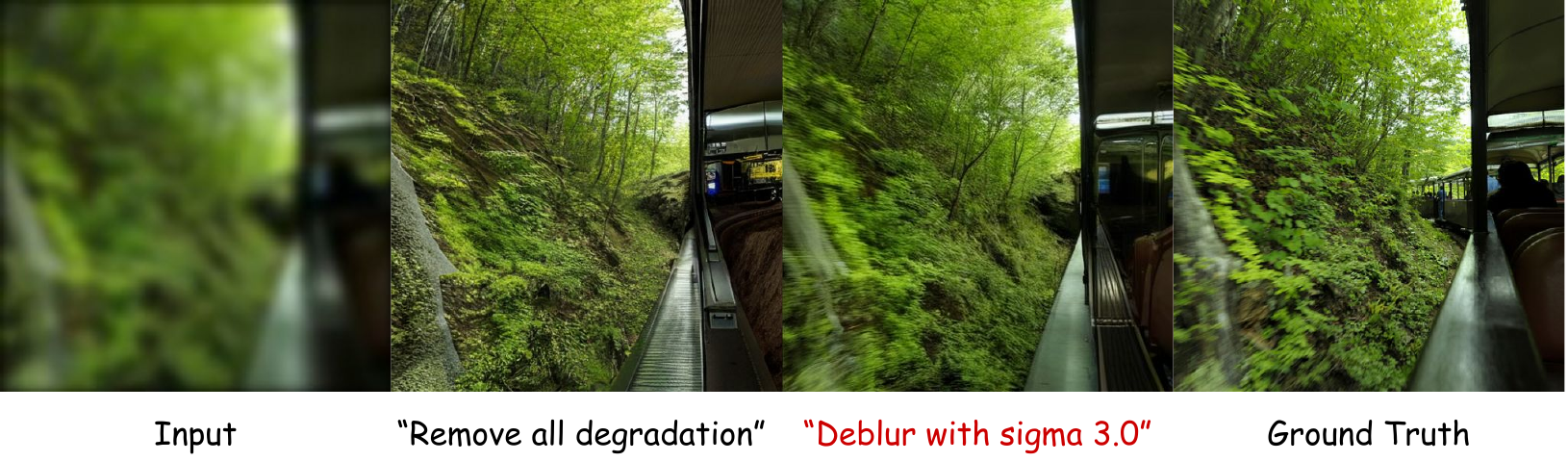}
    \vspace{-2.5em}
  \end{center}
  \caption{\textbf{Degradation ambiguities in real-world problems}. By adjusting the restoration prompt, our method can preserve the motion effect that is coupled with Gaussian blur, while fully blind restoration models do not provide this 
  flexibility.}
 \label{fig:degradation_ambiguity}
    \vspace{-1.5em}
\end{wrapfigure}

%% file: sec/tabs/main_table.tex
\begin{table*}[!t]
\centering
\setlength{\tabcolsep}{1.6pt}
\renewcommand{\arraystretch}{1.0}
\resizebox{1\linewidth}{!}{
\begin{tabular} 
{lcccccccccccccr}

\toprule
 & \multicolumn{2}{l}{Prompts} & \multicolumn{6}{c}{Parameterized Degradation with synthesized $\cres{}$} & \multicolumn{6}{c}{Real-ESRGAN Degradation without $\cres{}$ }
\\
\cmidrule(l{0mm}r{1mm}){2-3} 
\cmidrule(l{1mm}r{1mm}){4-9}
\cmidrule(l{1mm}r{0mm}){10-15}
Method & Sem & Res  & FID$\downarrow$ & LPIPS$\downarrow$ & PSNR$\uparrow$ & SSIM$\uparrow$ &
\small{CLIP-I}$\uparrow$ & \small{CLIP-T}$\uparrow$ & 
FID$\downarrow$ & LPIPS$\downarrow$ & PSNR$\uparrow$ & SSIM$\uparrow$ & 
\small{CLIP-I}$\uparrow$ & \small{CLIP-T}$\uparrow$ \\
\midrule
SwinIR~\cite{liang2021swinir} & \textcolor{BrickRed}{\XSolidBrush} & \textcolor{BrickRed}{\XSolidBrush} & 43.22 & 0.423 & \textbf{24.40} & \textbf{0.717} & 0.856 & 0.285 & 48.37 & 0.449 & \textbf{23.45} & \textbf{0.699} & 0.842 & 0.284\\
StableSR~\cite{wang2023stablesr} & \textcolor{BrickRed}{\XSolidBrush} & \textcolor{BrickRed}{\XSolidBrush} &
20.55 & 0.313 & 21.03 & 0.613 & 0.886 & 0.298 &
25.75 & 0.364 & 20.42 & 0.581 & 0.864 & 0.298
\\

DiffBIR~\cite{Lin2023diffbir} & \textcolor{BrickRed}{\XSolidBrush} & \textcolor{BrickRed}{\XSolidBrush} &
17.26 & 0.302 & 22.16 & 0.604 & 0.912 & 0.297 & 19.17 & 0.330 & 21.48 & 0.587 & 0.898 & 0.298 \\

ControlNet-SR~\cite{controlnet} & \textcolor{BrickRed}{\XSolidBrush} & \textcolor{BrickRed}{\XSolidBrush} &
13.65 & 0.222 & 23.75 & 0.669 & 0.938 & 0.300 & 16.99 & 0.269 & 22.95 & 0.628 & 0.924 & 0.299\\
Ours w/o text & \textcolor{BrickRed}{\XSolidBrush} & \textcolor{BrickRed}{\XSolidBrush} & 
12.70 & 0.221 & {23.84} & 0.671 & 0.939 & 0.299 & 16.25 & \textbf{0.262} & {23.15} & {0.636} & 0.929 & 0.300\\
\hline

DiffBIR~\cite{Lin2023diffbir} + SDEdit~\cite{sdedit} & \textcolor{teal}{\Checkmark} & \textcolor{BrickRed}{\XSolidBrush} & 19.36 & 0.362 & 19.39 & 0.527 & 0.891 & 0.305 & 17.51 & 0.375 & 19.15 & 0.521 & 0.887 & \textbf{0.308} \\
DiffBIR~\cite{Lin2023diffbir} + CLIP~\cite{radford2021clip}& \textcolor{teal}{\Checkmark} & \textcolor{BrickRed}{\XSolidBrush} & 18.46 & 0.365 & 20.50 & 0.526 & 0.896 & \textbf{0.308} & 20.31 & 0.374 & 20.45 & 0.539 & 0.885 & 0.307
\\
ControlNet-SR + CLIP~\cite{radford2021clip}& \textcolor{teal}{\Checkmark} & \textcolor{BrickRed}{\XSolidBrush} & 13.00 & 0.241 & 23.18 & 0.648 & 0.937 & 0.307 & 15.16 & 0.286 & 22.45 & 0.610 & 0.926 & \textbf{0.308}
\\
\hline

Ours & \textcolor{teal}{\Checkmark} & \textcolor{teal}{\Checkmark} & \textbf{11.34} & \textbf{0.219} & 23.61 & 0.665 & \textbf{0.943} & 0.306 & \textbf{14.42} & \textbf{0.262} & 23.14 & 0.633 & \textbf{ 0.935} & \textbf{0.308} \\
\bottomrule
\end{tabular}
}
\caption{Quantitative results on the MS-COCO dataset (with $c_s$) using our parameterized degradation (left) and Real-ESRGAN degradation (right). We also denote the prompt choice at test time. `Sem' stands for semantic prompt; `Res' stands for restoration prompt. The first group of baselines are tested without prompt. The second group are combined with semantic prompt in zero-shot way.
}
\label{tab:merged main-table}
\vspace{-1em}
\end{table*}

%% file: sec/tabs/degradation_pipeline.tex
\vspace{-1em}
\begin{table}[t]
\centering
\setlength{\tabcolsep}{1.6pt}
\renewcommand{\arraystretch}{1.0}
\resizebox{1.0\linewidth}{!}{
\begin{tabular}{l@{\hspace{3mm}}cr|l@{\hspace{3mm}}cr}
\toprule
\multicolumn{3}{c|}{Parameterized} & \multicolumn{3}{c}{Real-ESRGAN} \\
Degradation Process & $p(\text{choose})$ & Restoration Prompt & Degradation Process & $p(\text{choose})$ & Restoration Prompt \\
\midrule
Gaussian Blur & 0.5 & Deblur with \{sigma$\in [0.2, 3.0]$\} or Deblur & Blur          & 1.0 & \multirow{4}{*}{Remove all degradation} \\
Downsample    & 0.5 & Upsample to \{resizing factor$\in [1.0, 7.0]$\} or Upsample & Resize        & 1.0 &  \\
Gaussian Noise & 0.5 & Denoise with \{sigma$\in [0.0, 0.12]$\} or Denoise & Noise         & 1.0 &  \\
JPEG          & 0.5 & Dejpeg with quality \{quality factor$\in [30, 92]$\} or Dejpeg & JPEG          & 1.0 &  \\
\bottomrule
\end{tabular}
}
\caption{\textbf{Our training degradation} is randomly sampled in these two pipeline with $50\%$ each. 
(1) Images generated by our \textbf{parameterized pipeline} are paired with either a restoration type 
(\eg,\emph{``Deblur’’}) 
or a restoration parameter prompt
(\eg,\emph{``Deblur with sigma 0.3;''}). 
(2) In other $50\%$ iterations, degraded images $\y{}$ synthesized by Real-ESRGAN are paired with the same restoration prompt $\cres{}=$ \emph{``Remove all degradation''}
}

\label{tab:Degradation_pipeline}
\vspace{-2em}
\end{table}

%% file: sec/4_exp.tex
\section{Experiments}
\label{sec:result}

\subsection{Text-based Training Data and Benchmarks}
\label{sec:Training_data_Benchmarks}
Our method opens up an entirely new paradigm of instruction-based image restoration. However, existing 
image restoration datasets such as DIV2K~\cite{div2k} and Flickr2K~\cite{flicker2k} do not provide high-quality semantic prompts, and Real-ESRGAN degradation is for blind restoration. To address this, we construct a new setting including training data generation and test benchmarks for
evaluation.

\noindent\textbf{Our parameterized degradation pipeline} is based on the Real-ESRGAN~\cite{wang2021real-esrgan} which contains a large number of degradation types. 
Since the original full Real-ESRGAN degradation is difficult to be parameterized and represented as user-friendly natural language (\eg, not every one understands \emph{``anisotropic plateau blur with beta 2.0"}), we choose the 4 most general degradations which are 
practical for parameterization to
support degradation parameter-aware restoration, 
as shown in~\cref{tab:Degradation_pipeline}. Our parameterized pipeline skips each degradation stage with a $50\%$ probability to increase the diversity of restoration prompts and support single task restoration. The descriptions of the selected degradations are appended following the order of the image degradations to synthesize $\cres{}$.
In real application, users can control the restoration type and its strength by modifying restoration prompt. As shown in third column of~\cref{fig:degradation_ambiguity}, driven by restoration prompt \emph{"Deblur with sigma 3.0"}, our framework removes the Gaussian blur, while preserving motion blur.
The representation of the restoration strength is also general enough to decouple different tasks in natural language as shown in~\cref{fig:restoration_decoupling}.
More visual results are presented in our supplement.

\vspace{.3em}
\noindent\textbf{Training data construction.} 
In the training stage, we sample near 100 million text-image pairs from 
internal data source
and use the alternative text label with highest relevance as the semantic prompt $\boldsymbol{c}_s$ in the framework. We drop out the semantic prompt to empty string $\varnothing$ by a probability of $10\% $ to support semantic-agnostic restoration. We mix the Real-ESRGAN degradation pipeline and our parameterized pipeline by $50\%$ each in training iterations.

\input{sec/figs/sec4_exp/main_visuals_merge}

\vspace{.3em}
\noindent\textbf{Evaluation Setting.}
We randomly choose 3000 pairs of image $\bm{x}$ and semantic prompt $\bm{c}_s$ from the MS-COCO~\cite{lin2014mscoco} validation set. Then, we build two test sets on parameterized and the Real-ESRGAN processes separately. 
At the left half of~\cref{tab:merged main-table}, images are sent to the 
parameterized degradation
to form the synthesize degradation image $\bm{y}$ with restoration prompt $ \bm{c_r}$. In this setting, we expect ``Ours'' model to fully utilize the $\bm{c_s}$ describing the semantic and the $\bm{c_r}$ describing the degradations.
On the right half, the same images are fed to the 
Real-ESRGAN degradation---so the degradation is aligned with open-sourced checkpoints~\cite{liang2021swinir,Lin2023diffbir,wang2023stablesr}.
Because not every model can utilize both the semantic ($\bm{c}_s$) and restoration ($\bm{c}_r$) information, 
~\cref{tab:merged main-table} provides a ``Prompts'' column denoting what information the model relied on.
In summary, our constructed benchmarks based on MS-COCO can serve multiple purposes---from fully-blind, 
content-prompted, to all-prompted restoration.

\vspace{.3em}
\noindent\textbf{Evaluation Metrics.}
We use the generative metric FID~\cite{FID} and perceptual metric LPIPS~\cite{zhang2018perceptual} for quantitative evaluation of image quality.
PSNR, and SSIM are reported for reference. 
We also evaluate similarity scores in the clip embedding space with ground truth image (CLIP-I) and caption (CLIP-T). 


\input{sec/tabs/div2k_eval}

\subsection{Comparing with baselines}
We compare our method with three categories of baselines: 
\begin{itemize}[leftmargin=*]
    \item Open-sourced image restoration checkpoints, including the regression model SwinIR~\cite{liang2021swinir} and the latent diffusion models StableSR~\cite{wang2023stablesr} and DiffBIR~\cite{Lin2023diffbir}. \vspace{.2em}
    
    \item Combining the state-of-the-art method~\cite{Lin2023diffbir} with the zero-shot post-editing~\cite{sdedit} (DiffBIR+SDEdit) or zero-shot injection through CLIP~\cite{radford2021clip} (DiffBIR + CLIP).\vspace{.2em}
    
    \item ControlNet-SR (retrained): We adapt the image-to-image ControlNet model~\cite{controlnet}  for super-resolution. For consistency, we maintain the same training iterations as our own model. We also evaluate its zero-shot capabilities with CLIP guidance (ControlNet-SR + CLIP). 
    
\end{itemize}

\noindent\textbf{Quantitative comparison with baselines}
is presented in~\cref{tab:merged main-table}. On the left, we evaluate our full model with the parameterized degradation test set. 
Since open-sourced baselines are pretrained on Real-ESRGAN degradation, not perfectly aligned with our parameterized degradation, we also evaluate our full model on MS-COCO with Real-ESRGAN degradation by setting our restoration prompt to \emph{``Remove all degradation''}.
Thanks to our prompts guidance and architecture improvements,
our full model achieves best FID, LPIPS, CLIP-Image score, 
which means better image quality and semantic restoration.
``Ours w/o text'' is the same checkpoint conditioned on only degradation image and has high pixel-level similarity (the second best PSNR and SSIM) with GT. Although SwinIR has highest PSNR, its visual results are blurry.
Although combining semantic prompt in zero-shot way can bring marginal improvement in FID and CLIP similarity with caption, it deteriorates the image restoration capacity and results in worse LPIPS and CLIP-Image. In contrast, semantic prompt guidance improves the CLIP image similarity of our full model. 
~\cref{tab:other_testset} shows the evaluation results of our model on DIV2K test set provided by StableSR. 
In zero-shot test setting, our model
has lower FID and LPIPS than DiffBIR.
To fairly compare with StableSR, 
we finetune our model 
on their training set.
Our finetuned model also outperforms StableSR in FID and CLIP-I.

\vspace{.3em}
\noindent\textbf{Qualitative comparison with baselines} is presented in~\cref{fig:main-visuals}, where the corresponding semantic prompt is provided below each row. Image-to-image baselines such as DiffBir and the retrained ControlNet-SR can easily generate artifacts (\eg, hydrant in the first row) and blurry images (\eg, hands example in third row) that may look close to an ``average'' estimation 
. Besides, naive combination with semantic prompt in the zero-shot approach results in heavy hallucinations and may fail to preserve the object identity in the input image 
(\eg, giraffes on the second row). 
Unlike the existing methods, our full model considers semantic prompt, degradation image and restoration prompt in both training and test stages, which makes its results more aligned with all conditions.

\subsection{Real-world restoration}
\input{sec/figs/sec4_exp/real-world-example}

\input{sec/tabs/real-world}

Besides evaluation on synthetic degradations, we present
an analysis on real-world images in 
~\cref{tab:real-world-result}
and 
~\cref{fig:real-world restoration}
. We choose RealPhoto60~\cite{SUPIR} as our real-world test-set,
and use non-reference metrics CLIP-IQA~\cite{wang2022clipiqa}, MUSIQ~\cite{ke2021musiq} and MANIQA~\cite{yang2022maniqa} scores
to study the influence of the semantic and restoration prompts.
First, we fix the semantic prompt to the empty string $\varnothing$ and conduct an ablation study on the restoration prompt. As the \textit{deblurring} strength increases from \textit{0.6} to \textit{2.0}, the image quality is improved consistently. A blind prompt \textit{"Remove all degradation"} can provide better results than the manually controlled restoration prompts. This showcases that the proposed model is flexible to work with generic restoration prompts, especially when coming up with the perfect prompt is challenging. We also fix the blind restoration prompt and change the semantic prompts.
Following ~\cite{SUPIR}, we feed degraded images to LLaVA~\cite{liu2023llava} and generate corresponding semantic prompts. 
Our experiments show that detailed \textcolor{teal}{full} descriptions (about 60 words) of the degraded image from LLaVA can be more effective to improve the visual results than
\textcolor{teal}{short} prompts that have less than 20 words.  
\textcolor{teal}{Unmatched} prompts simulate the case when a user provides wrong prompts by randomly re-sampling unmatched long prompts from the data set.
An interesting observation is that unmatched long prompts also improve the aesthetic quality of images against a short prompt or a blind restoration. Nonetheless, it can lead to more ``hallucination'' when the semantic input prompt is unmatched with the real ground truth.
In contrast to concurrent work~\cite{sun2023coser} that leverages implicit clip embedding, our text-based formulation provides better control of hallucinations by adjusting the LLM-synthesized or manually-designed prompt at test-time.
Note that many studies~\cite{blau2018pd_tradeoff,NTIRE2022_IQA} find current image restoration metrics are biased towards specific datasets and not fully aligned with human's visual perception. The quantitative results of previous works~\cite{wang2021real-esrgan,Lin2023diffbir} are provided for reference. More results and discussions are provided in the supplements.

\input{sec/figs/sec4_exp/content_prompt}

\input{sec/figs/sec4_exp/degradation_prompt}

\subsection{Prompting the \TIP{}}

\noindent\textbf{Semantic Prompting.}
Our model can be driven by user-defined semantic prompts.
As shown in~\cref{fig:content_prompting}, the blind restoration with empty string generates ambiguous and unrealistic identity. In contrast, our framework reconstructs sharp and realistic results.
The object identity accurately follows user's semantic prompt,
while the global layout and color tone remain consistent with input.


\noindent\textbf{Restoration Prompting.}
Users also have the freedom to adjust the degradation types and strengths in the restoration prompts.
As shown in~\cref{fig:restoration_decoupling}, the denoising and deblurring restoration are well decoupled. Using the task-specific prompts, our method can get a clean but blur (second row), or sharp but noisy (third row) result. In addition, our model learns the continuous space of restoration as the result becomes cleaner progressively if we modify the ``Denoise with sigma 0.06'' to ``0.24.''. 
Beyond purely relying on numerical description, our model ``understand'' the difference of strength definition for tasks, such as the range difference of denoise sigma and deblur sigma, which makes it promising to be a universal representation for restoration strength control.



\subsection{Ablation Study}
\input{sec/tabs/ablation}


We conduct several ablations in~\cref{tab:ablation_study}. 
Providing both prompts $c_s$, $c_r$ 
to \newline ``ControlNet-SR'' baseline improves the FID by 1.5. 
Adding the modulation fusion layer (``Ours'') further improves the generative quality (0.8 for FID) and pixel-level similarity (0.11 dB).
Embedding degradation parameters in the restoration prompt, not only enables restoration strength prompting (\cref{fig:restoration_decoupling}), but also improves image quality (FID and LPIPS). More details are in the supplement.














%% file: sec/figs/sec4_exp/main_visuals_merge.tex



\newcommand{\imgwidth}{0.26\columnwidth}

\begin{figure*}[t]
\centering
\setlength\tabcolsep{1pt}
{
\renewcommand{\arraystretch}{0.6}
\resizebox{1\linewidth}{!}{
\begin{tabular}{@{}*{7}{c}@{}}
     Input & DiffBIR~\cite{Lin2023diffbir} & DiffBIR + SDEdit~\cite{sdedit} & DiffBIR + CLIP~\cite{radford2021clip} &
     \small{ControlNet-SR~\cite{controlnet}}
     & Our \TIP{} Model & Ground-Truth \\
     \includegraphics[width=\imgwidth]{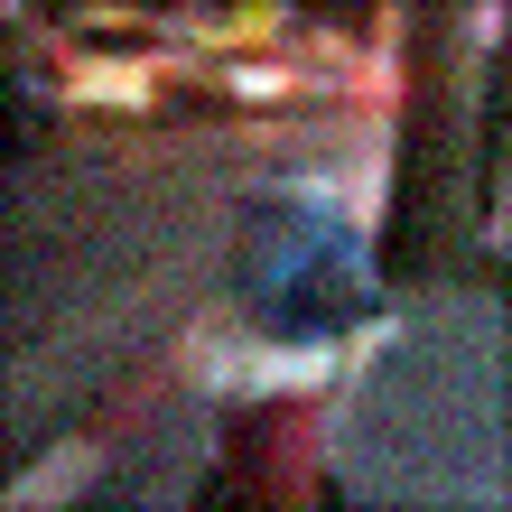} & 
     \includegraphics[width=\imgwidth]{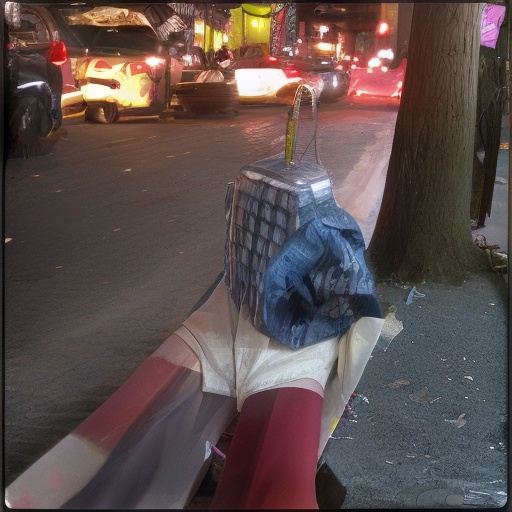} &
     \includegraphics[width=\imgwidth]{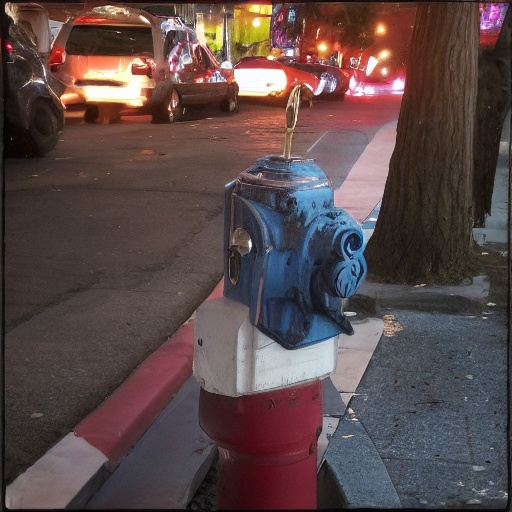} & 
     \includegraphics[width=\imgwidth]{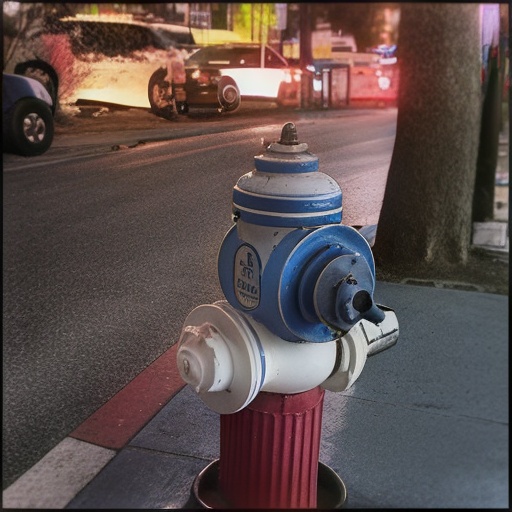} & 
     \includegraphics[width=\imgwidth]{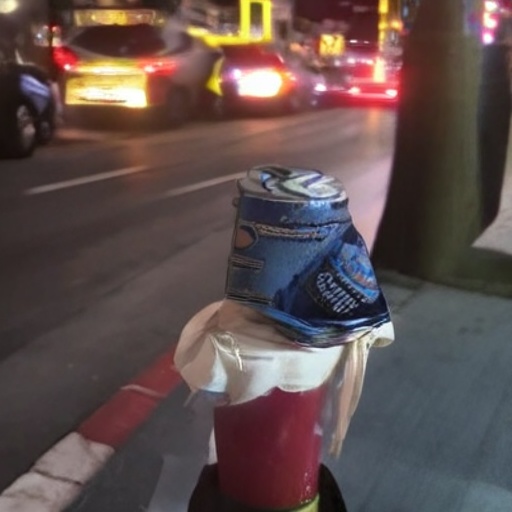} 
     & \includegraphics[width=\imgwidth]{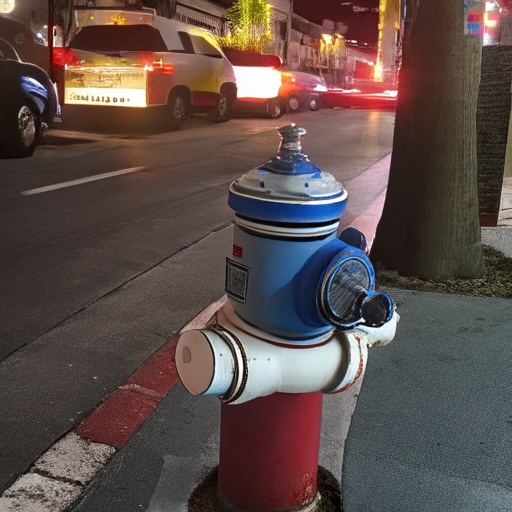} & \includegraphics[width=\imgwidth]{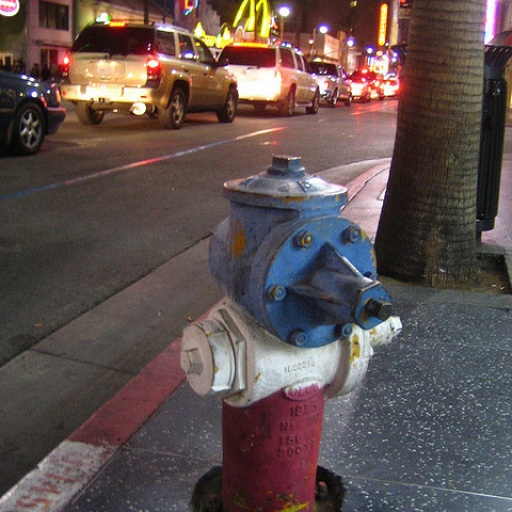}\\
     \multicolumn{7}{c}{$\boldsymbol{c}_s=$\emph{``A blue, white and red fire hydrant sitting on a sidewalk.''}} \\
     \includegraphics[width=\imgwidth]{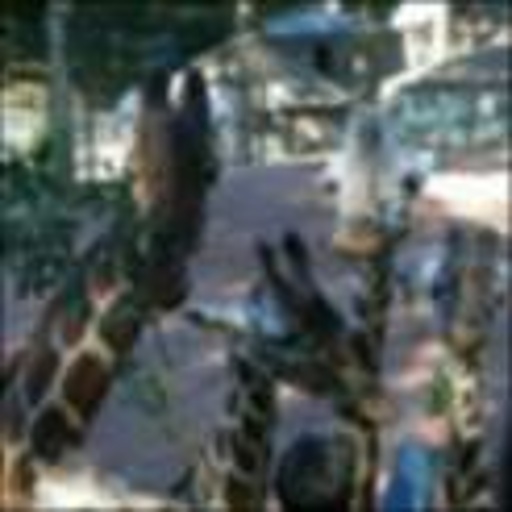} &
     \includegraphics[width=\imgwidth]{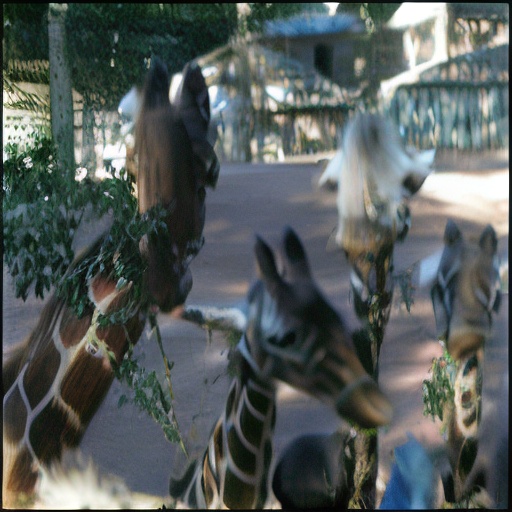} & 
     \includegraphics[width=\imgwidth]{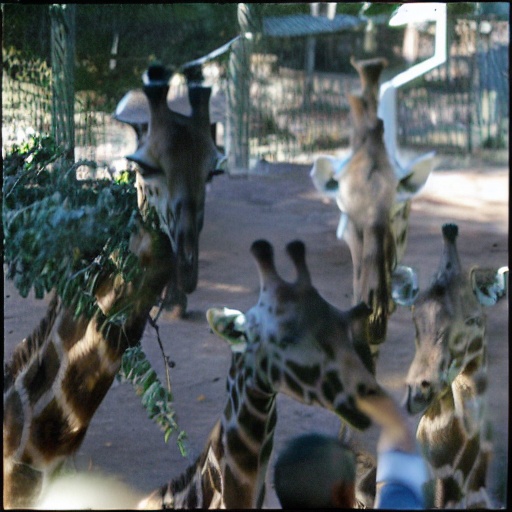} &
      \includegraphics[width=\imgwidth]{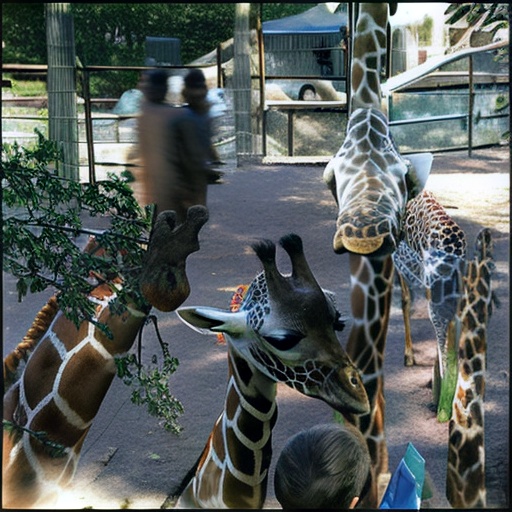} &
     \includegraphics[width=\imgwidth]{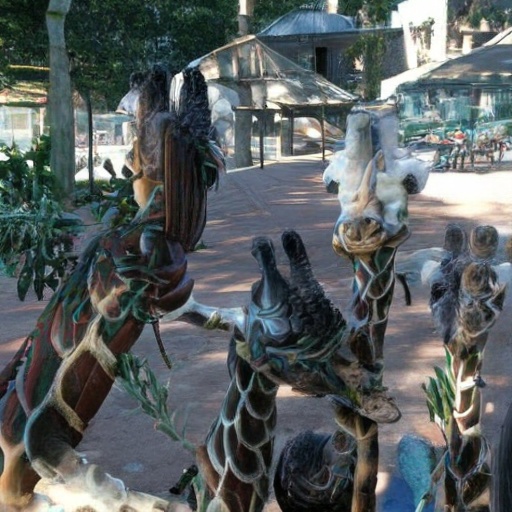} & 
     \includegraphics[width=\imgwidth]{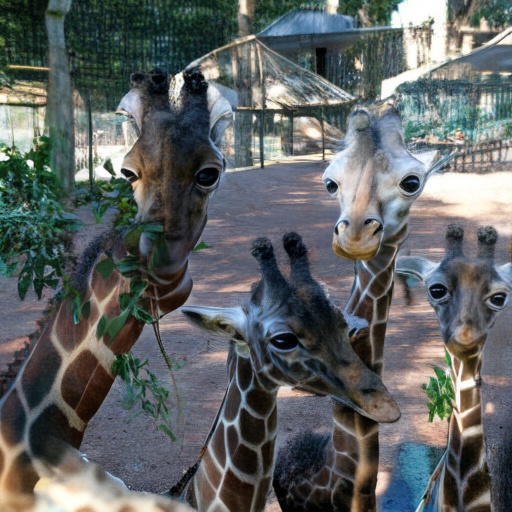} 
     & \includegraphics[width=\imgwidth]{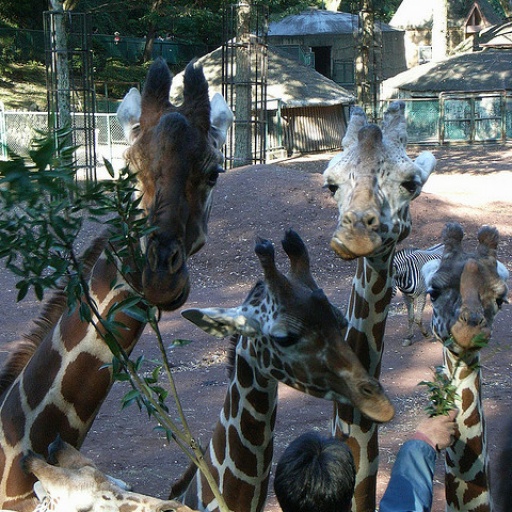}\\
     \multicolumn{7}{c}{$\boldsymbol{c}_s=$\emph{``Four young giraffes in a zoo, with one of them being fed leaves by a person.''}} \\
     \includegraphics[width=\imgwidth]{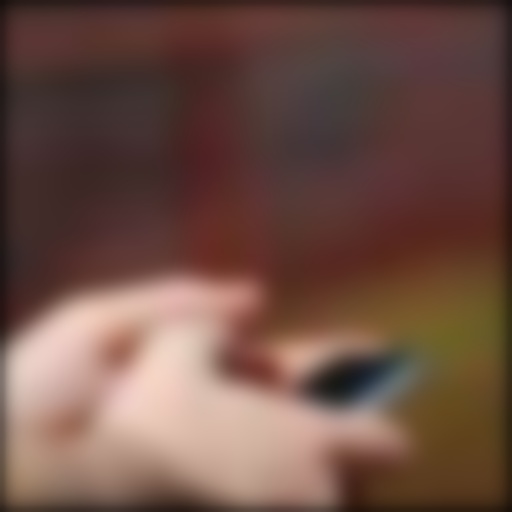} & 
     \includegraphics[width=\imgwidth]{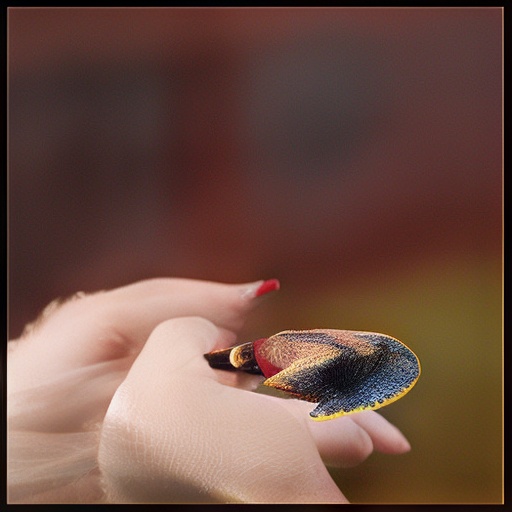} &
     \includegraphics[width=\imgwidth]{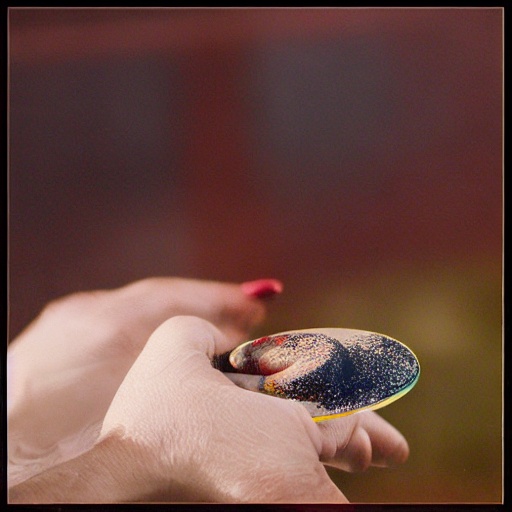} &
      \includegraphics[width=\imgwidth]{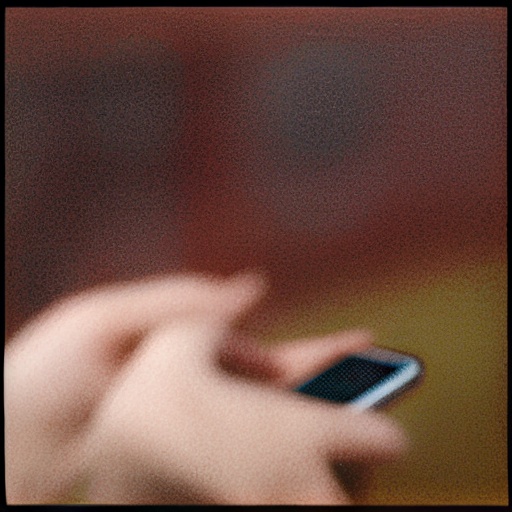} &
     \includegraphics[width=\imgwidth]{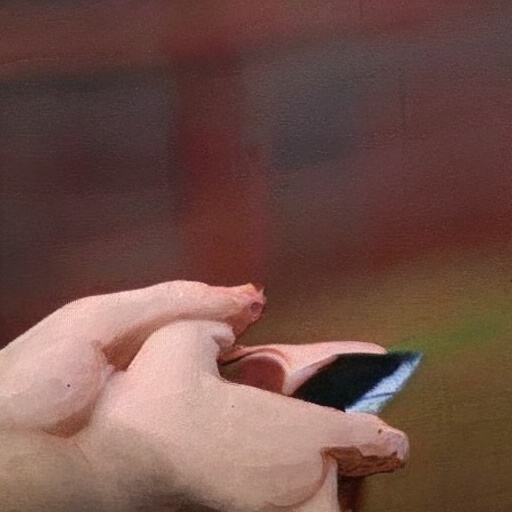} & 
     \includegraphics[width=\imgwidth]{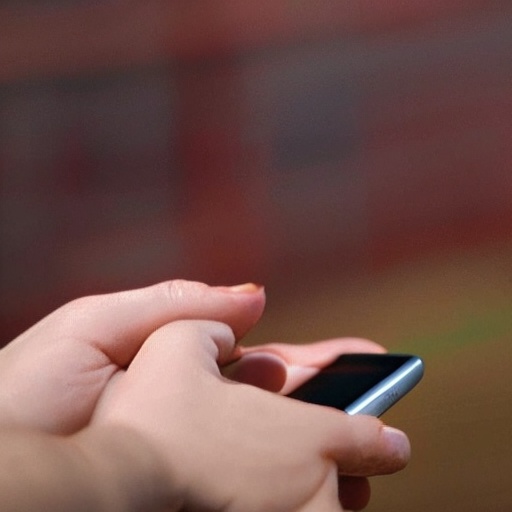} 
     & \includegraphics[width=\imgwidth]{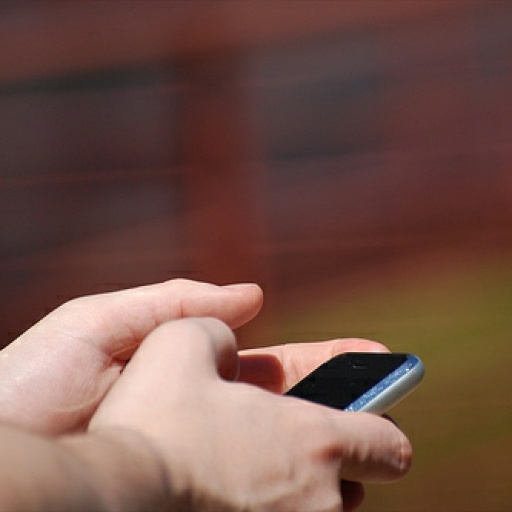}\\ 
     \multicolumn{7}{c}{$\boldsymbol{c}_s=$\emph{``Two hands holding and dialing a cellular phone.''}} \\

     \includegraphics[width=\imgwidth]{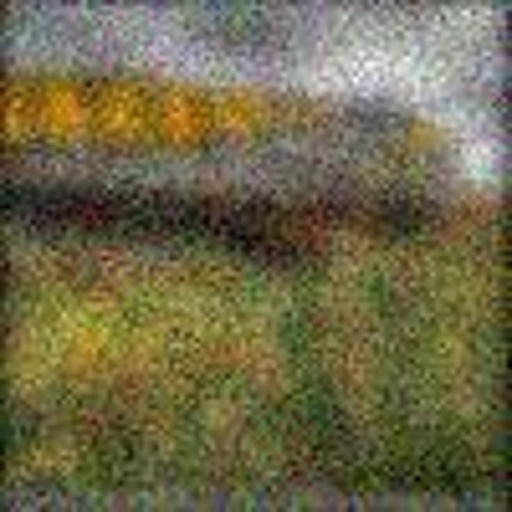} & 
     \includegraphics[width=\imgwidth]{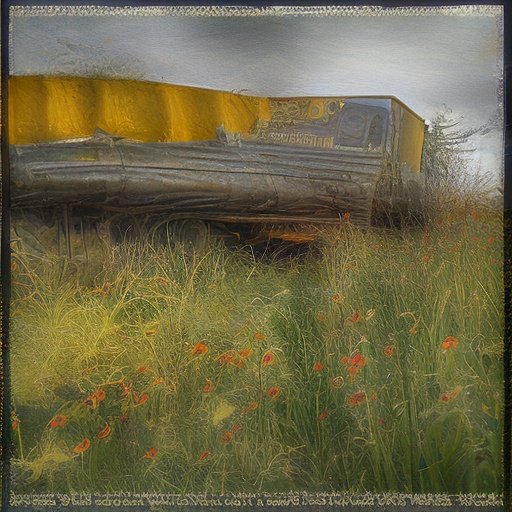} &
     \includegraphics[width=\imgwidth]{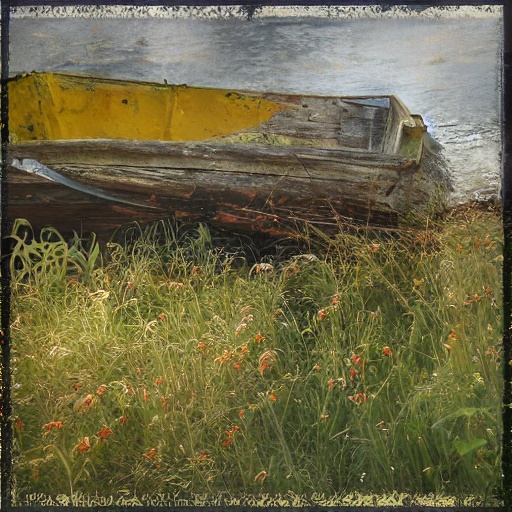} &
      \includegraphics[width=\imgwidth]{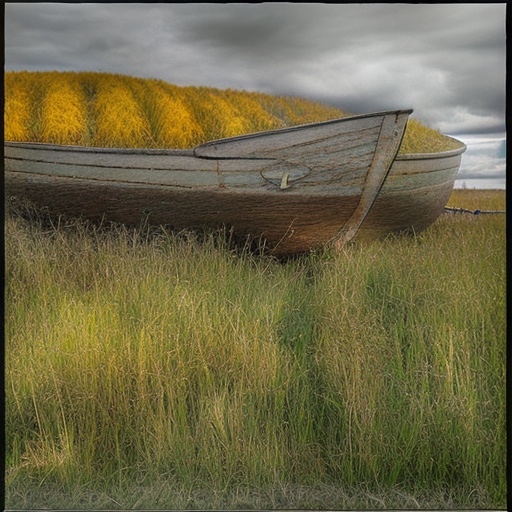} &
     \includegraphics[width=\imgwidth]{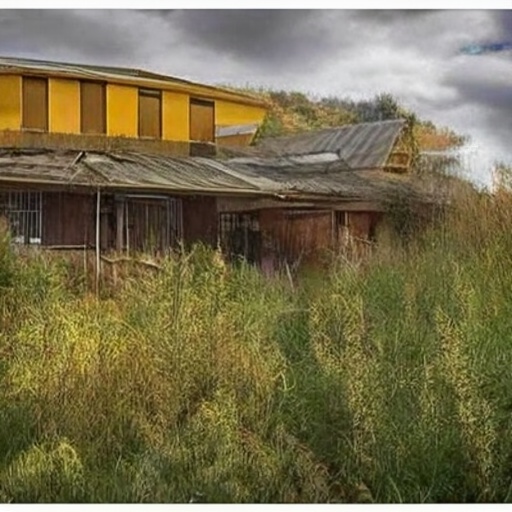} & 
     \includegraphics[width=\imgwidth]{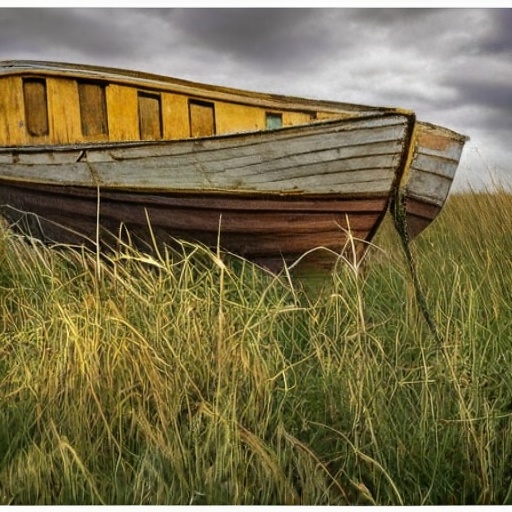} 
     & \includegraphics[width=\imgwidth]{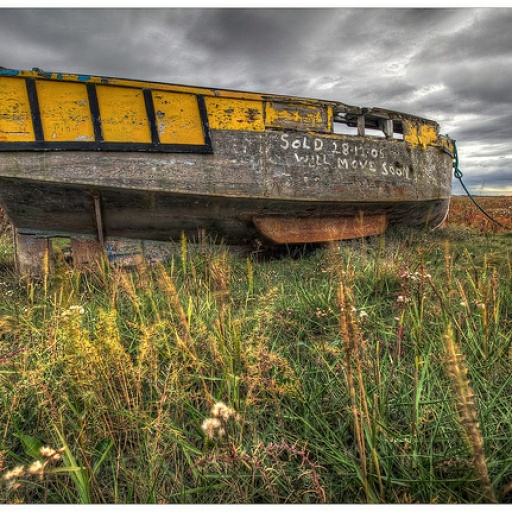}\\ 
     \multicolumn{7}{c}{$\boldsymbol{c}_s=$\emph{``An old boat sitting in the middle of a field.''}} \\

\end{tabular}
}
}
\vspace{-1em}
\caption{\textbf{Visual Comparison with other baselines}. 
Our method of integrating both the semantic prompt $\cs{}$ and the restoration prompt $\cres{}$ outperforms imge-to-image restoration (DiffBIR, Retrained ControlNet-SR) and naive zero-shot combination with semantic prompt.
It achieves more sharp and clean results while maintaining consistency with the degraded image.  
}
\vspace{-1.5em}
\label{fig:main-visuals}
\end{figure*}

%% file: sec/tabs/div2k_eval.tex
\begin{table}[t]

\begin{minipage}[t]{0.7\textwidth}
      \centering
\resizebox{\linewidth}{!}{
\begin{tabular} 
{@{}l@{\hspace{2mm}}*{4}{c@{\hspace{3mm}}}c@{\hspace{1mm}}r@{}}
\toprule
Method & FID$\downarrow$ & LPIPS$\downarrow$ & PSNR$\uparrow$ & SSIM$\uparrow$ &
\small{CLIP-I}$\uparrow$ 
\\
\midrule
Real-ESRGAN~\cite{wang2018esrgan}  & 32.37 & 0.312 & \textbf{22.52} & \textbf{0.646} & 0.683  \\
DiffBIR~\cite{Lin2023diffbir} (zero-shot) & 30.71 & 0.354 & {22.01} & 0.526 & 0.921 \\

StableSR & 24.44 & \textbf{0.311} & 21.62 & 0.533 & 0.928 \\

Ours w/o text (zero-shot) & 28.80 & 0.352 & 21.68 & {0.549} & 0.927 \\ 
Ours w/o text (finetuned) & \textbf{22.45} & 0.321 & 21.38 & 0.532 & \textbf{0.932} \\
\bottomrule
\end{tabular}
  \vspace{-1em}
}
      \end{minipage}
  \vspace{-1em}
    \begin{minipage}[t]{0.25\textwidth}
            \vspace{-20pt}
  \caption{Numerical results on the DIV2K testset without any prompt. }
  \label{tab:other_testset}
  \vspace{-1em}
    \end{minipage}
\end{table}
  \vspace{-1em}

%% file: sec/figs/sec4_exp/real-world-example.tex
\newcommand{\tuneimgwidth}{0.25\columnwidth}

\begin{figure}[t]
\centering
\setlength\tabcolsep{1pt}
{
\renewcommand{\arraystretch}{0.6}
\vspace{-0.35cm}
\resizebox{1\linewidth}{!}{
\begin{tabular}{@{\hspace{0mm}}*{6}{c}@{\hspace{0mm}}}

    \includegraphics[width=\tuneimgwidth]{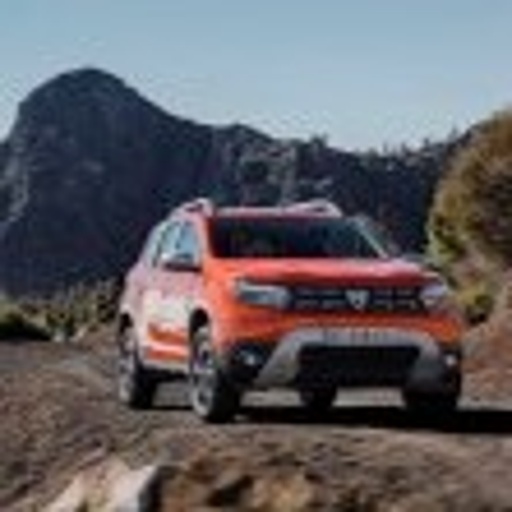} & 
    \includegraphics[width=\tuneimgwidth]{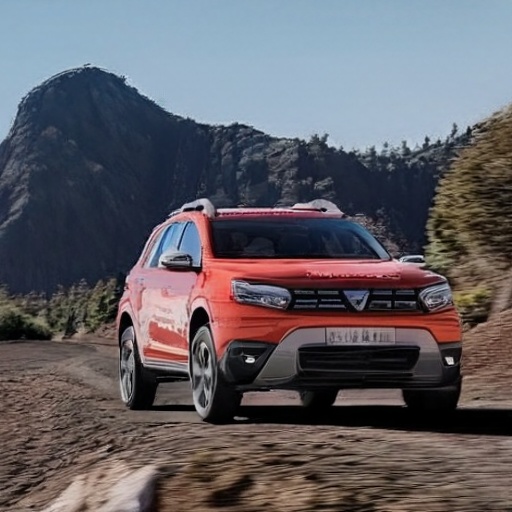} &

    \includegraphics[width=\tuneimgwidth]{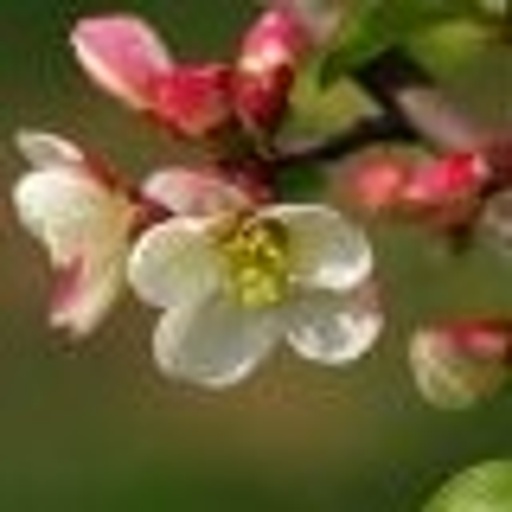} & 
    \includegraphics[width=\tuneimgwidth]{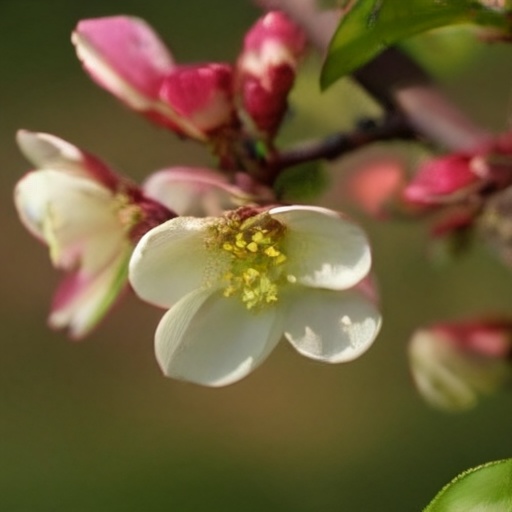} & 
    
    \includegraphics[width=\tuneimgwidth]{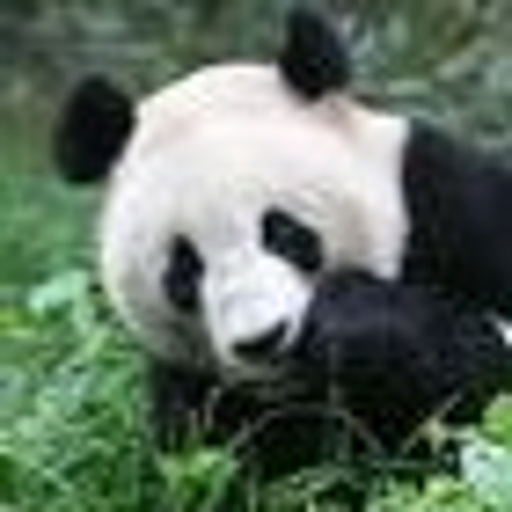} & 
    \includegraphics[width=\tuneimgwidth]{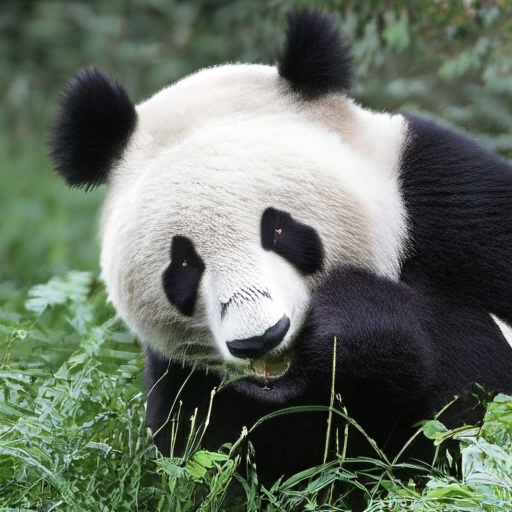}
     \\ 
    \includegraphics[width=\tuneimgwidth]{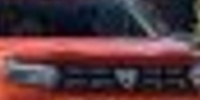} & 
    \includegraphics[width=\tuneimgwidth]{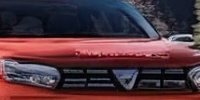} & 

    \includegraphics[width=\tuneimgwidth]{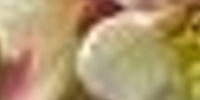} & 
    \includegraphics[width=\tuneimgwidth]{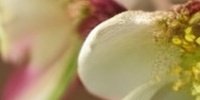} & 

    \includegraphics[width=\tuneimgwidth]{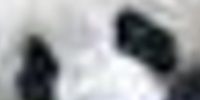} & 
    \includegraphics[width=\tuneimgwidth]{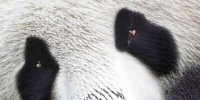} 
     \\ 


     Input & \textit{``SUV''} & Input & \textit{``Flower''} & Input & \textit{``Panda bear''} 
     \\
\end{tabular}
}
}
\vspace{-1.2em}
\caption{Real-world image restorations. Our framework enhances real-world low-quality images following semantic prompts provided by language models or user input. 
}
\label{fig:real-world restoration}
\vspace{-1em}
\end{figure}

%% file: sec/tabs/real-world.tex
\begin{table}[t]
\centering
\setlength{\tabcolsep}{1.6pt}
\renewcommand{\arraystretch}{1.0}

\resizebox{1\linewidth}{!}{

\begin{tabular}{@{}l@{\hspace{1mm}}*{9}{c@{\hspace{4mm}}}r}

\toprule

Metric & SwinIR~\cite{liang2021swinir}
& Diffbir~\cite{Lin2023diffbir}  & \multicolumn{4}{c}{Ablation of restoration prompt $\cres{}$ } & \multicolumn{4}{c}{Ablation of semantic prompt $\cs{}$} \\
\cmidrule(l{0mm}r{1mm}){2-3} 
\cmidrule(l{1mm}r{1mm}){4-7}
\cmidrule(l{1mm}r{0mm}){8-11}
Sem $\cs{}$ & \textcolor{BrickRed}{\XSolidBrush} 
& \textcolor{BrickRed}{\XSolidBrush} &

\multicolumn{4}{c}{\textcolor{teal}{\textcolor{BrickRed}{\XSolidBrush}}} &

\textcolor{BrickRed}{\XSolidBrush} & \textcolor{teal}{short} & \textcolor{teal}{unmatched} & \textcolor{teal}{full} \\


Res $\cres{}$ & \textcolor{BrickRed}{\XSolidBrush} 
& \textcolor{BrickRed}{\XSolidBrush} 

& \textit{"Deblur 0.6"} & \textit{"... 1.0"}  & \textit{"... 2.0"} & \textit{"... 5.0"}

&\multicolumn{4}{c}{ \textit{"Remove all degradation"}} 
\\
CLIP-IQA$\uparrow$ & 0.6190

& 0.6983 

&0.4074 & 0.4159 & 0.4831 & 0.4797 

& 0.5818 & 0.5750 &
{0.6460} & {0.6610}

 \\

MUSIQ$\uparrow$ & 63.65
& 69.69 

& 35.82 & 39.87 & 49.98 & 48.81

& 61.68 & 60.27 &

{67.54} & {67.82} 

 \\
 
MANIQA$\uparrow$ & 0.4035
& 0.2619 

& 0.2299 & 0.2487 & 0.2970 & 0.2821

& 0.3694 & 0.3637 & 
{0.4355} & {0.4474} 

 \\
\bottomrule
\end{tabular}

}
\caption{Quantitative results on real-world images. 
Our aesthetic quality
is improved when increasing restoration strength or using a more accurate and detailed prompt.
}
\label{tab:real-world-result}
\vspace{-2.5em}
\end{table}

%% file: sec/figs/sec4_exp/content_prompt.tex
\newcommand{\tuneimgwidtha}{0.25\columnwidth}

\begin{figure*}[!t]
\centering
\setlength\tabcolsep{1pt}
{
\renewcommand{\arraystretch}{0.6}
\resizebox{1\linewidth}{!}{
\begin{tabular}{@{}*{6}{c}@{}}
    \includegraphics[width=\tuneimgwidtha]{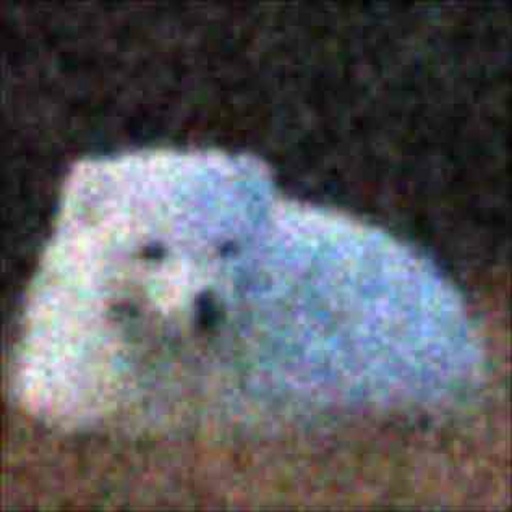} & 
    \includegraphics[width=\tuneimgwidtha]{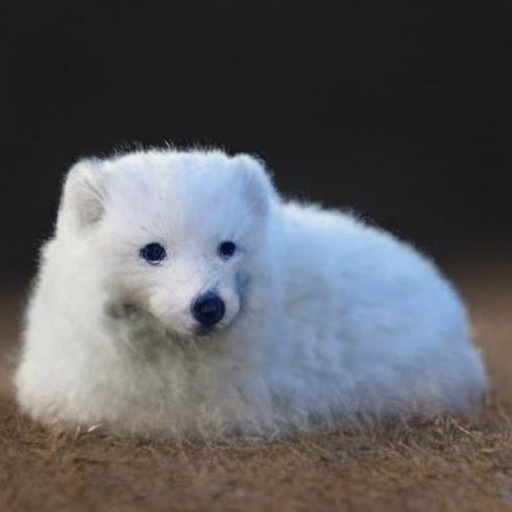} & 
    \includegraphics[width=\tuneimgwidtha]{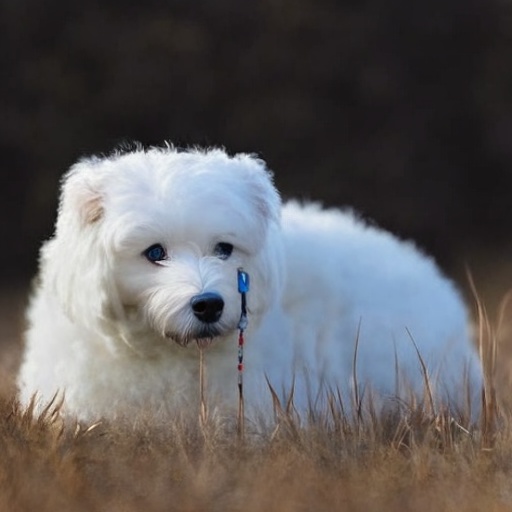} & 
    \includegraphics[width=\tuneimgwidtha]{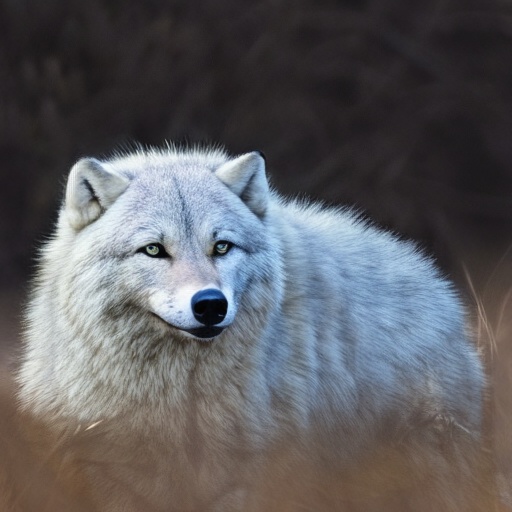} & 
    \includegraphics[width=\tuneimgwidtha]{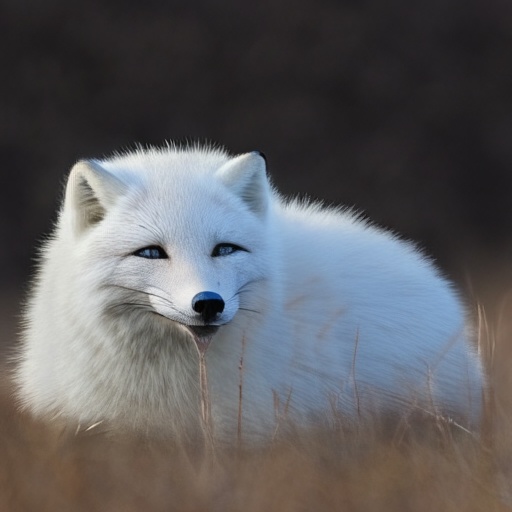} & 
    \includegraphics[width=\tuneimgwidtha]{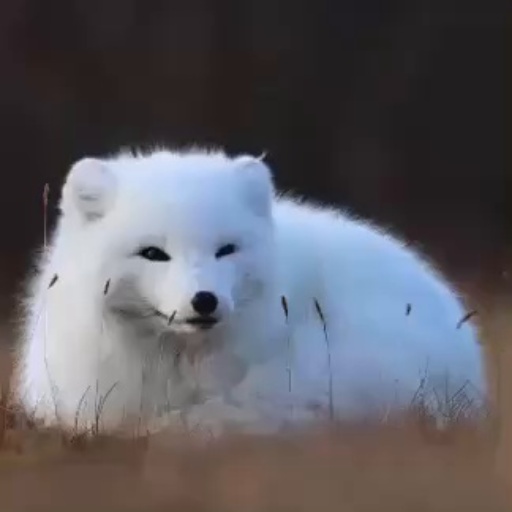}

     \\ 


     Input & \emph{``''} & \emph{``Bichon Frise dog''} & \emph{``grey wolf''} & \emph{``white fox''} & Reference  \\

    \includegraphics[width=\tuneimgwidtha]{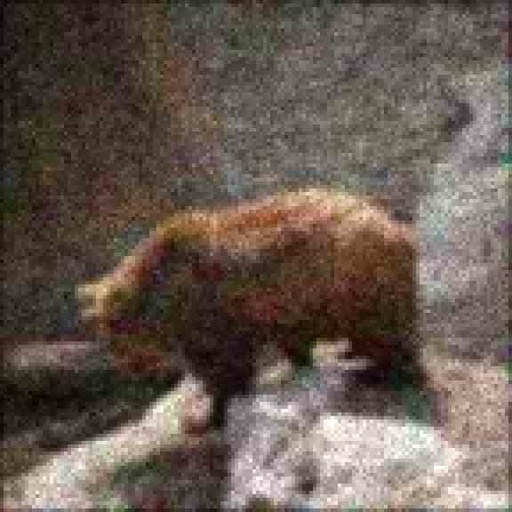} & 
    \includegraphics[width=\tuneimgwidtha]{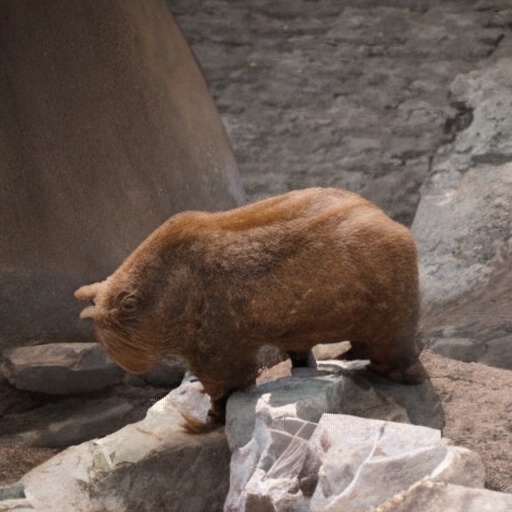} & 
    \includegraphics[width=\tuneimgwidtha]{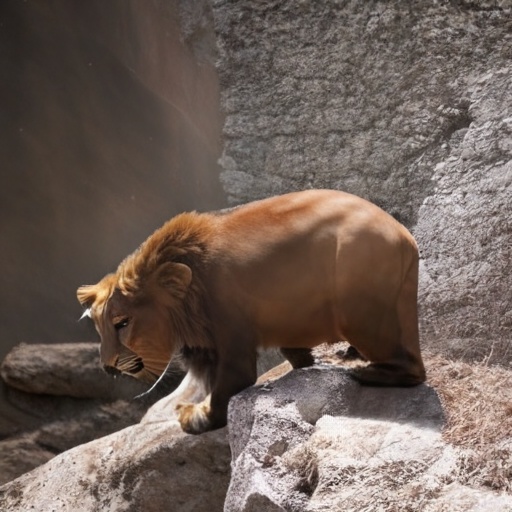} & 
    \includegraphics[width=\tuneimgwidtha]{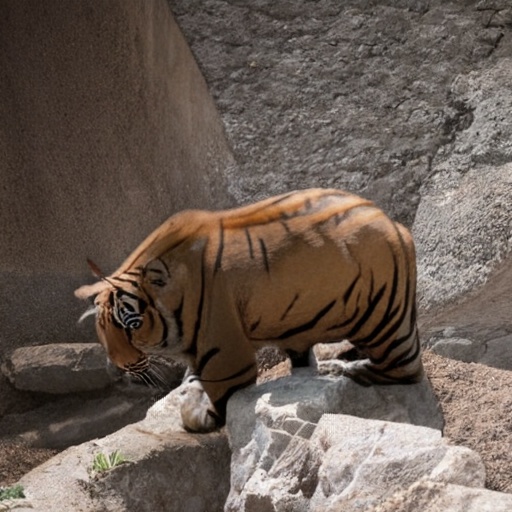} & 
    \includegraphics[width=\tuneimgwidtha]{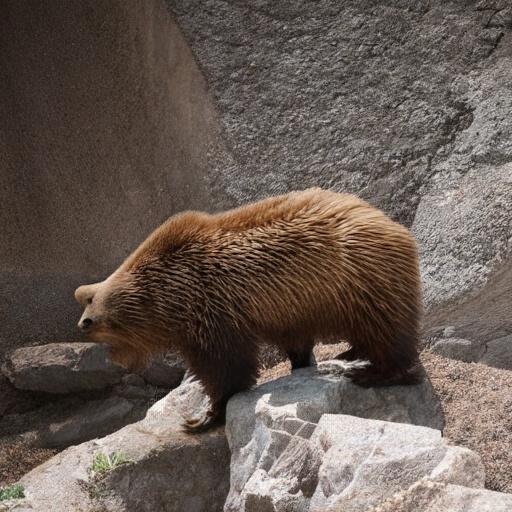} & 
    \includegraphics[width=\tuneimgwidtha]{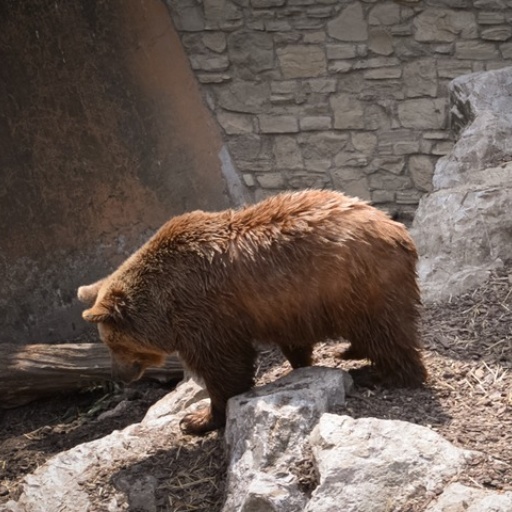}

     \\ 

     Input & \emph{ ``''} & \emph{``lion''} & \emph{`` tiger''} & \emph{``bear''} & Reference

    
\end{tabular}
}
}
\vspace{-1em}
\caption{\textbf{Test-time semantic prompting}. Our framework restores degraded images guided by flexible semantic prompts, while unrelated background elements and global tones remain aligned with the degraded input conditioning.  }
\label{fig:content_prompting}
\vspace{-1em}
\end{figure*}

%% file: sec/figs/sec4_exp/degradation_prompt.tex
\vspace{-1.5em}

\begin{figure}[t]
    \begin{minipage}[t]{0.65\textwidth}
  \centering

   \includegraphics[width=1\linewidth]{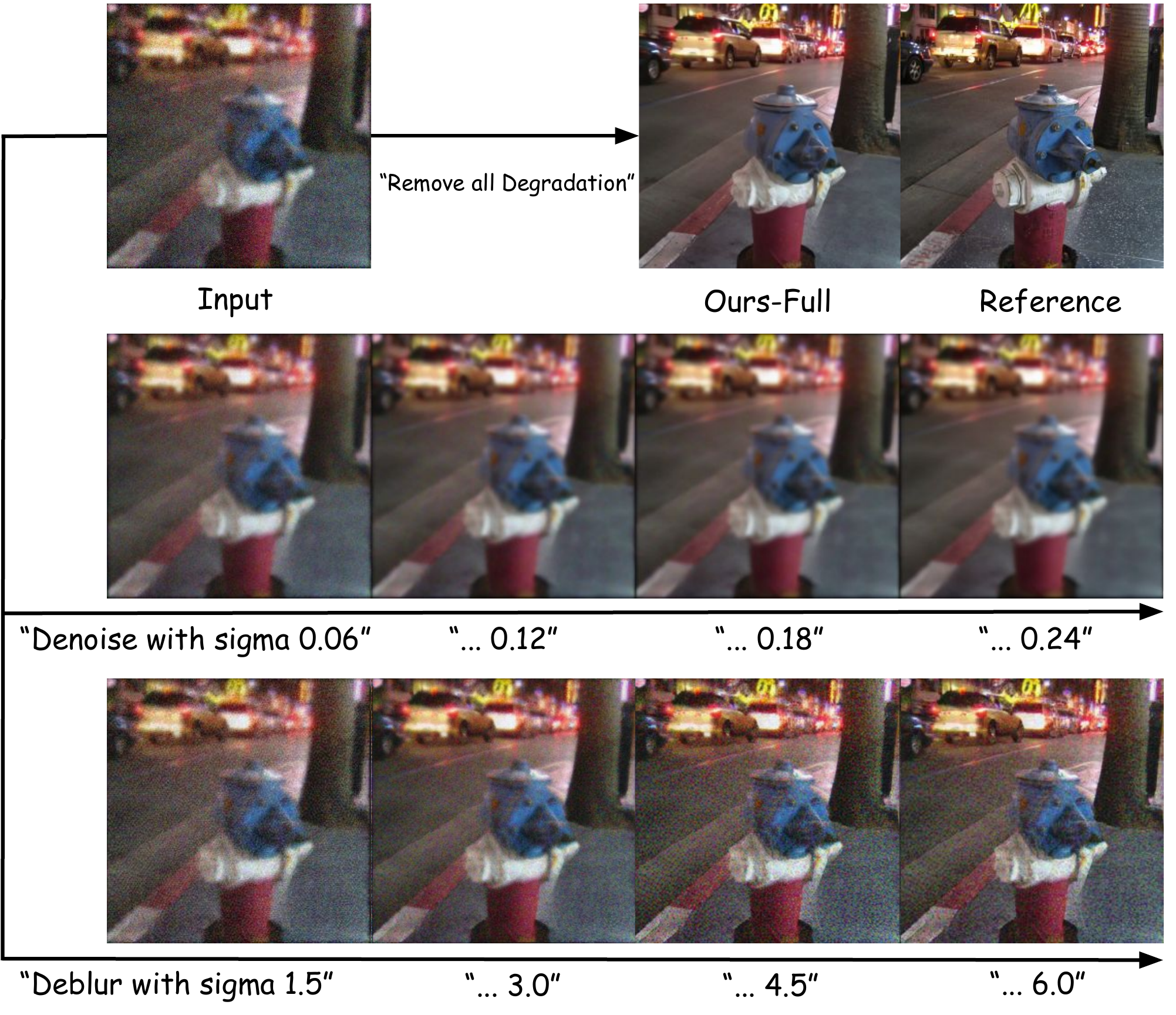}
    \vspace{-3em}
        \end{minipage}
    \begin{minipage}[t]{0.34\textwidth}
      \vspace{-21em}
   \caption{\textbf{Prompt space walking visualization for the restoration prompt}. Given the same degraded input (upper left) and empty semantic prompt $\varnothing$, our method can decouple the restoration direction and strength via only prompting the \textbf{quantitative number in natural language}. An interesting finding is that our model learns a continuous range of restoration strength from discrete language tokens. 
   }
  \label{fig:restoration_decoupling}
   \end{minipage}
 \vspace{-3em}
\end{figure}


%% file: sec/tabs/ablation.tex
\begin{table}[t]
\centering
\setlength{\tabcolsep}{1.6pt}
\renewcommand{\arraystretch}{1.0}
\resizebox{0.9\linewidth}{!}{
\begin{tabular} 
{@{}l@{\hspace{1mm}}*{6}{c@{\hspace{1mm}}}c@{\hspace{1mm}}r@{}}
\toprule
Method & Sem $\cs{}$ & Res $\cres{}$ & FID$\downarrow$ & LPIPS$\downarrow$ & PSNR$\uparrow$ & SSIM$\uparrow$ & \small{CLIP-I}$\uparrow$ & \small{CLIP-T}$\uparrow$ \\
\midrule
ControlNet-SR & \textcolor{red}{\XSolidBrush} & \textcolor{red}{\XSolidBrush} &
13.65 & 0.222 & 23.75 & 0.669 & 0.938 & 0.300\\
ControlNet-SR & \textcolor{teal}{\Checkmark} & \textcolor{teal}{\Checkmark} & 12.14 & 0.223 & 23.50 & 0.660 & 0.940 & \textbf{0.306}\\
Ours &\textcolor{teal}{\Checkmark} &  wo param & 11.58 & 0.223 & 23.61 & 0.665 & 0.942 & 0.305 \\\hline

Ours & \textcolor{teal}{\Checkmark} & \textcolor{teal}{\Checkmark} &  \textbf{11.34} & \textbf{0.219} & 23.61 & 0.665 & \textbf{0.943} & 0.305\\
\bottomrule
\end{tabular}
}
\caption{Ablation of architecture and degradation strength in $c_r$ }
\label{tab:ablation_study}
 \vspace{-3em}
\end{table}

%% file: sec/5_conclusion.tex
\section{Conclusion}

We have presented a unified text-driven framework for instructed image restoration using semantic and restoration prompts.
We design our model in a decoupled way to better preserve the semantic text-to-image generative prior while efficiently learning to control both the restoration direction and its strength.
To the best of our knowledge, this is the first framework to support both semantic and parameter-embedded restoration instructions simultaneously, allowing users to flexibly prompt-tune the results up to their expectations.
Our extensive experiments have shown that our method significantly outperforms prior works in terms of both quantitative and qualitative results.
Our model and evaluation benchmark have established a new paradigm of instruction-based image restoration, paving the way for further multi-modality generative imaging applications.


%% file: sec/X_suppl.tex


\section{Implementation Details}
\subsection{Derivation Details in Equation 2}
We propose to compute the distribution of latents $p(\boldsymbol{z}_t |\bm{y}, \cs{}, \cres{})$ conditioned on degraded image $\y{}$, semantic prompt $\cs{}$ and restoration prompt $\cres{}$. Using the Bayes' decomposition similar to score-based inverse problem~\cite{ncsn,song2022inverse_medical_imaging}, we have
\begin{eqnarray}
{ p(\boldsymbol{z}_t |\bm{y}, \cs{}, \cres{})} = { p(\boldsymbol{z}_t, \bm{y}, \cs{}, \cres{}) / p( \bm{y}, \cs{}, \cres{}).}
\end{eqnarray}
Then, we compute gradients with respect to $\boldsymbol{z}_t$, and remove the gradients of input condition $\nabla_{\ztlatents{}}\log p( \y{}, \cs{}, \cres{})=0$ as: 
\begin{eqnarray}
&&\hspace{-3em} \nabla_{\ztlatents{}}{\log p(\boldsymbol{z}_t |\bm{y}, \cs{}, \cres{})} \nonumber \\
&=&\nabla_{\ztlatents{}}\log p(\boldsymbol{z}_t, \y{}, \cs{}, \cres{}) \\
&=&\nabla_{\ztlatents{}}\log [p(\cs{})\cdot p(\boldsymbol{z}_t| \cs{}) \cdot p(\bm{y},\cres{}|\cs{}, \boldsymbol{z}_t)] \\
&=&\nabla_{\ztlatents{}}\log [ p(\boldsymbol{z}_t| \cs{}) \cdot p(\bm{y},\cres{}|\cs{}, \boldsymbol{z}_t)] \\
&=&\nabla_{\ztlatents{}}\log p(\boldsymbol{z}_t| \cs{}) + \nabla_{\ztlatents{}}\log p(\bm{y}, \cres{}|\cs{}, \boldsymbol{z}_t).
\end{eqnarray}

We assume $\y{}$ is generated through a degradation pipeline as $\y{}=\text{Deg}(\bm{x}, \cres{})$, thus it is independent of $\cs{}$ with $\bm{x}$ and $\cres{}$ provided as condition. Removing redundant $\cs{}$ condition, the second term in the last equation can be approximated as:
%
\begin{eqnarray}
    && \hspace{-3em}\nabla_{\ztlatents{}}\log p(\bm{y}, \cres{}|\cs{},
  \boldsymbol{z}_t) \nonumber \\ 
  & \approx& \nabla_{\ztlatents{}}\log p(  \y{}, \cres{}| \boldsymbol{z}_t) \\ 
  &=& \nabla_{\ztlatents{}}\log p( \cres{}| \boldsymbol{z}_t) +\nabla_{\ztlatents{}}\log p( \y{} | \boldsymbol{z}_t, \cres{}) \\ 
  &=& \nabla_{\ztlatents{}}\log p(  \y{}| \boldsymbol{z}_t,\cres{})
\end{eqnarray}

In summary of the above equations, we derive the Equation 2 in the main manuscript
\begin{equation}
\begin{split}
    &\nabla_{\ztlatents{}}{\log p(\boldsymbol{z}_t |\bm{y}, \cs{}, \cres{})} \\
    &\approx
   \underbrace{\nabla_{\ztlatents{}}\log p(\boldsymbol{z}_t |\cs{})}_{\text{Semantic-aware (\textcolor{cvprblue}{frozen})}} + \underbrace{\nabla_{\ztlatents{}}\log p(  \y{}| \boldsymbol{z}_t,\cres{}),}_{\text{Restoration-aware (\textcolor{BrickRed}{learnable})}}
\end{split}
\end{equation}
where $\nabla_{\ztlatents{}}\log p(  \y{}| \boldsymbol{z}_t,\cres{})$ is synthesized using stochastic degradation pipeline $\y{} = \text{Deg}( x,\cres{})$ to train our ControlNet.

\input{sec/algs/alg_degradation}
\input{sec/tabs/supp_prompt_ablation}

\input{sec/tabs/supp_arch_ablation}

\subsection{Pseudo Code for Degradation Synthesis}
To support the learning of restoration-aware term $\nabla_{\ztlatents{}}\log p(  \y{}| \boldsymbol{z}_t,\cres{})$, we synthesize the degradation image $\bm{y}$ using clean image $\bm{x}$ with the algorithm presented in~\cref{alg:degradation_pipeline}. 
First, we randomly choose one from Real-ESRGAN pipeline and our parameterized degradation.
Then the degraded image from Real-ESRGAN pipeline is paired with restoration prompt $\cres{}=$\emph{``Remove all degradation''}.
In our parameterized degradation, all processes are paired with restoration prompts $\cres{}$ listed in Table 2 of the main manuscript (\eg, \emph{Deblur with sigma 3.0}).







\section{More Ablation Study}
~\cref{tab:ablation_prompts} provides more comprehensive ablations of text prompts by providing different information to our image-to-image baseline. Semantic prompts significantly improve image quality as shown in better FID and CLIP-Image, but reduce the similarity with ground truth image. Restoration types and parameters embedded in the restoration prompts both improve image quality and fidelity.
~\cref{tab:ablation_arch} presents a comparison of our skip feature modulation $f_{skip}$ with that in StableSR~\cite{wang2023stablesr}  which modulates both skip feature $f_{skip}$ from encoder and upsampling feature $f_{up}$ from decoder. We observe that modulating $f_{up}$ does not bring obvious improvements. One possible reason is that $\gamma$ and $ \beta$ of the middle layer adapts to the feature in the upsampling layers.

\input{sec/figs/supp/semantic_prompt}
\input{sec/figs/supp/mj_result}

\input{sec/figs/sec4_exp/degradation_prompt_1}

\vspace{2em}
\section{Multiple Objects Semantic Prompting}
Besides single semantic restoration, real applications may involve multiple objects with different semantic categories (\eg ~\cref{fig:supp_content_prompting}). In each column, we guide the upper part of the image with \textit{peppers}, \textit{bananas} or \textit{leaves}, while the lower part can be restored as \textit{potatoes} or \textit{stones}.
Thanks to the cross attention mechanism, multiple semantics can be spatially decoupled and recombined following the user's prompts, thus yielding better restoration for both objects.

\section{More Restoration Prompting}

~\cref{fig:midjourney_restoration} shows the application of restoration prompt on images with different degradations and content, including Midjourney image and real-world cartoon. Since these images are not in our training data domain, a blind enhancement with prompt \emph{``Remove all degradation''} can not achieve satisfying results. Utilizing restoration prompting  (\eg, \emph{``Upsample to 6.0x; Deblur with sigma 2.9;''}) can successfully guide our model to improve the details and color tones of the Midjourney image. In the right half, a manually designed restoration prompt also reduces the jagged effect to smooth the lines in the cartoon image.

To study whether the model follows restoration instructions, a dense walking of restoration prompt is presented in~\cref{fig:restoration_decoupling_supp}. From left to right, we increase the strength of denoising in the restoration prompt. From top to the bottom, the strength of deblurring gets larger.
The results demonstrate that our restoration framework refines the degraded image continuously following the restoration prompts
\input{sec/figs/supp/real-input-long-prompt}




\section{Real-world images}
Although our model is trained on synthetic degradation, it generalize to real-world data RealPhoto60~\cite{SUPIR}, as shown in the ~\cref{fig:real-input-long-prompt}.
Compared to a model without semantic prompt, the synthetic semantic prompts from LLAVA~\cite{liu2023llava} enhance fine-level details in ~\cref{fig:real-input-no-prompt-long-prompt} (\eg, grass under sheep in the upper left figure, and the staircase in the mountain in lower right photo). 
These results demonstrate an additional potential advantage of employing language prompts in real-world restoration: the ease of leveraging the logical reasoning capabilities in pre-trained large language models.


\section{Limitation}
\input{sec/figs/sec4_exp/limitation}

\input{sec/figs/supp/real-input-no-prompt-long-prompt}

Although our framework can generate high-fidelity results following semantic and restoration prompts, it is prone to occasional hallucinations. As shown in ~\cref{fig:hallucinations}, the image quality is degraded and the semantic is unclear when the input prompt (\textit{"Snow leopard"}) is misaligned with the ground truth (\textit{"Panda bear"}). Instead of relying on user input or frozen language models, one future direction can be fine-tuning multimodal language models to automatically provide more accurate instructions, thus reducing hallucinations. 
In addition, we plan to scale up our model parameters and extend it to more diverse and realistic degradation types in our future work.

\input{sec/figs/rebuttal/r1q2_derain_dehaze}

\section{Mixed and universal degradation.}

Our method can also restore mixed degradation 
(Figure 1 and Table 1 in the
paper
).
For unseen degradations such as haze or rain,
our pretrained model can still handle  them properly, as shown in the figure below, since our pretrained prior and training data contains those concepts (\eg, "A clear sky").

\section{ Comparison of inference cost}
Our method takes 1.4s (50 DDIM steps ) to run on TPUv4 and 1130~GFLOPS per step). Our UNet model
has 1240~M (275~M trainable) parameters. The overall computation cost is comparable with 1203~M StableSR (19.3s on GPU) and 1510~M DiffBir (based on Stable Diffusion, 10.9s on GPU), and less than SUPIR (based on 2.6B SD-XL).

\input{sec/tabs/rebuttal_task_comparison}

\section{Comparison with more models}
We follow the test set design of task-specific SwinIR.
In the table~\ref{tab:swinir_comparison}, our method outperforms the task-specific SwinIR and achieves lower LPIPS in evaluation.
Following StableSR and Real-esrgan, our model is trained on large-scale open-domain images with ESRGAN degradations, which has a noticeable difference with the degradation in all-in-one restoration mentioned by reviewers (\eg, DA-CLIP considers raindrop but Real-esrgan does not).
Thus, comparing our framework and other concurrent work (\eg, DiffBir) to all-in-one restoration techniques proves difficult. To alleviate the concern, we evaluate our framework on the denoising testset of CBSD68 and our method achieves a comparable LPIPS (0.305) with DA-CLP (0.294). 



\section{More Visual Comparison}
More visual comparisons with baselines are provided in ~\cref{fig:main-visuals-supp1}.

\input{sec/figs/sec4_exp/main_visuals_supp_merge}



%% file: sec/algs/alg_degradation.tex
\algnewcommand{\ElseIIf}[1]{\algorithmicelse\ #1}
\begin{algorithm}[t]
\caption{ \texttt{\TIP{}} Degradation Pipeline in Training}
\label{alg:degradation_pipeline}
\begin{algorithmic}

\State \textbf{Inputs:} $\bm{x}$: \text{Clean image} \vspace{.25em}

\State {\bf Outputs:} $\y{}$: \text{Degraded image}; $\cres{}$: \text{Restoration prompt}
\vspace{.5em}

\noindent \text{type} $\gets$ \Call{RandChoice}{Real-ESRGAN, Param} \vspace{.5em}

\If{type = Real-ESRGAN} // Real-ESRGAN degradation
    \State $\y{} \gets  \bm{x}$
    \State Deg $\gets$ \Call{Random}{Real-ESRGAN-Degradation}
    \For{\Call{Process}{} \textbf{in} Deg}:
        \State $\bm{y}$ $\gets$ \Call{Process}{$\bm{y}$}
    \EndFor
    \State $\cres{}$ $\gets$ \emph{``Remove all degradation''}
\Else // Parameterized degradation
    \State $\cres{} \gets \emptyset$
    \State $\y{} \gets \bm{x}$
    \State Deg $\gets$ \Call{Random}{Parametrized-Degradation}
    \For{\Call{Process}{}, $\cres{}_p$ in  Deg}:
        \State $\bm{y}$ $\gets$ \Call{Process}{$\bm{y}$, $\cres{}_p$}
        \State $\cres{}$ $\gets$ \Call{Concat}{$\cres{}, \cres{}_p$}

    \EndFor
\EndIf \\
\Return $\y{}$, $\cres{}$
\end{algorithmic}
\end{algorithm}

%% file: sec/tabs/supp_prompt_ablation.tex
\begin{table}[t]
\centering
\setlength{\tabcolsep}{1.6pt}
\renewcommand{\arraystretch}{1.0}
\resizebox{1\linewidth}{!}{
\begin{tabular} 
{@{}l@{\hspace{3mm}}*{2}{c@{\hspace{4mm}}}*{5}{c@{\hspace{3mm}}}c@{\hspace{3mm}}c@{}}
\toprule
Method & Sem & Res Type & Res Param & FID$\downarrow$ & LPIPS$\downarrow$ & PSNR$\uparrow$ & SSIM$\uparrow$ & \small{CLIP\textsubscript{im}}$\uparrow$ & \small{CLIP\textsubscript{tx}}$\uparrow$ \\
\midrule
Ours 
& \textcolor{red}{\XSolidBrush} & \textcolor{red}{\XSolidBrush}  & \textcolor{red}{\XSolidBrush} & 13.60 & 0.221 & 23.65 & 0.664 & 0.939 & 0.300
\\
Ours 
& \textcolor{teal}{\Checkmark} & \textcolor{red}{\XSolidBrush}  & \textcolor{red}{\XSolidBrush}
& 11.71 & 0.226 & 23.55 & 0.663 & 0.941 & 0.305\\
Ours & \textcolor{teal}{\Checkmark} & \textcolor{teal}{\Checkmark} & \textcolor{red}{\XSolidBrush}  & 11.58 & 0.223 & 23.61 & 0.665 & 0.942 & 0.305 \\
\hline

Ours & \textcolor{teal}{\Checkmark} & \textcolor{teal}{\Checkmark} & \textcolor{teal}{\Checkmark} & \textbf{11.34} & \textbf{0.219} & 23.61 & 0.665 & \textbf{0.943} & \textbf{0.306}\\

\bottomrule
\end{tabular}
}
\caption{Ablation of prompts provided during both training and testing. We use an image-to-image model with our modulation fusion layer as our baseline. Providing semantic prompts significantly increases the image quality (1.9 lower FID) and semantic similarity (0.002 CLIP-Image), but results in worse pixel-level similarity. In contrast, degradation type information embedded in restoration prompts improves both pixel-level fidelity and image quality. Utilizing degradation parameters in the restoration instructions further improves these metrics. }
\label{tab:ablation_prompts}
 \vspace{-2em}
\end{table}

%% file: sec/tabs/supp_arch_ablation.tex
\begin{table}[t]
\centering
\setlength{\tabcolsep}{1.6pt}
\renewcommand{\arraystretch}{1.0}
\begin{tabular} 
{@{}l@{\hspace{1mm}}*{3}{c@{\hspace{1mm}}}c@{\hspace{1mm}}r@{}}
\toprule
Method & Modulate $f_{skip}$ & Modulate $f_{up}$ & Relative Param & FID$\downarrow$ & LPIPS$\downarrow$ \\
\midrule
Ours w/ prompts & \textcolor{red}{\XSolidBrush} & \textcolor{red}{\XSolidBrush} & 1 &12.14 & 0.223 \\
Ours w/ prompts & \textcolor{teal}{\Checkmark} & \textcolor{teal}{\Checkmark} & 1.06 & 11.21 & 0.219 \\
Ours w/ prompts & \textcolor{teal}{\Checkmark} &  \textcolor{red}{\XSolidBrush} & 1.03 & 11.34 & 0.219 \\
\bottomrule
\end{tabular}
\caption{Ablation of the architecture. Modulating the skip feature $f_{skip}$ improves the fidelity of the restored image with 3\% extra parameters in the adaptor, while further modulating the backbone features $f_{up}$ does not bring obvious advantage.}
\label{tab:ablation_arch}
 \vspace{-2em}
\end{table}

%% file: sec/figs/supp/semantic_prompt.tex
\newcommand{\tuneimgwidths}{0.25\columnwidth}

\begin{figure*}[tb]
\centering
\setlength\tabcolsep{1pt}
{
\renewcommand{\arraystretch}{0.6}
\resizebox{1\linewidth}{!}{
\begin{tabular}{@{}*{7}{c}@{}}

    \includegraphics[width=\tuneimgwidths]{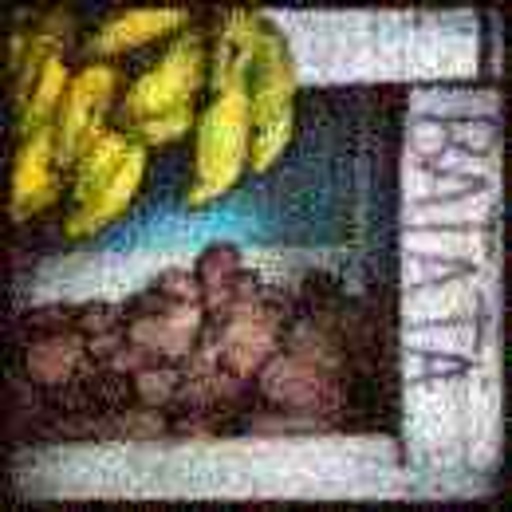} & 
    \includegraphics[width=\tuneimgwidths]{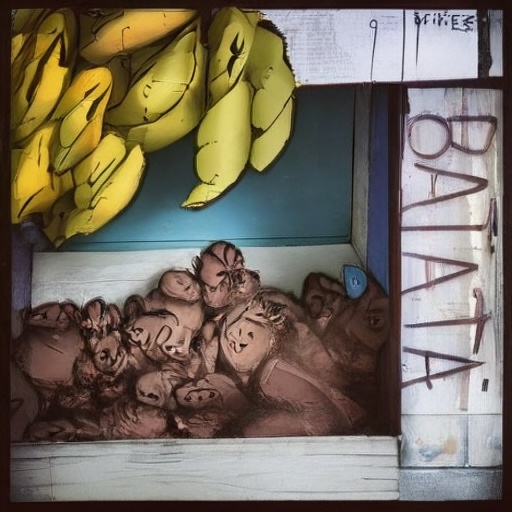} & 
    \includegraphics[width=\tuneimgwidths]{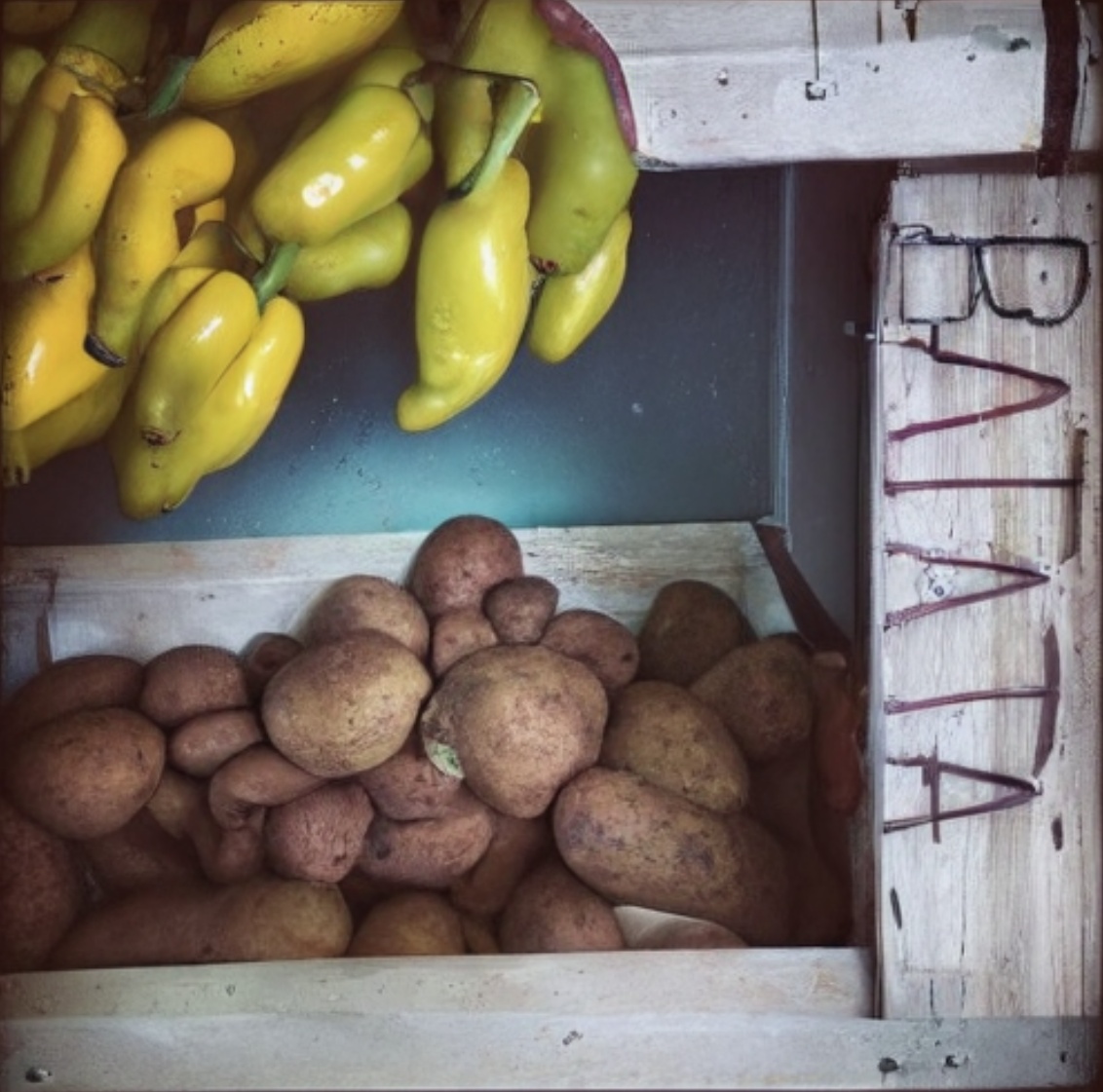} & 
    \includegraphics[width=\tuneimgwidths]{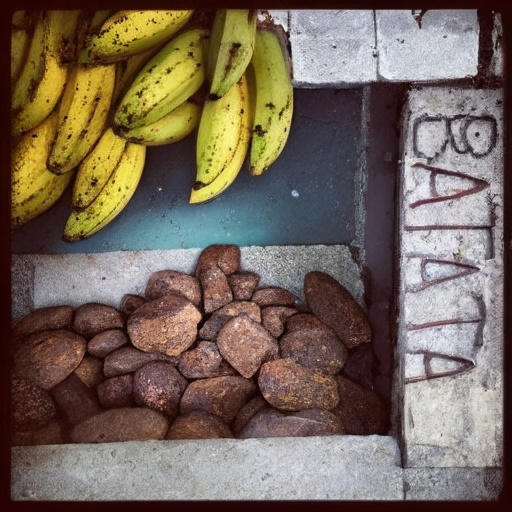} & 
    \includegraphics[width=\tuneimgwidths]{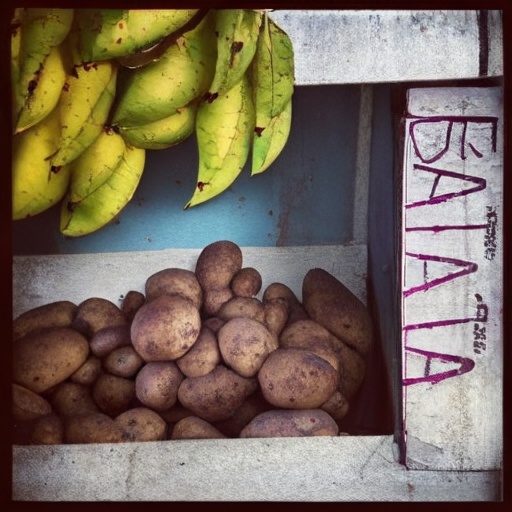} & 
    \includegraphics[width=\tuneimgwidths]{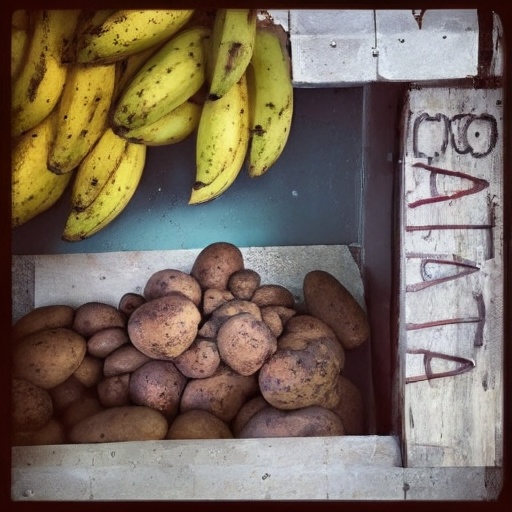} & 
    \includegraphics[width=\tuneimgwidths]{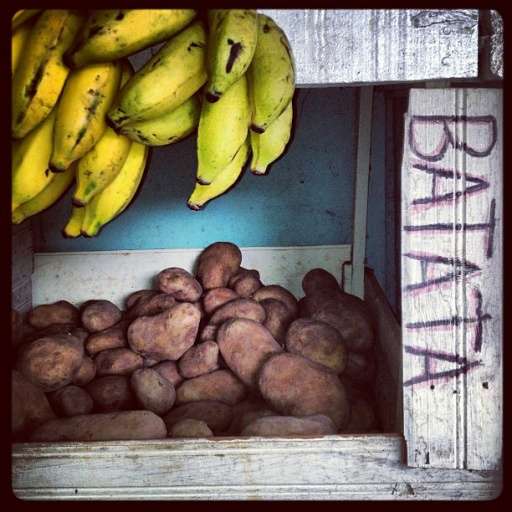}
     \\ 
     Input & \emph{``''}, & \emph{``peppers ... potatoes''},& \emph{``bananas... stones''} & \emph{``leaves... potatoes''} & \emph{``bananas...potatoes''} & Reference  \\
\end{tabular}
}
}
\vspace{-1em}
\caption{\textbf{More semantic prompting for images with multiple objects}.   }
\label{fig:supp_content_prompting}
\vspace{-1em}
\end{figure*}

%% file: sec/figs/supp/mj_result.tex
\newcommand{\tuneimgwidthx}{0.3\columnwidth}

\begin{figure}[t]
\centering
\setlength\tabcolsep{1pt}
{
\renewcommand{\arraystretch}{0.6}
\resizebox{1\linewidth}{!}{
\begin{tabular}{@{}*{6}{c}@{}}
    \includegraphics[width=\tuneimgwidthx]{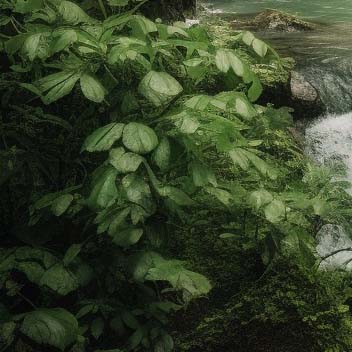} & 
    \includegraphics[width=\tuneimgwidthx]{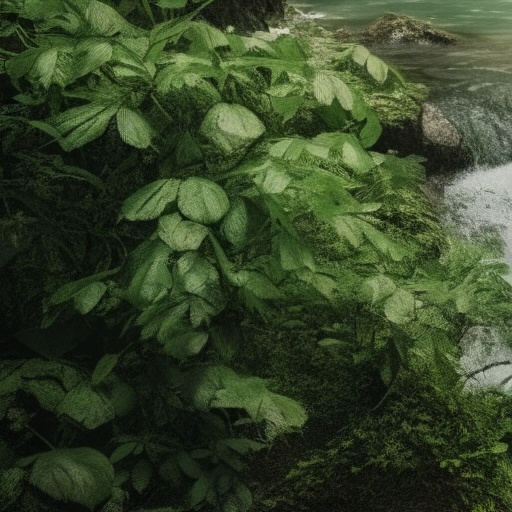} & 
    \includegraphics[width=\tuneimgwidthx]{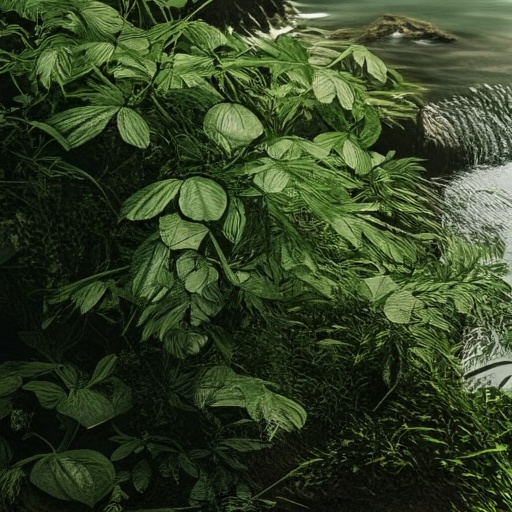} 
    
     
    &
    \includegraphics[width=\tuneimgwidthx]{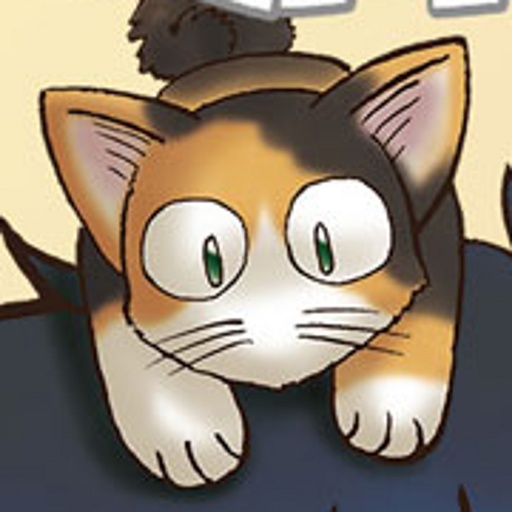} & 
    \includegraphics[width=\tuneimgwidthx]{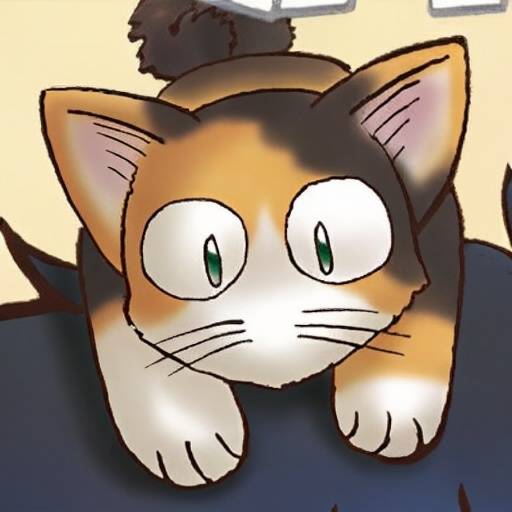} & 
    \includegraphics[width=\tuneimgwidthx]{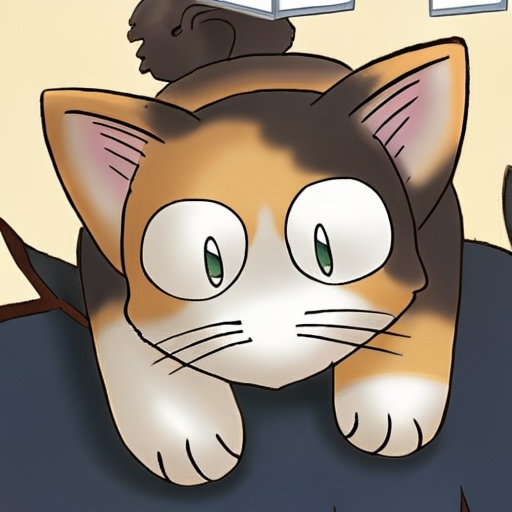}

     \\
          Input & \small{\emph{Remove all degradation}} 
     & \small{\emph{Upsample... Deblur...}} 
     &
     
     Input & \small{\emph{Remove all degradation}} 
     & \small{\emph{Upsample...Deblur...Dejpeg}} 
     \\

\end{tabular}
}
}
\vspace{-1em}
\caption{\textbf{Restoration prompting for images from internet}.   }
\label{fig:midjourney_restoration}
\end{figure}

%% file: sec/figs/sec4_exp/degradation_prompt_1.tex
\vspace{-3em}

\begin{figure*}[h]
  \centering
   \includegraphics[width=0.79\linewidth]{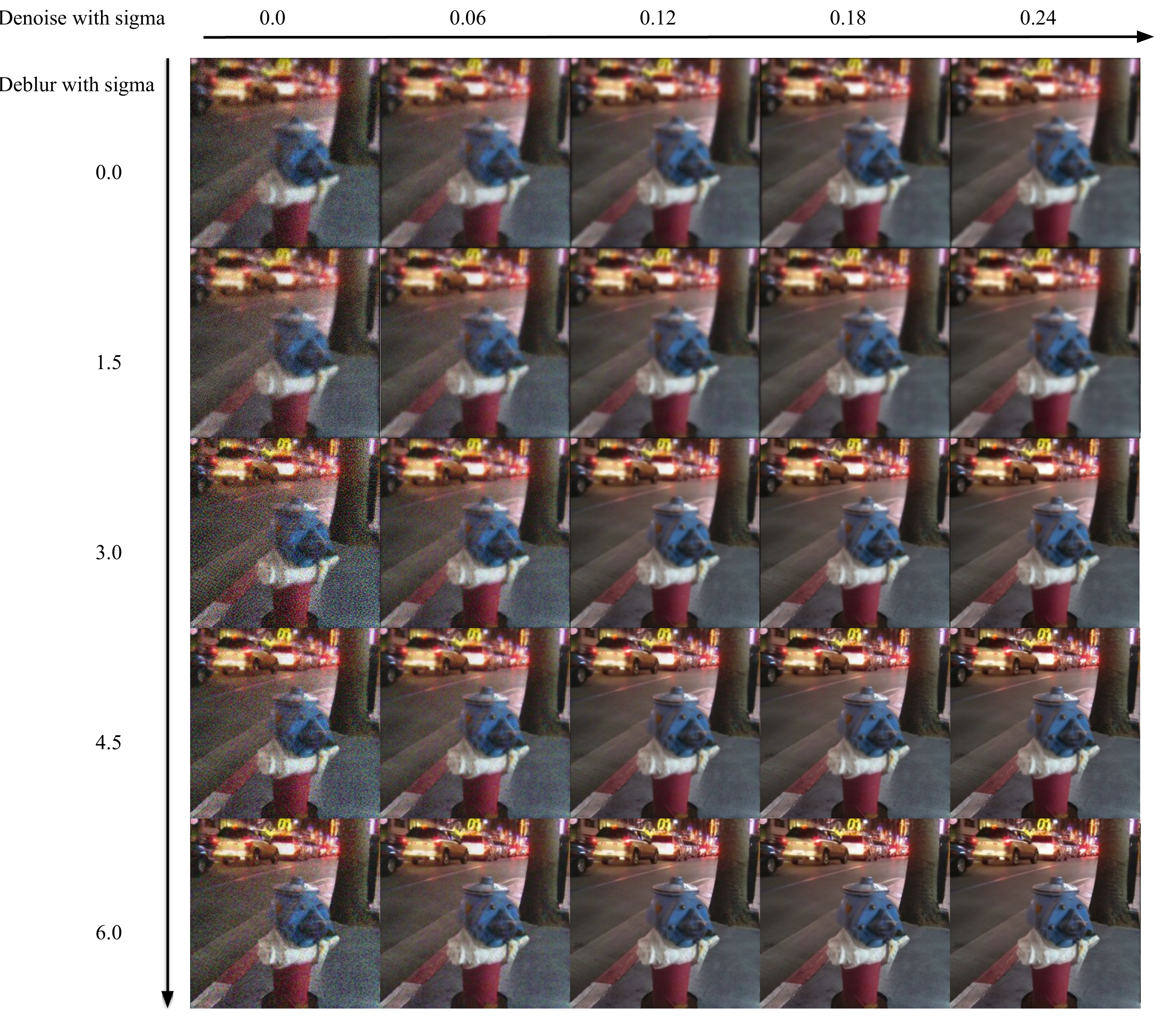}
    \vspace{-0.5em}
   \caption{\textbf{Prompt space walking visualization for the restoration prompt}. Given the same degraded input (upper left) and empty semantic prompt $\varnothing$, our method can decouple the restoration direction and strength via only prompting the \textbf{quantitative number in natural language}. An interesting finding is that our model learns a continuous range of restoration strength from discrete language tokens. 
   }
   \label{fig:restoration_decoupling_supp}
\end{figure*}

%% file: sec/figs/supp/real-input-long-prompt.tex
\newcommand{\imgwidtha}{0.6\columnwidth}

\begin{figure*}[t]
\centering
\setlength\tabcolsep{1pt}
{
\renewcommand{\arraystretch}{0.6}
\resizebox{1\linewidth}{!}{
\begin{tabular}
{*{2}{c@{\hspace{1mm}}}c@{\hspace{3mm}}*{1}{c@{\hspace{1mm}}}}
\\
     Input & Ours SPIRE Model & Input & Ours SPIRE Model \\
 \includegraphics[width=\imgwidtha]{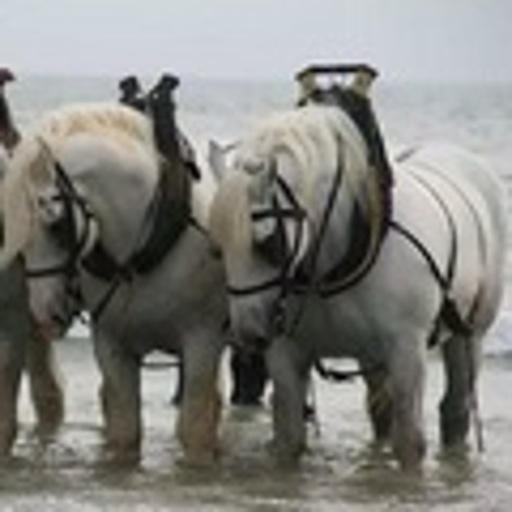} &
     \includegraphics[width=\imgwidtha]{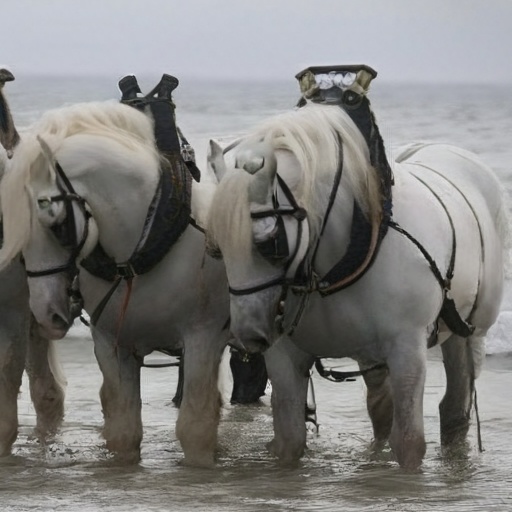} &
    \includegraphics[width=\imgwidtha]{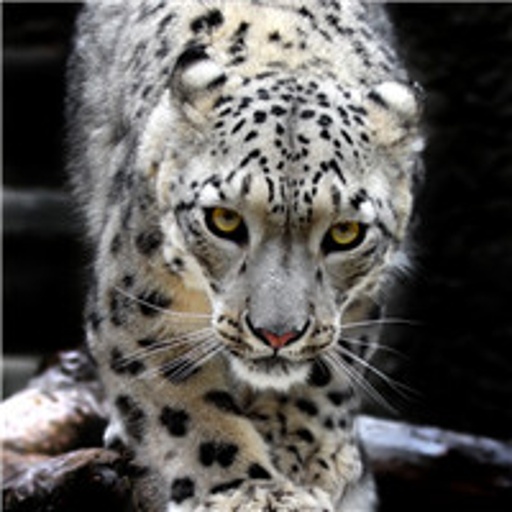} & 
     \includegraphics[width=\imgwidtha]{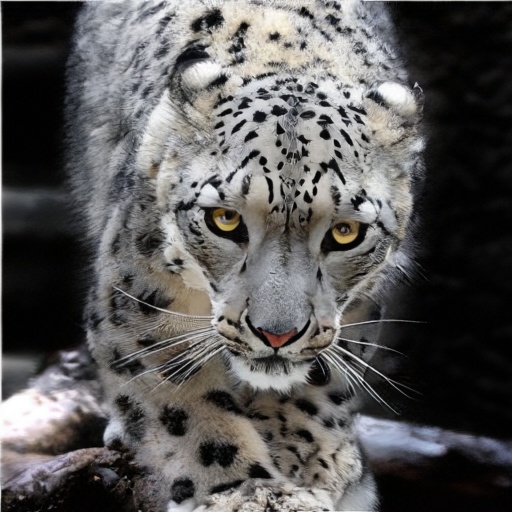} 
     \\
     \multicolumn{2}{c}{LLAVA caption: \emph{A group of white horses standing in shallow water...}} &
      \multicolumn{2}{c}{LLAVA caption: \emph{A close-up of a snow leopard's face...}}
     \\   
     \includegraphics[width=\imgwidtha]{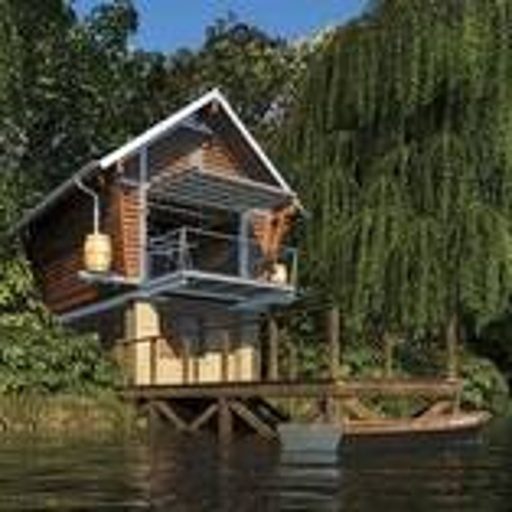} & 
     \includegraphics[width=\imgwidtha]{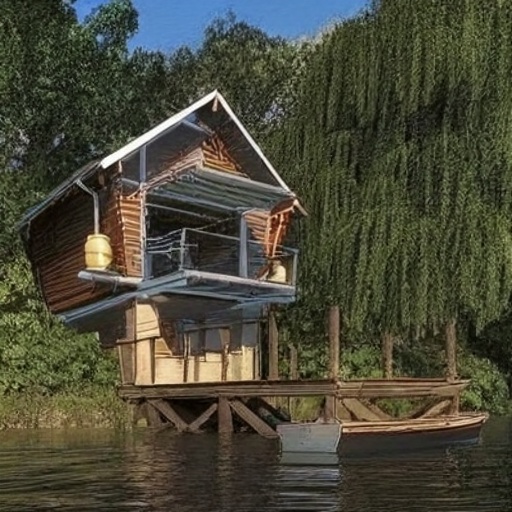} &
     \includegraphics[width=\imgwidtha]{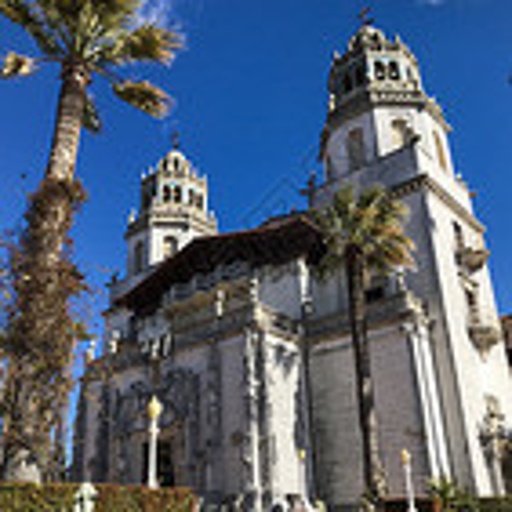} & 
     \includegraphics[width=\imgwidtha]{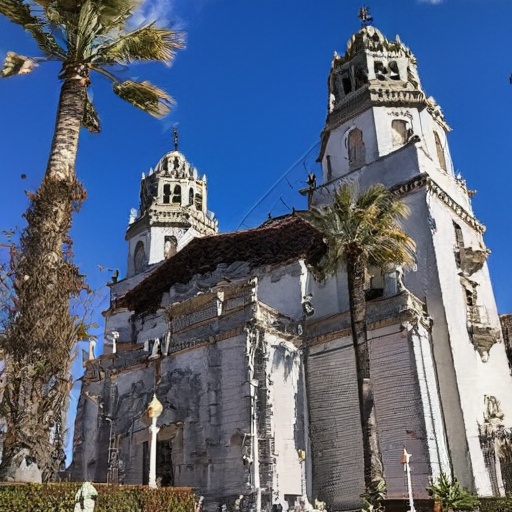}
\\ 
     \multicolumn{2}{c}{LLAVA caption: \emph{A small wooden house situated on a body of water, possibly a
  lake or a river...} } 
    &
     \multicolumn{2}{c}{LLAVA caption: \emph{A large, ornate church with a clock tower and two towers on top...} } 
\\ \includegraphics[width=\imgwidtha]{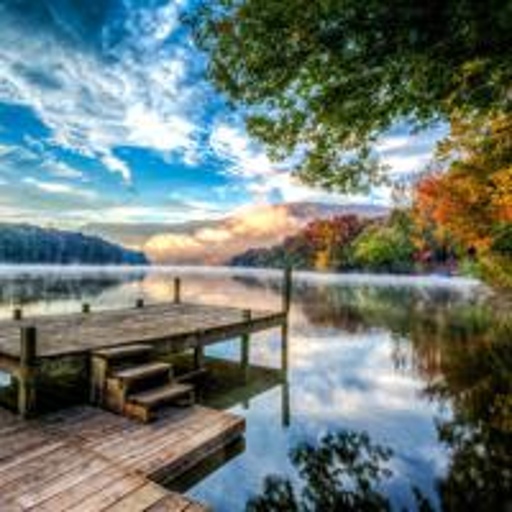} & 
     \includegraphics[width=\imgwidtha]{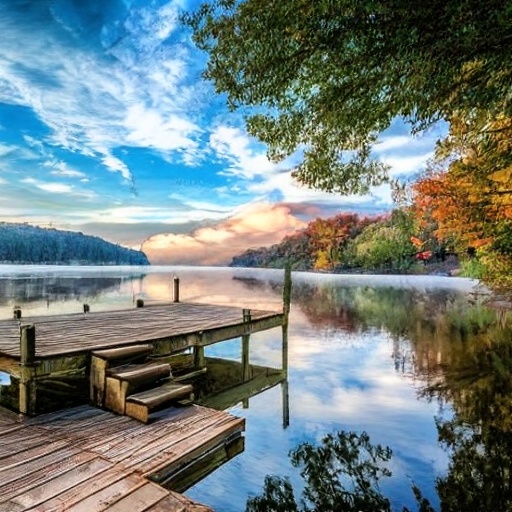} &
      \includegraphics[width=\imgwidtha]{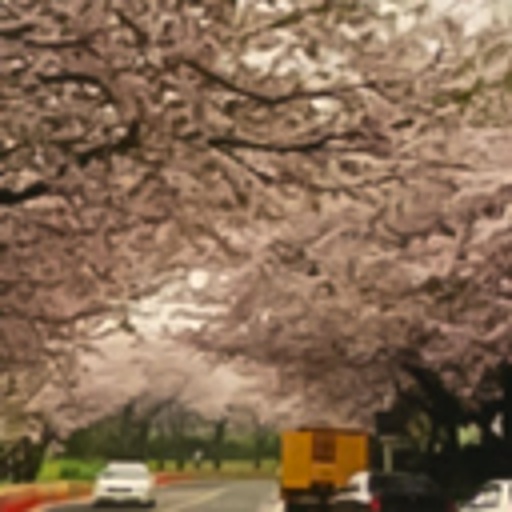} & 
     \includegraphics[width=\imgwidtha]{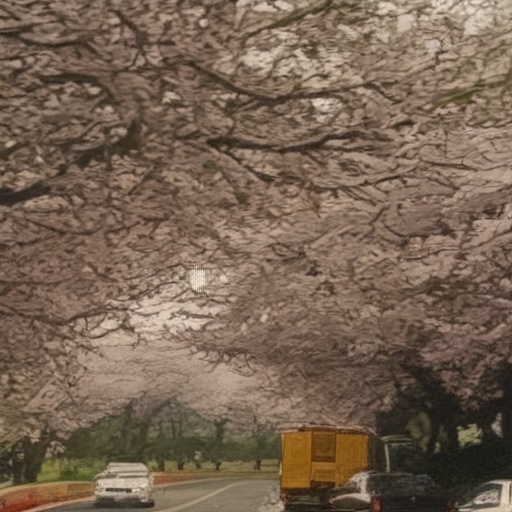}
\\ \multicolumn{2}{c}{LLAVA caption: \emph{A serene scene of a dock situated on a lake....} } 
     &
      \multicolumn{2}{c}{LLAVA caption: \emph{A tree-lined street with a yellow truck, surrounded by a beautiful blossoming tree.} } 
\end{tabular}
}
}
\caption{Qualitative result on real-world images.}
\label{fig:real-input-long-prompt}
\end{figure*}

%% file: sec/figs/sec4_exp/limitation.tex
\newcommand{\tuneimgwidthb}{0.25\columnwidth}

\begin{wrapfigure}{r}{0.7\textwidth}
\centering
\setlength\tabcolsep{1pt}
{
\renewcommand{\arraystretch}{0.6}
\resizebox{1\linewidth}{!}{
\begin{tabular}{@{\hspace{0mm}}*{2}{c}@{\hspace{0mm}}}

    \includegraphics[width=\tuneimgwidthb]{sec/figs/rebuttal/real_images/panda_input.jpg} & 

    \includegraphics[width=\tuneimgwidthb]{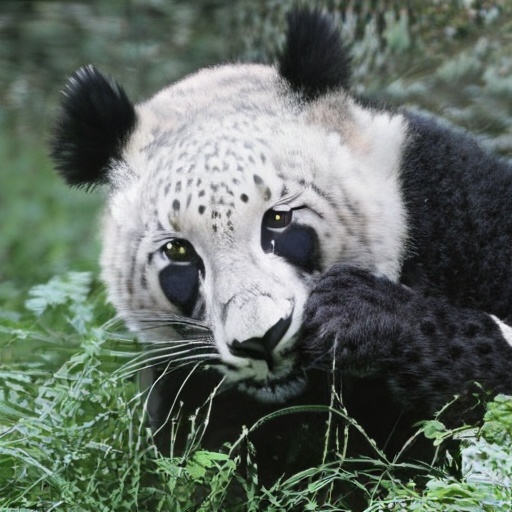}

     \\ 


     Input 
     & \textit{``Snow leopard''}
     
     \\

    
\end{tabular}
}
}
\vspace{-1.2em}
\caption{Hallucinations when the prompt is unmatched with input image restorations.
}
\label{fig:hallucinations}
\vspace{-2em}
\end{wrapfigure}

%% file: sec/figs/supp/real-input-no-prompt-long-prompt.tex
\newcommand{\imgwidthc}{0.35\columnwidth}

\begin{figure*}[t]
\centering
\setlength\tabcolsep{1pt}
{
\renewcommand{\arraystretch}{0.6}
\resizebox{1\linewidth}{!}{
\begin{tabular}
{*{2}{c@{\hspace{1mm}}}c@{\hspace{3mm}}*{4}{c@{\hspace{1mm}}}}

     \includegraphics[width=\imgwidthc]{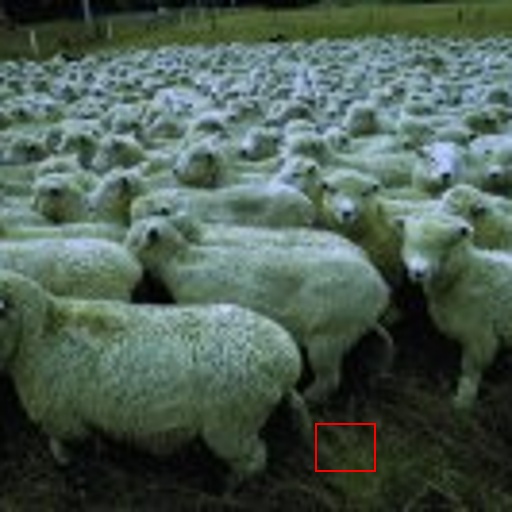} & 
    \includegraphics[width=\imgwidthc]{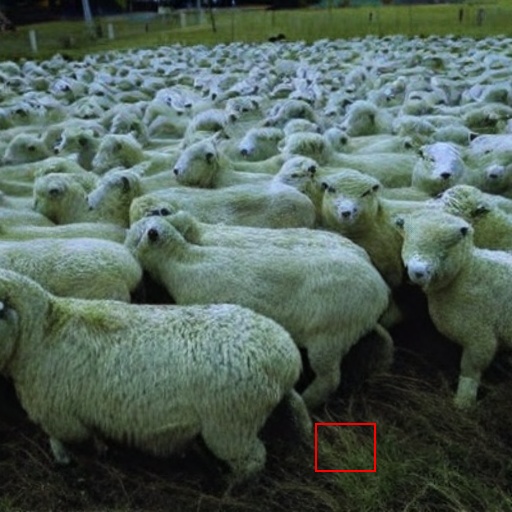} &  
     \includegraphics[width=\imgwidthc]{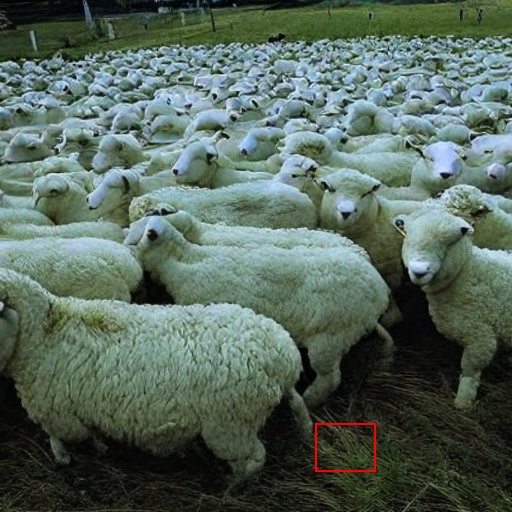} &

     \includegraphics[width=\imgwidthc]{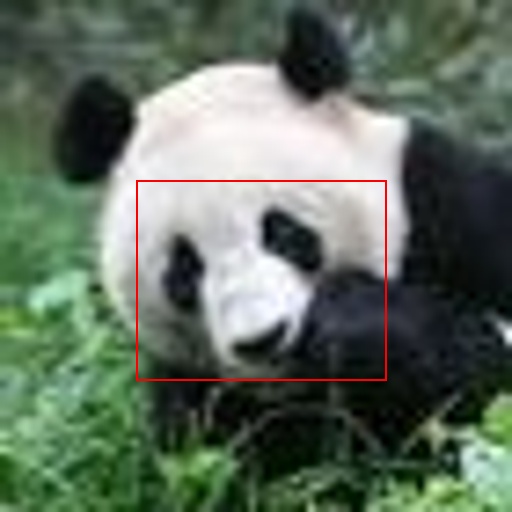} & 
  \includegraphics[width=\imgwidthc]{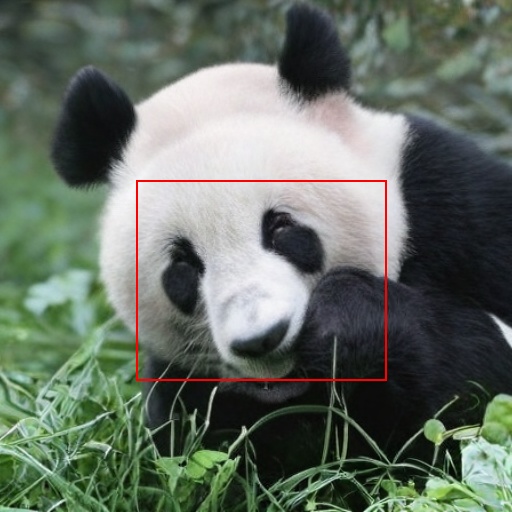} &
     \includegraphics[width=\imgwidthc]{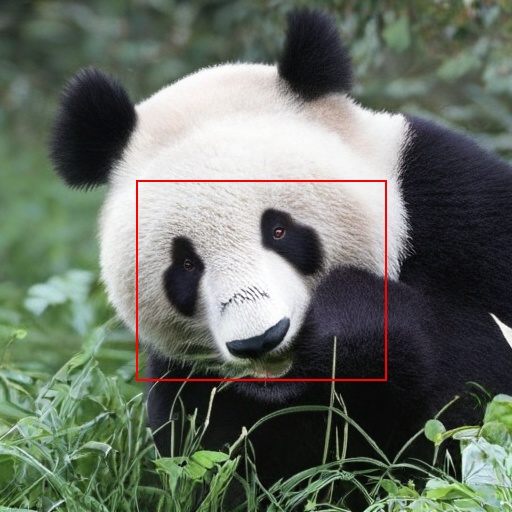} &

\\ 
     \includegraphics[width=\imgwidthc]{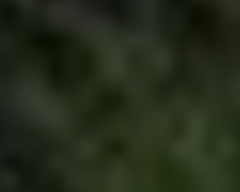} & 
     \includegraphics[width=\imgwidthc]{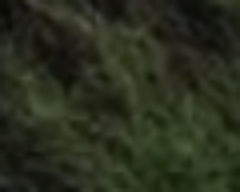} &
     \includegraphics[width=\imgwidthc]{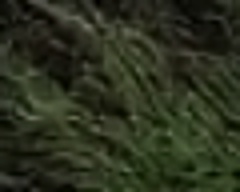} &
     
     \includegraphics[width=\imgwidthc]{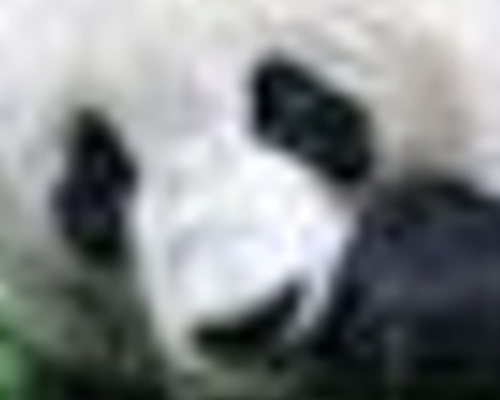} & 
     \includegraphics[width=\imgwidthc]{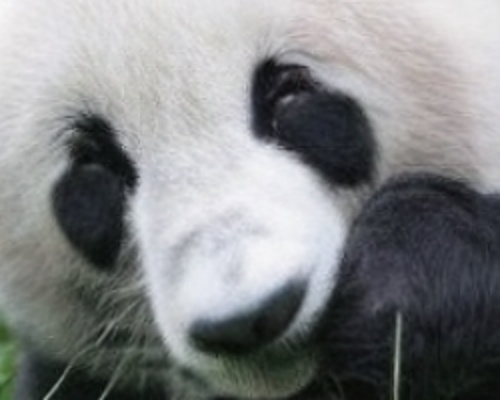} &
     \includegraphics[width=\imgwidthc]{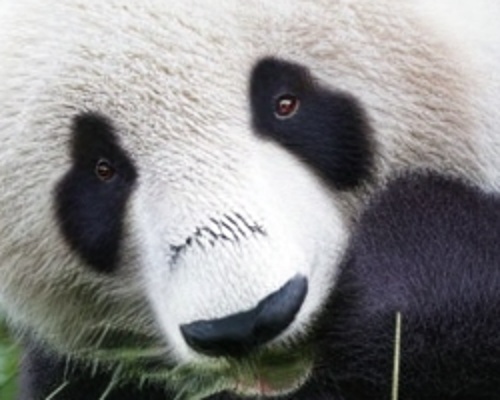} &
\\

     Input & Without semantic prompt &  \emph{sheep in a grassy field ...} &
     Input & Without semantic prompt &  \emph{panda bear ...}

\\
\\
     \includegraphics[width=\imgwidthc]{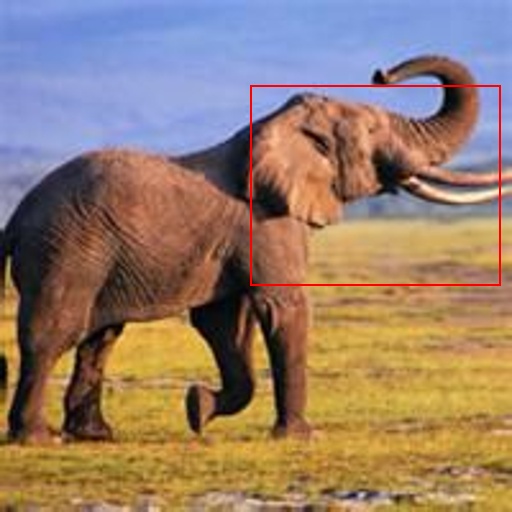} & 
      \includegraphics[width=\imgwidthc]{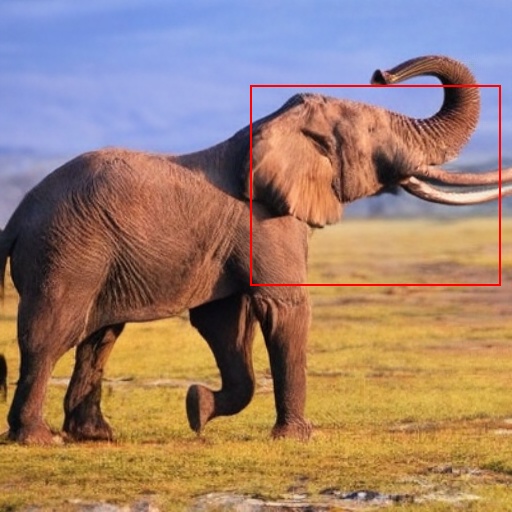} &
     \includegraphics[width=\imgwidthc]{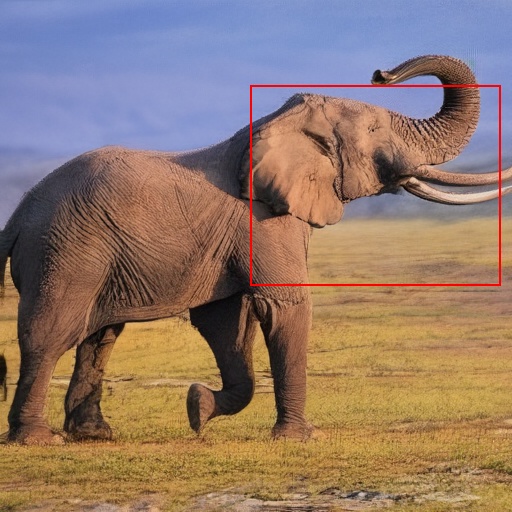} &

     \includegraphics[width=\imgwidthc]{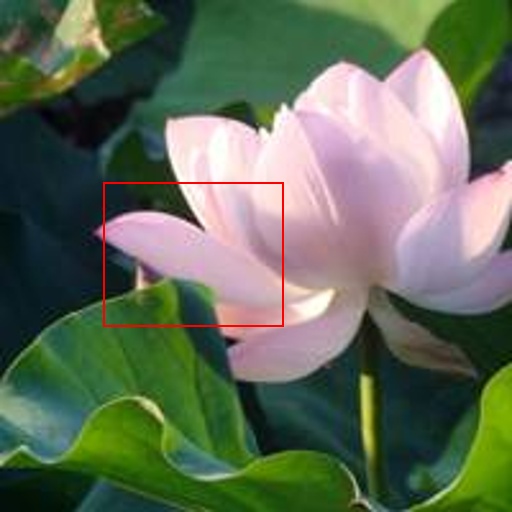} & 

     \includegraphics[width=\imgwidthc]{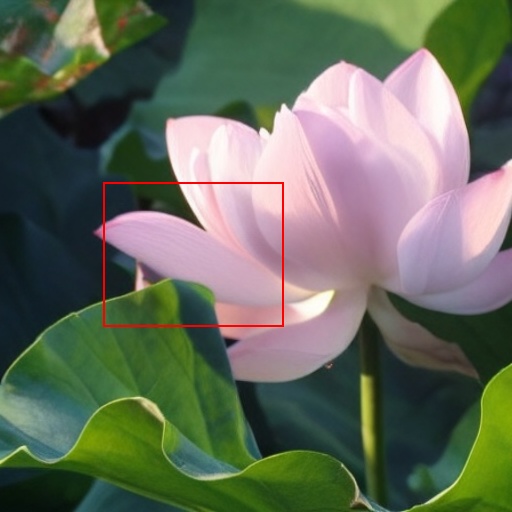} &
      \includegraphics[width=\imgwidthc]{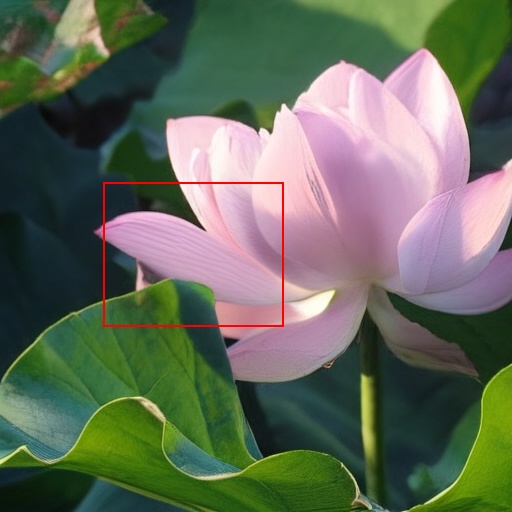} &
     
\\ 
     \includegraphics[width=\imgwidthc]{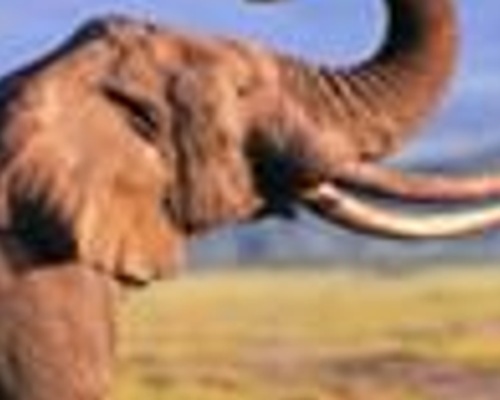} & 
     \includegraphics[width=\imgwidthc]{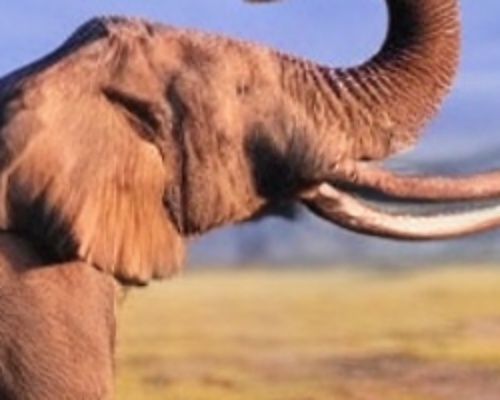} &
     \includegraphics[width=\imgwidthc]{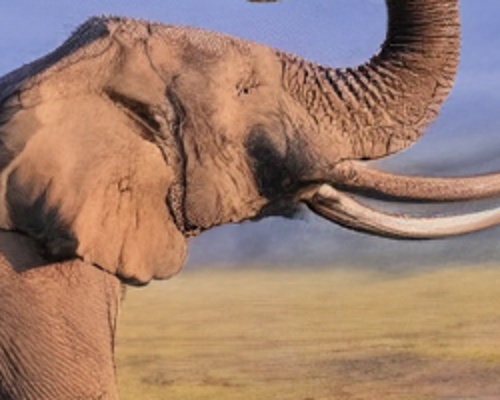} &
     
     \includegraphics[width=\imgwidthc]{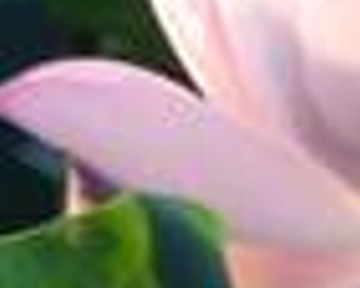} & 
     \includegraphics[width=\imgwidthc]{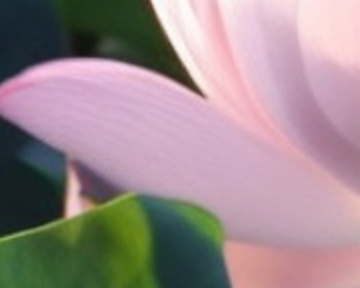} &
     \includegraphics[width=\imgwidthc]{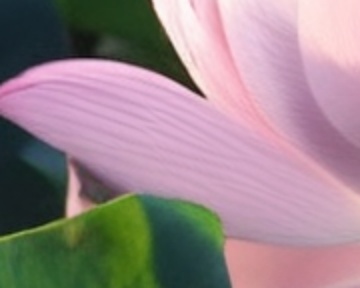} &
\\

     Input & Without semantic prompt & 
     \emph{elephant ... grassy field ...} 
     &
     Input & Without semantic prompt & 
     \emph{pink flower ... delicate petals ...} 
\\
\\

     \includegraphics[width=\imgwidthc]{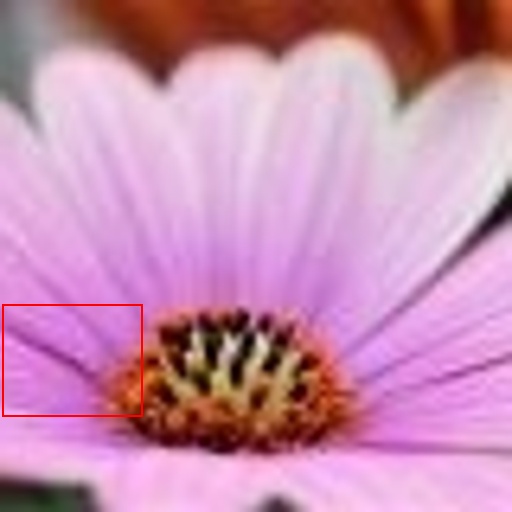} & 
      \includegraphics[width=\imgwidthc]{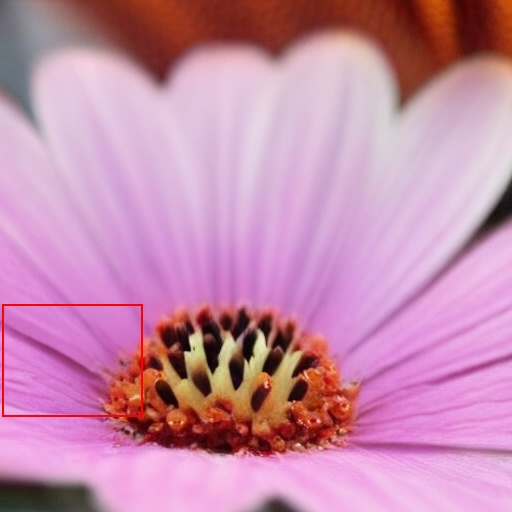} &
     \includegraphics[width=\imgwidthc]{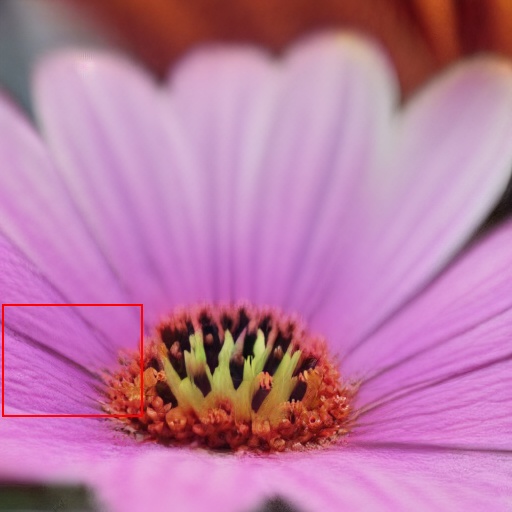} &

     \includegraphics[width=\imgwidthc]{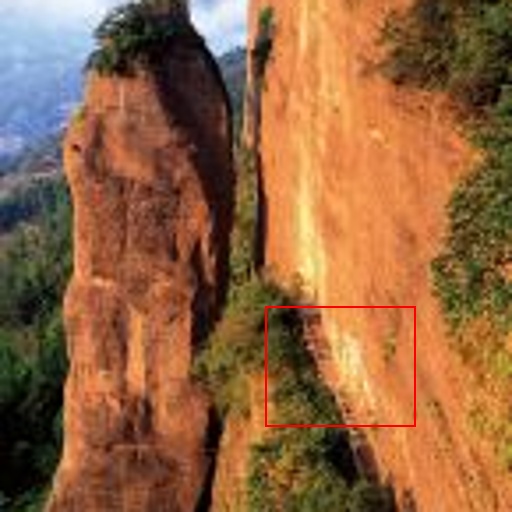} & 
  \includegraphics[width=\imgwidthc]{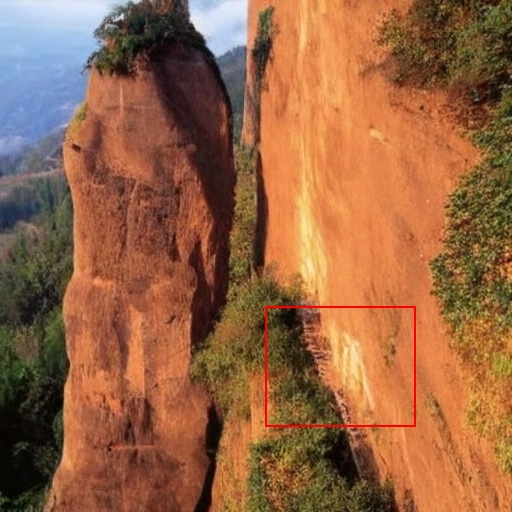} &
     \includegraphics[width=\imgwidthc]{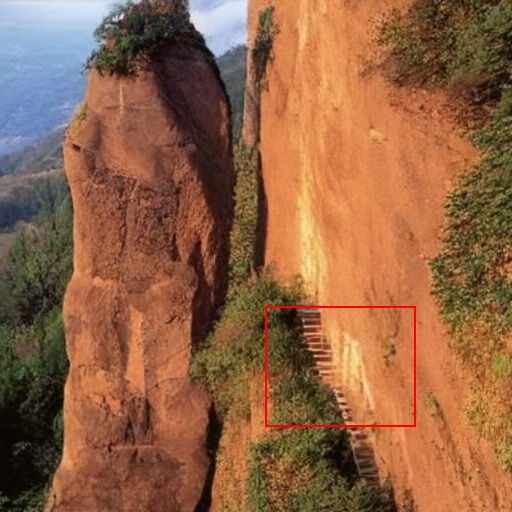} &

\\ 
     \includegraphics[width=\imgwidthc]{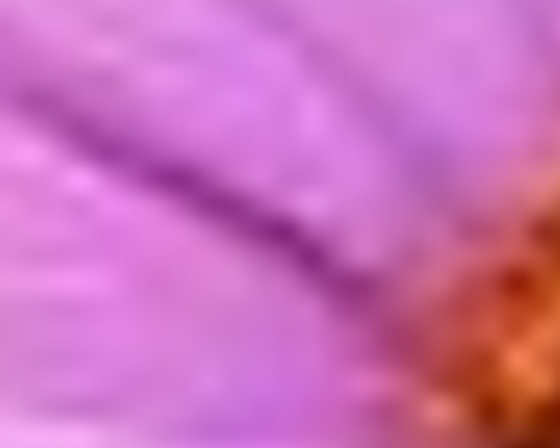} & 
     \includegraphics[width=\imgwidthc]{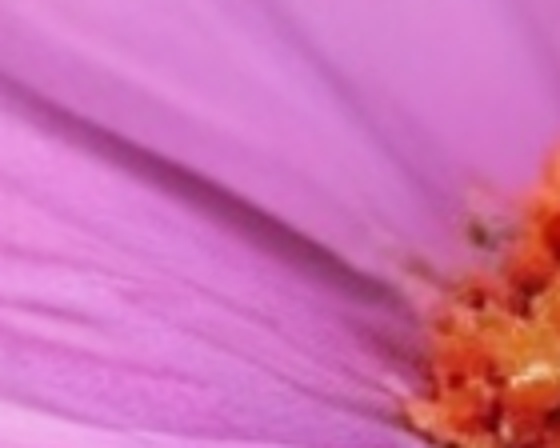} &
     \includegraphics[width=\imgwidthc]{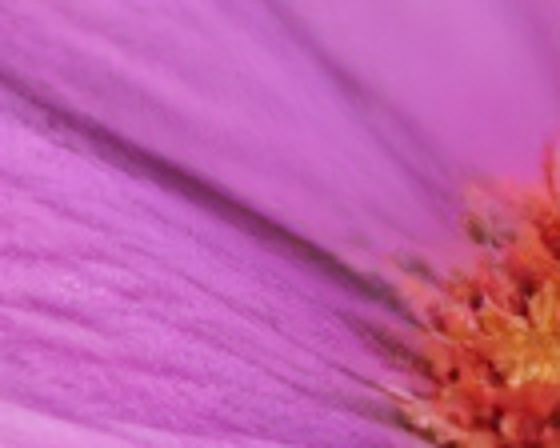} &
     
     \includegraphics[width=\imgwidthc]{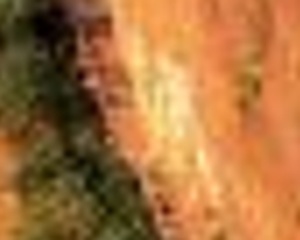} & 
     \includegraphics[width=\imgwidthc]{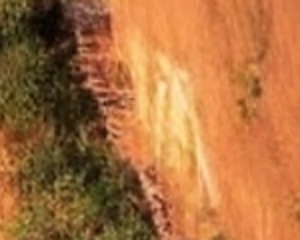} &
     \includegraphics[width=\imgwidthc]{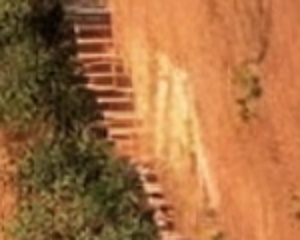} &
\\
     Input & Without semantic prompt & 
     \emph{purple flower...pink petal...}
     &
     Input & Without semantic prompt &
     \emph{rocky cliff with a staircase ... }
\end{tabular}
}
}
\caption{Real-world examples showing the effect of semantic prompts.}
\label{fig:real-input-no-prompt-long-prompt}
\end{figure*}

%% file: sec/figs/rebuttal/r1q2_derain_dehaze.tex
\begin{figure}[H]
\centering
{
\vspace{-1em}
\includegraphics[width=0.8\linewidth]{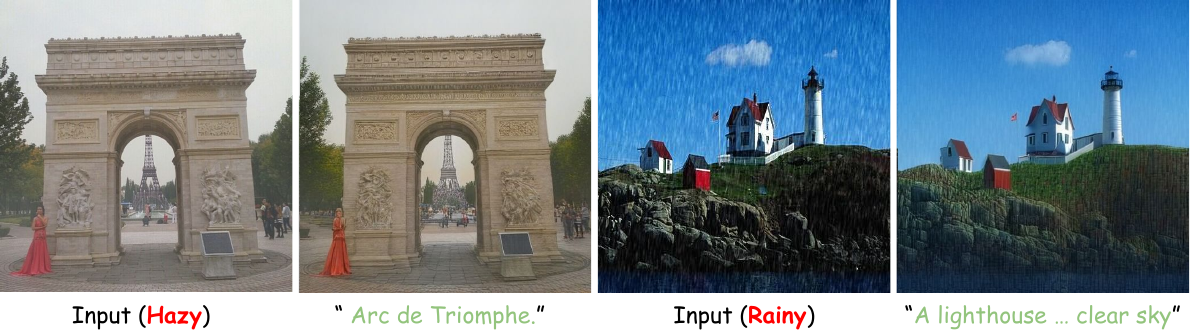}
}

\vspace{-1em}
\caption{Image restoration for unseen degradations.}
\end{figure}

%% file: sec/tabs/rebuttal_task_comparison.tex
\begin{table}[t]
\footnotesize

\centering
\setlength{\tabcolsep}{6pt}
\begin{tabular} 
{lccc}
Model / Task & $\times 4 $ super-resolution  & denoising & de-jpeg
\\
\toprule
SwinIR (task-specific) & 0.309 & 0.361 & 0.319 \\
Ours & \textbf{0.265} & \textbf{0.305} & \textbf{0.214} \\
\bottomrule
\end{tabular}
\caption{Comparison with task-specific SwinIR.}
\vspace{-10mm}

\label{tab:swinir_comparison}
\end{table}

%% file: sec/figs/sec4_exp/main_visuals_supp_merge.tex
\newcommand{\imgwidthx}{0.27\columnwidth}

\begin{figure*}[t]
\centering
\vspace{-2em}
\setlength\tabcolsep{1pt}
{
\renewcommand{\arraystretch}{0.6}
\resizebox{1\linewidth}{!}{
\vspace{-1em}
\begin{tabular}{@{}*{7}{c}@{}}
     Input & DiffBIR~\cite{Lin2023diffbir} & DiffBIR + SDEdit~\cite{sdedit} & DiffBIR + CLIP~\cite{radford2021clip} & ControlNet-SR~\cite{luo2023controlling} & Ours \TIP{} Model & Ground-Truth \\

     \includegraphics[width=\imgwidthx]{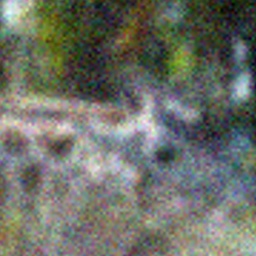} &
     \includegraphics[width=\imgwidthx]{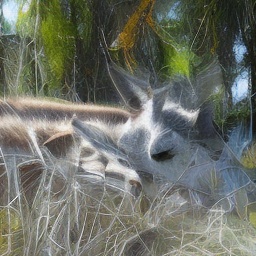} & 
     \includegraphics[width=\imgwidthx]{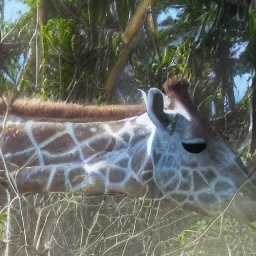} &
      \includegraphics[width=\imgwidthx]{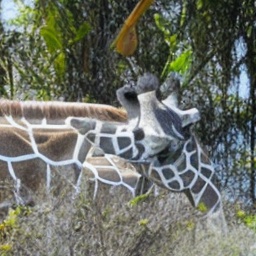} &
     \includegraphics[width=\imgwidthx]{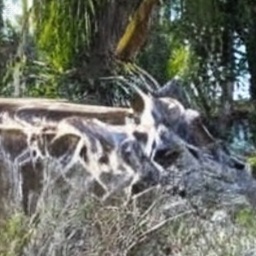} & 
     \includegraphics[width=\imgwidthx]{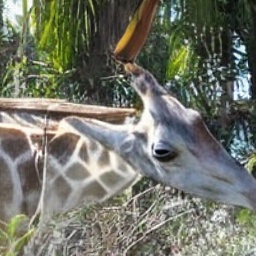} 

     & \includegraphics[width=\imgwidthx]{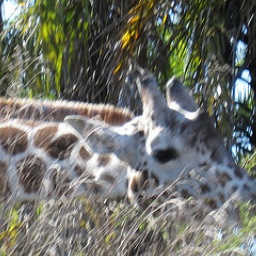}\\
     
     \multicolumn{7}{c}{\emph{A tall giraffe eating leafy greens in a jungle.}} \\

     \includegraphics[width=\imgwidthx]{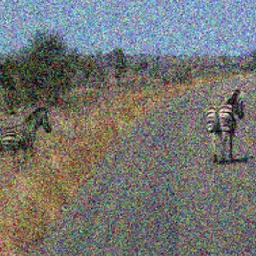} & 
     \includegraphics[width=\imgwidthx]{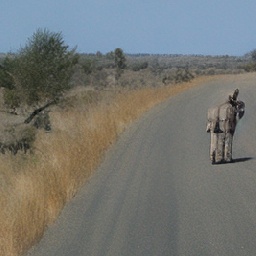} &
     \includegraphics[width=\imgwidthx]{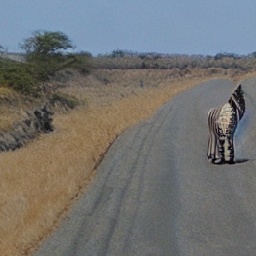} &
      \includegraphics[width=\imgwidthx]{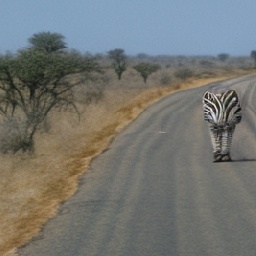} &
     \includegraphics[width=\imgwidthx]{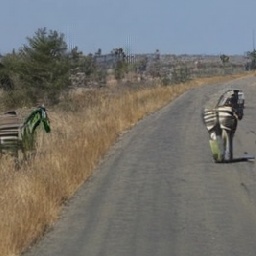} & 
     \includegraphics[width=\imgwidthx]{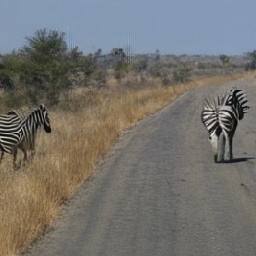} 
     & \includegraphics[width=\imgwidthx]{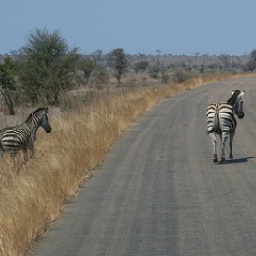}\\ 

     \multicolumn{7}{c}{\emph{Zebras crossing a bitumen road in the savannah.}} \\     

     \includegraphics[width=\imgwidthx]{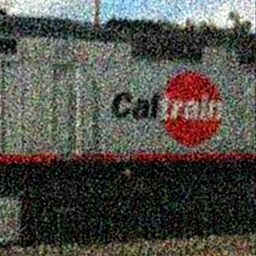} & 
     \includegraphics[width=\imgwidthx]{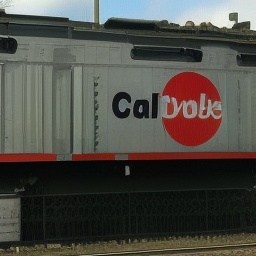} &
     \includegraphics[width=\imgwidthx]{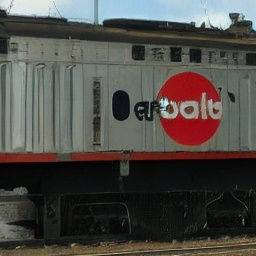} & 
     \includegraphics[width=\imgwidthx]{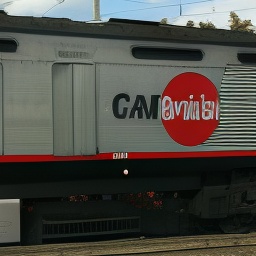} & 
     \includegraphics[width=\imgwidthx]{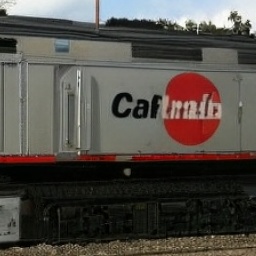} 
     & \includegraphics[width=\imgwidthx]{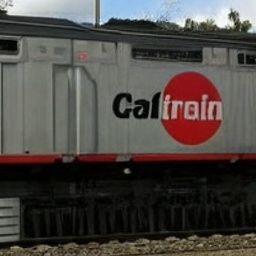} & \includegraphics[width=\imgwidthx]{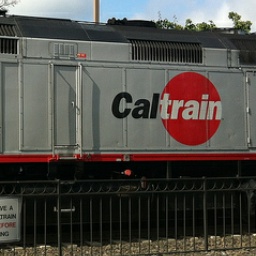}\\
     \multicolumn{7}{c}{\emph{A gray train riding on a track.}} \\
     
     \includegraphics[width=\imgwidthx]{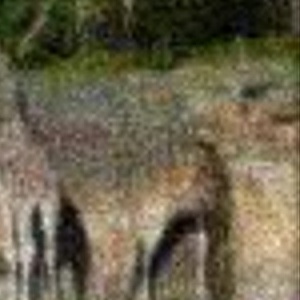} & 
     \includegraphics[width=\imgwidthx]{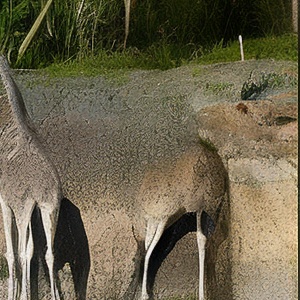} &
     \includegraphics[width=\imgwidthx]{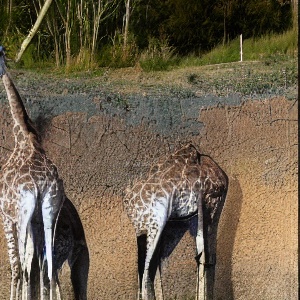} &
      \includegraphics[width=\imgwidthx]{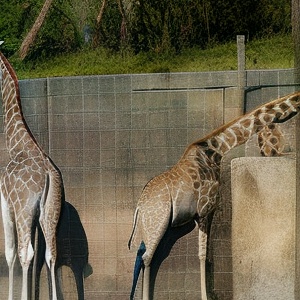} &
     \includegraphics[width=\imgwidthx]{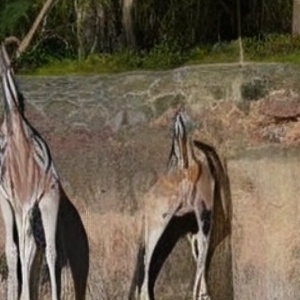} & 
     \includegraphics[width=\imgwidthx]{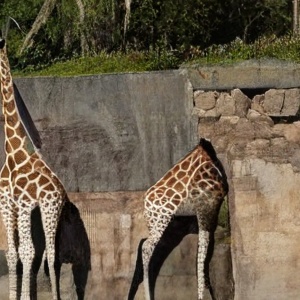} 
     & \includegraphics[width=\imgwidthx]{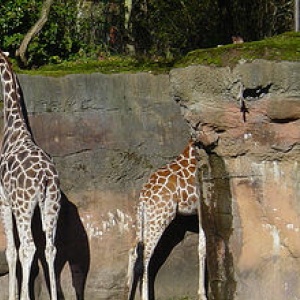}\\ 
     \multicolumn{7}{c}{\emph{Two giraffes are standing together outside near a wall.}} \\



     \includegraphics[width=\imgwidthx]{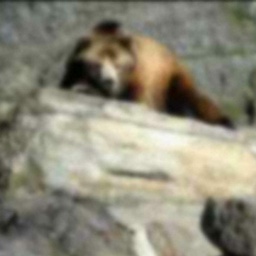} & 
     \includegraphics[width=\imgwidthx]{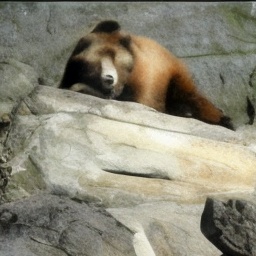} &
     \includegraphics[width=\imgwidthx]{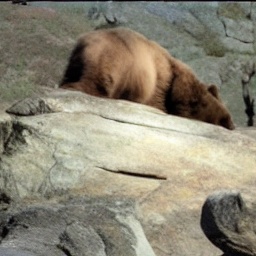} &
     \includegraphics[width=\imgwidthx]{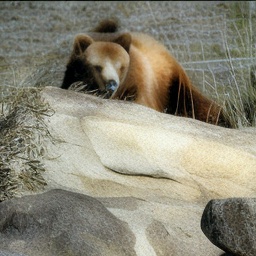} &
     \includegraphics[width=\imgwidthx]{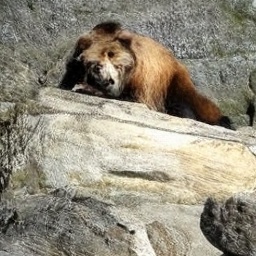} & 
     \includegraphics[width=\imgwidthx]{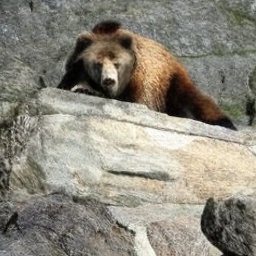}  & \includegraphics[width=\imgwidthx]{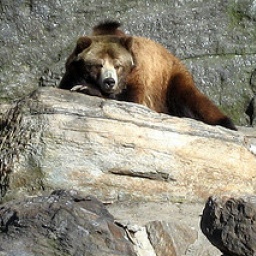}\\ 

     \multicolumn{7}{c}{\emph{two brown bears on some rocks.}} \\



     \includegraphics[width=\imgwidthx]{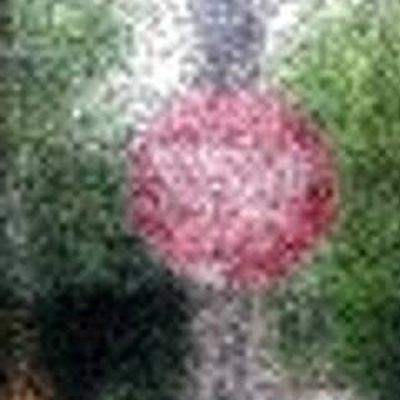} & 
     \includegraphics[width=\imgwidthx]{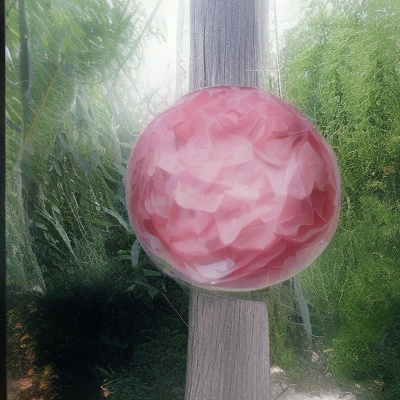} &
     \includegraphics[width=\imgwidthx]{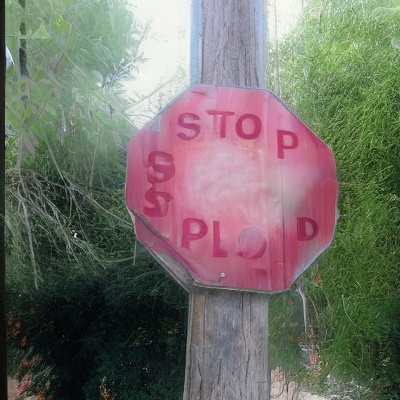} &
     \includegraphics[width=\imgwidthx]{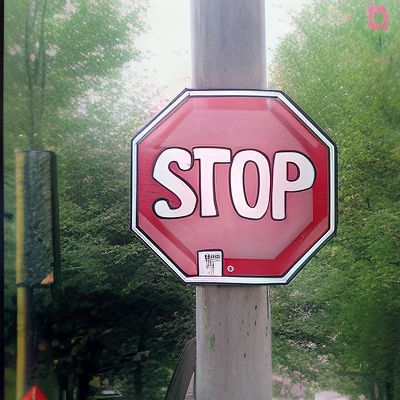} &
     \includegraphics[width=\imgwidthx]{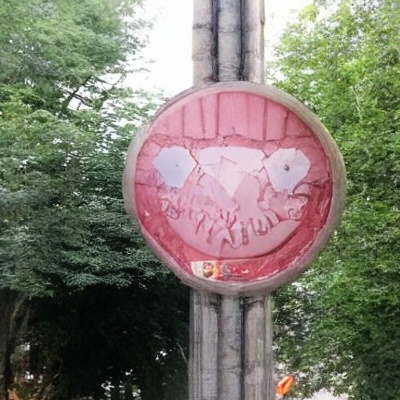} & 
     \includegraphics[width=\imgwidthx]{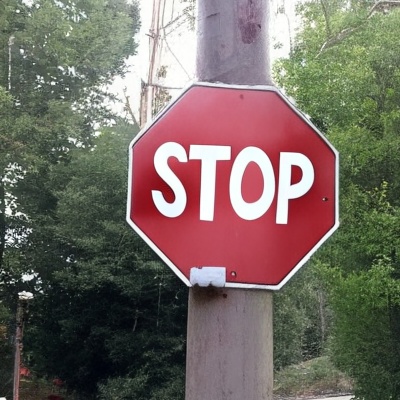}  & \includegraphics[width=\imgwidthx]{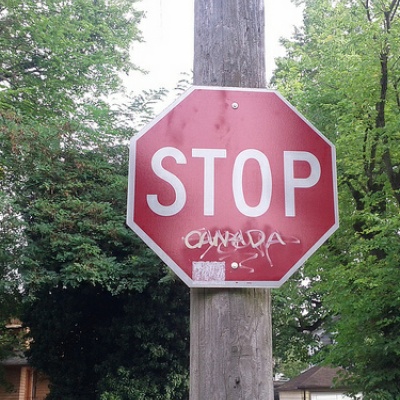}\\ 

     \multicolumn{7}{c}{\emph{A stop sign with graffiti on it nailed to a pole.}} \\
     
     \includegraphics[width=\imgwidthx]{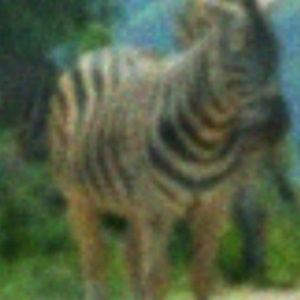} &
     \includegraphics[width=\imgwidthx]{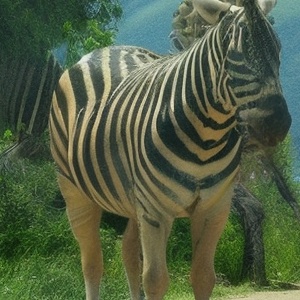} & 
     \includegraphics[width=\imgwidthx]{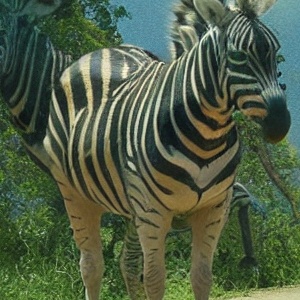} &
      \includegraphics[width=\imgwidthx]{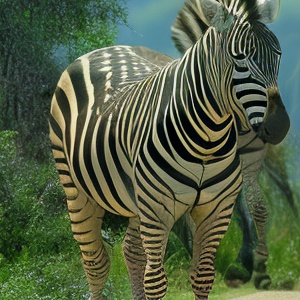} &
     \includegraphics[width=\imgwidthx]{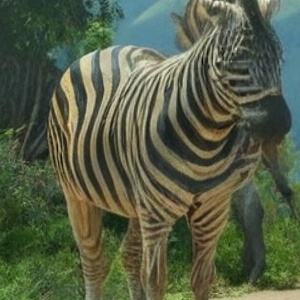} & 
     \includegraphics[width=\imgwidthx]{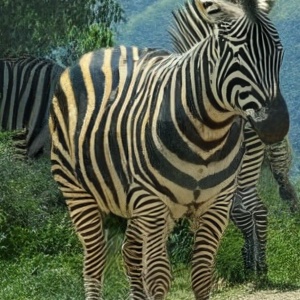} 
     & \includegraphics[width=\imgwidthx]{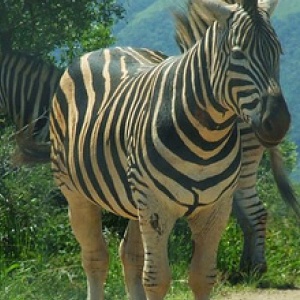}\\
     \multicolumn{7}{c}{\emph{Two zebras are heading into the bushes as another is heading in the opposite direction.}} \\


     \includegraphics[width=\imgwidthx]{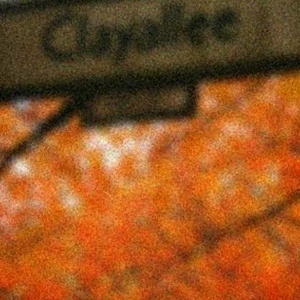} & 
     \includegraphics[width=\imgwidthx]{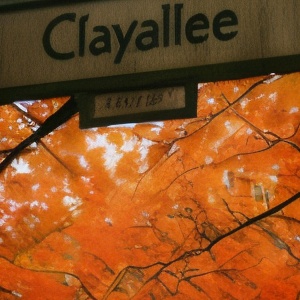} &
     \includegraphics[width=\imgwidthx]{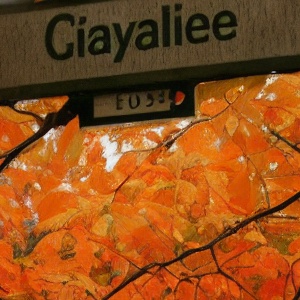} &
      \includegraphics[width=\imgwidthx]{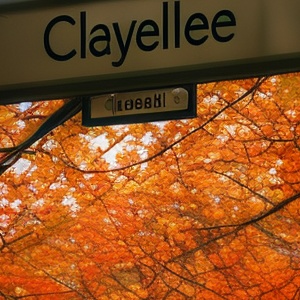} &
     \includegraphics[width=\imgwidthx]{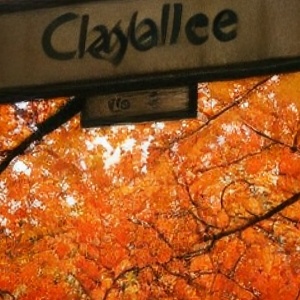} & 
     \includegraphics[width=\imgwidthx]{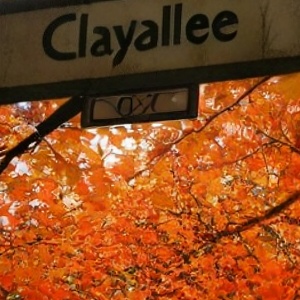} 
     & \includegraphics[width=\imgwidthx]{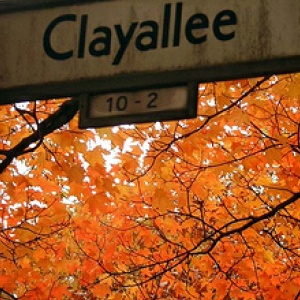}\\ 
     \multicolumn{7}{c}{\emph{A street sign surrounded by orange and red leaves.}} \\
     
     \includegraphics[width=\imgwidthx]{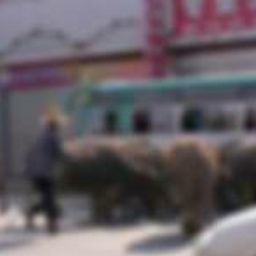} &
     \includegraphics[width=\imgwidthx]{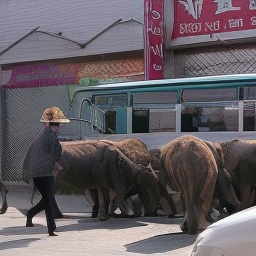} & 
     \includegraphics[width=\imgwidthx]{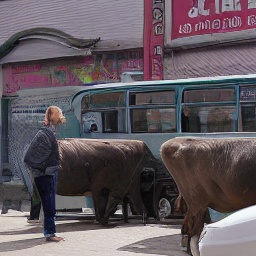} &
      \includegraphics[width=\imgwidthx]{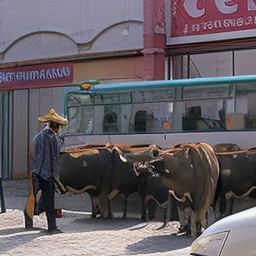} &
     \includegraphics[width=\imgwidthx]{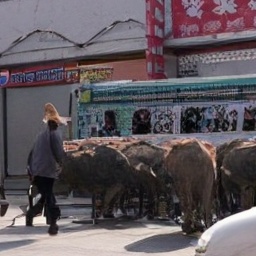} & 
     \includegraphics[width=\imgwidthx]{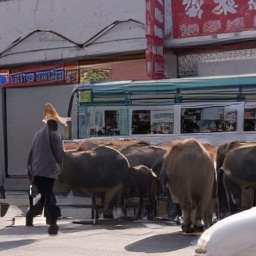} 
     & \includegraphics[width=\imgwidthx]{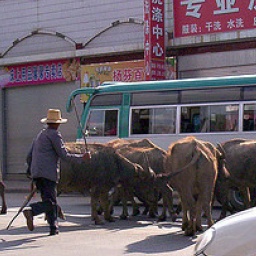}\\   

     \multicolumn{7}{c}{\emph{A group of cows on street next to building and bus.}} \\

\end{tabular}
}
\vspace{-1em}
}
\caption{Main visual comparison with baselines. (Zoom in for
details)    }
\label{fig:main-visuals-supp1}
\end{figure*}